\documentclass[lettersize,journal]{IEEEtran}
\usepackage{amsmath,amsfonts}
\usepackage{algorithmic}
\usepackage{algorithm}
\usepackage{array}
\usepackage{subcaption}
\usepackage{textcomp}
\usepackage{stfloats}
\usepackage{url}
\usepackage{verbatim}
\usepackage{graphicx}
\usepackage{xcolor}

\usepackage{cite}
\begin{document}

\title{Exploiting Expertise of Non-Expert and Diverse Agents in Social Bandit Learning: A Free Energy Approach}

\author{Erfan Mirzaei, Seyed Pooya Shariatpanahi, Alireza Tavakoli, Reshad Hosseini, Majid Nili Ahmadabadi~\IEEEmembership{}
\thanks{This is the accepted version of the paper.The final published version is available via IEEE Xplore: https://doi.org/10.1109/TCDS.2025.3648042.\\The authors are with the Cognitive Systems Lab., School of Electrical and Computer Engineering, College of Engineering, University of Tehran, Tehran, Iran.}}
\markboth{}%
{Shell \MakeLowercase{\textit{et al.}}: A Sample Article Using IEEEtran.cls for IEEE Journals}


\maketitle

\begin{abstract}
Personalized AI-based services involve a population of individual reinforcement learning agents. However, most reinforcement learning algorithms focus on harnessing individual learning and fail to leverage the social learning capabilities commonly exhibited by humans and animals. Social learning integrates individual experience with observing others' behavior, presenting opportunities for improved learning outcomes. In this study, we focus on a social bandit learning scenario where a social agent observes other agents' actions without knowledge of their rewards. The agents independently pursue their own policy without explicit motivation to teach each other. We propose a free energy-based social bandit learning algorithm over the policy space, where the social agent evaluates others' expertise levels without resorting to any oracle or social norms. Accordingly, the social agent integrates its direct experiences in the environment and others' estimated policies. The theoretical convergence of our algorithm to the optimal policy is proven. Empirical evaluations validate the superiority of our social learning method over alternative approaches in various scenarios. Our algorithm strategically identifies the relevant agents, even in the presence of random or suboptimal agents, and skillfully exploits their behavioral information. In addition to societies including expert agents, in the presence of relevant but non-expert agents, our algorithm significantly enhances individual learning performance, where most related methods fail. Importantly, it also maintains logarithmic regret. 
\end{abstract}


\section{Introduction}
Reinforcement learning (RL) in general, and bandit learning as a single-state RL in particular, is rooted in the study of animal and human learning \cite{thorndike2017animal,schultz1997neural}. However, humans and several animals, as social beings, employ a diverse set of social learning methods in harmony with RL. This type of learning is used to make optimal decisions in bandits and acquire complex behavior, resulting in faster adaptation to new tasks and environments, and to explore safer and more efficiently \cite{bandura1977social, laland2004social, SocialAINature}. 
 
 The complexity of human societies would not have been possible if they relied solely on individual learning to solve everyday decision-making problems \cite{SecretOfSuccess, laland2017, boyd2011cultural}. Ignoring the social aspects of reinforcement learning algorithms can be a contributing factor to their poor performance compared to humans, particularly in simple tasks. 

 Additionally, due to the rapid advances in human-in-the-loop technologies \cite{mosqueira2023human, chai2020human} and AI-based personalization of products and services, interactions between humans and artificial intelligence (AI) have become an integral part of daily life \cite{tsiakas2024unpacking}. In this setting, we encounter societies of diverse artificial interactive learning agents that engage in complex bandit tasks. These factors present an opportunity to develop an effective combination of individual and social learning, addressing the main disadvantages of bandit learning, namely, high regret and slow learning. 
 
According to the theory of social learning, two different resources contribute to the learning of new tasks and skills: direct interaction with the environment and observation of others' behavior and consequences \cite{bandura1977social}. The former is more closely related to reinforcement learning, while the latter is more associated with observational learning in the literature. This research focuses on applying social learning to bandit problems, which offers a simple yet elegant framework for studying the explore-exploit trade-off faced by an agent in an unknown environment. 

Our setting, known as Social Bandit Learning (SBL), closely resembles many real-world problems, including personal AI assistants and personal education systems. In personal education systems, each student interacts with an AI-powered tutor. These tutors can observe each other's actions but not the rewards and other private data. The tutors can form a society while interacting with humans who may possess diverse expertise and utilities. The same is true for personalized recommender systems. These systems can form a society, too.  By integrating social learning methods in parallel with individual learning, we can achieve lower regret and faster learning. 

Our setting involves a Social Agent (SA) that observes the actions of other agents, called individual agents (IAs), in the same environment, without access to their corresponding rewards and other private information. The agents independently pursue their own policy without any direct incentive to teach one another, and we make no assumptions about their expertise or the relevance of their policy to our SA's task.
Therefore, in the absence of any social evaluations and norms and due to reward privacy, the SA should mainly evaluate other agents' relevance as well as the levels and scope of their expertise based on a self-referenced evaluation \cite{yaman2022meta, witt2024humans}.

The challenge here is that, due to a lack of sufficient experience in early trials, typical measures for self-referenced evaluation are not accurate enough in the early stages of learning. This leads to ignoring social learning or high regret in the early trials. Our solution to this challenge is to include global and absolute measures for assessing the suitability of information besides self-referenced evaluation. The next challenge for the SA is how to integrate its own direct experiences (action and reward) and the observed behavior data of others to enhance its policy and reduce its regret. In this research, we propose a free energy solution in the policy space that resolves the two mentioned challenges in a unified method.  Free energy minimization models the behavior of bounded-rational decision-makers, aiming to strike a balance between expected utility and information-processing cost \cite{ortega2013thermodynamics}. The free energy in our method includes the entropy of candidate behavior policy as a measure of suitability for the observed behavior and the divergence of candidate policies from a reference policy as a self-referenced evaluation measure.  


Our approach differentiates from other multi-agent reinforcement learning methods by not assuming that agents can share private information, such as rewards, observations, or gradients, during training. Such assumptions are unlikely to hold in real-world social learning scenarios that involve competing agents, agents with intrinsic motivations, and agents from different companies. Importantly, these assumptions may not generalize to human-in-the-loop or human-AI interaction scenarios \cite{jaques2019social, hawkins2023flexible}. It is worth noting that this problem should be distinguished from other multi-agent bandit problems, specifically cooperative bandits, where several agents share information about actions and rewards obtained \cite{landgren2016distributed}. Our method inherently includes imitative learning and observational learning but it is neither limited to the presence of an expert nor to sharing a common utility function among the agents.  


In the next section, we review the literature related to social learning in the RL bandit framework. We then present the problem statement and the assumptions. For the sake of completeness, we explain the TS policy estimation as well as the free energy model of decision-making and mention the individual learning methods employed in this research. Subsequently, we describe our method and demonstrate its convergence to the optimal policy. The experimental results in various scenarios are presented next. Lastly, we analyze the empirical findings, discuss their implications and limitations, and propose promising research directions.

\section{Related Work}

Learning from other agents, whether artificial or human beings, has been extensively studied across diverse research domains. In complex robotics tasks, imitation learning has gained popularity. Imitation learning methods aim to train an agent to closely approximate an expert's behavior (see \cite{duan2017one, hua2021learning} as recent examples). However, pure imitation learning methods can be prone to failure when even minor changes occur in the environment. To mitigate this issue, these methods are often combined with RL or use multiple experts \cite{cheng2020policy, lerer2019learning}.  

Having a well-identified expert in society, as well as assuming either homogeneity or having a mapping to handle heterogeneity among the expert and the learner, is hard to achieve in real-world problems. As a result, identification of an expert, if one exists, falls upon the SA when the society of IAs is heterogeneous; e.g., agents have diverse utility functions, which is the case in many real-world applications, including autonomous driving and personalized services \cite{filos2021psiphilearning}. For this reason, an expert for one social learner may not be an expert for others. All said aside, in many real-world problems, the task is either novel to all learners or no individual learner can become an expert with a budgeted interaction opportunity \cite{ndousse2021emergent}.
Furthermore, curated expert trajectories can be unavailable, and imitation learning may not be suitable. A social learning method that can identify and exploit partial expertise can be beneficial for regret reduction and cover imitation learning as well. However, unlike some basic imitation learning methods, social learning methods do not condition the learner's behavior to the expert policy.  

In a similar vein, some papers study the effect of knowledge transfer (see \cite{zhu2023transfer}) or knowledge sharing (see \cite{979961}) in multi-agent scenarios.  However, as we mentioned, in some applications, the task is new to all agents, and there is no expert agent present in the environment. In addition, the assumption of sharing or transferring knowledge is not feasible in many scenarios, e.g., when there are competing agents, agents have internal intrinsic goals, or agents are owned by different institutions. Most crucially, these presumptions cannot be applied to learning from humans \cite{jaques2019social}.

Several works in the literature assume a similar setting to ours. The authors investigated if independent RL agents in a multi-agent environment can leverage social learning to enhance performance \cite{ndousse2021emergent}. Generalized social learning policies emerged by adding a model-based auxiliary loss to model-free deep RL. These policies learn advanced skills and adapt quickly to new environments by learning from experts. Additionally, other authors have proposed a deep RL model to optimize the social learning strategies of agents in a cooperative game in a multi-dimensional landscape and demonstrated its superior performance in different settings \cite{ha2023social}. Our work differs from these approaches because they focus explicitly on generating emerging social RL policies or optimizing them, while we aim to develop an algorithm that inherently possesses this ability.

The other work  \cite{borsa2017observational} is an example of using observational learning to enhance an agent's performance. They employed a deep RL algorithm with memory, enabling agents to learn new tasks from rewards provided by the environment and, if available, from the observed behavior of expert agents acting as a teacher. Their assumption entailed expert agents having access to hidden information about the environment, whereas, in our method, we do not make such an assumption.

In another work, the authors compared four computational models of imitation in RL that are assumed to be used by humans through conducting a social RL task \cite{najar2020actions}. The results showed that the \textit{Value Shaping} method can represent imitation better than the other models and self-value guides the imitation rate. In contrast to our method, the authors designed their hypotheses for a two-armed bandit environment with binary rewards and just one demonstrator. The authors in another similar study introduced a socially correlated bandit task that accommodates payoff differences among participants and proposed a method called Social Generalization \cite{witt2024humans}. This method was tested through evolutionary simulations in comparison with Value Shaping and Decision Biasing methods. However, a comparison with our work was not feasible due to the difference in problem setting; we consider a stochastic bandit problem while they consider a spatially correlated bandit problem.

A recent study examined the performance of success-based social learning strategies and conformity strategies in different levels of environmental volatility and uncertainty  \cite{yaman2022meta}. They also developed a meta-control of individual and social learning strategies that affords agents the leverage to resolve environmental uncertainty with minimal exploration cost by exploiting others’ learning as an external knowledge base. However, unlike our setting, they assumed that we could see the rewards received by the individuals. In a similar vein, in a study on social bandit learning, the authors examined social learning dynamics and provided some theoretical analysis where agents could observe the full history of previous agents' decisions and rewards \cite{NEURIPS2023_212b143b}.

Similar to the last article, the well-known research demonstrates the effectiveness of social learning through a multi-armed bandit tournament \cite{rendell2010copy}. The key difference from our method lies in their agents' access to noisy rewards for observed behaviors and an accurate reward for a random behavior. Our approach takes broader assumptions for real-world applicability.

Our work is mostly related to \cite{lupu2019leveraging} and ~\cite{Zong2020SocialBL}, who proposed a social bandit learning algorithm inspired by the Upper Confidence Bound (UCB) learning method \cite{auer2002finite} to enhance agents' decisions by considering other agents' actions. Both methods are based on the optimism principle about the average of observed policies. The underlying assumption here is that the probability of selecting the optimal action by the average observed policy is higher than a uniform probability. However, unlike their studies, we assumed that IAs might be diverse,  irrelevant, or even misleading to the SA and aim to investigate the potential of social cues to speed up learning in a more generalized setting.

\section{ Problem Statement and Assumptions}
We examine the stochastic bandit problem, which is the most conventional $K$-armed bandit problem, where $\mathcal{A}$  represents the available actions for all agents with $\left | \mathcal{A} \right |$ as the total number of actions. This action set is the same for all the agents unless it is explicitly mentioned. Each action corresponds to an unknown reward distribution, $\mathcal{P}_a$ \cite{lattimore2020bandit}. Thus, we can define a bandit instance $\nu = \mathcal{P}_a: a \in \mathcal{A}$. In unstructured bandits, there exist sets of distributions $\mathcal{M}_a$ for each $a \in \mathcal{A}$. Thus, the environment class can be defined as follows: 

\begin{equation*}
\mathcal{E} =\left \{ \nu = (\mathcal{P}_a: a \in \mathcal{A}): \mathcal{P}_a \in \mathcal{M}_a \hspace{2pt} for \hspace{2pt} all \hspace{2pt} a \in \mathcal{A} \right \}.
\end{equation*}
Here, $\mathcal{M}_a$ represents a set of diverse distributions, all belonging to the same family of distributions but characterized by distinct sets of parameters.
There are $N$ agents that interact with the bandit environment in the problem, each selecting an action $a_t \in \mathcal{A}$ at each trial $t$ and observing reward $r_t \sim \mathcal{P}_a$. Then we can define $\mu_a$ as the average reward of action $a$ :
\begin{equation*}
\mu_a(\nu) = \int_{-\infty}^{\infty } r\hspace{2pt} d\mathcal{P}_a(r).
\end{equation*}
The social agent (SA), agent $\it{N}$,  is the only agent capable of observing the actions of other agents, called individual agents (IAs), in trial $t$, $a_{t, i} \hspace{0.2cm} i=1,2,..., N-1$, without receiving any supplementary information, like rewards. Every leaner agent aims to achieve its individual target. The objective of the SA  is to maximize its expected reward, while we do not have such assumptions necessarily for IAs. The expected reward is defined as: 

\begin{equation} \label{eq1}
\mathbb{E}[R_t] = \sum_{a = 1}^{K} \pi_{t, N}(a)\mu_a(\nu).
\end{equation}
The policy or the probability of selecting action $a$ by the $N$th agent, SA, is denoted as $\pi_{t, N}(a)$. To evaluate and compare the algorithms, we utilize pseudo-regret as a metric. The pseudo-regret over $T$ trials, which is a random variable with respect to the stochastic choices of arms, is:

\begin{equation}
   \mathcal{R}_T = T \times \mu^{*}(\nu)- \sum_{t = 1}^{T}R_{t},
\end{equation}
where  $\mu^{*}(\nu)= \max_{a \in \mathcal{A}} \ \mu_a(\nu)$, and $R_t$  is the reward signal received at trial $t$. Nonetheless, in the discussion that follows, we use the term regret to refer to pseudo-regret, and we use the average of pseudo-regret as an estimation of regret \cite{lupu2019leveraging}.

\section{Preliminaries}
In this section, we first present Thompson Sampling (TS) for policy estimation in bandit problems. We provide a brief outline of how to approximate  TS policy utilizing a sampling approach. Following this, we present an overview of implementing the free energy paradigm in sequential decision-making tasks, as well as its association with the TS algorithm. Finally, we elaborate on the individual learning method employed by our SA.

\subsection{Thompson Sampling Policy Estimation}
The Thompson Sampling algorithm, initially introduced in \cite{thompson1933likelihood}, is utilized in this study. While the original work had certain limitations, recent developments have resolved these issues and provided theoretical guarantees, demonstrating that the method often closely approximates the optimal solution.
 
TS is a technique used in bandit problems that selects each action based on the probability of its optimality \cite{russo2018tutorial}. It incorporates both its estimation of the expected value of actions and uncertainty in the estimation. This probabilistic measure can be determined using belief distributions, which represent the probability distribution of values under model uncertainty for each available action. In the bandit learning framework, belief distributions are modified iteratively during the learning process. The policy of TS can be expressed as:

\begin{equation} \label{pi_TS_integral_eq}
\begin{gathered}
\pi_{T S}(a_i)=P\left(\bigcap_{j \neq i}\left\{\hat \mu_{a_i}(\nu)> \hat \mu_{a_j}(\nu)\right\}\right)= 
\\
\int_{-\infty}^{\infty} p_{\hat \mu}\left(x \mid a_i\right) \prod_{j \neq i} P\left(x>\hat \mu_{a_j}(\nu)\right) d x= \\
\int_{-\infty}^{\infty} p_{\hat \mu}\left(x \mid a_i\right) \prod_{j \neq i} \int_{-\infty}^{x} p_{\hat \mu}\left(y \mid a_j\right) d y d x,
\end{gathered}
\end{equation}
where $\hat \mu_{a_i}(\nu)$ is the estimated value corresponding to action $a_i$, which equals the expected value of reward distribution of action $a_i$ in bandit instance $\nu$. In addition, $p_{\hat \mu}\left(. \mid a_i\right)$ is the belief distribution for the estimated value corresponding to action $a_i$. To determine the policy using TS, belief distributions must be computed. Bayes' rule can be implemented directly to compute belief distributions. Nevertheless, computing the above integrals for TS policy at each trial can be a time-consuming process. Hence, it is advantageous to approximate TS policy by repeating the sampling process sufficiently. Samples are drawn from all actions' belief distributions, and the counter of the action with the highest value sample is increased each time. Ultimately, the count of each action with the maximum value is averaged, providing an approximation of the TS policy.

\subsection{Free Energy Model of Decision Making}

In the field of statistical physics, it is widely recognized that the Boltzmann distribution satisfies a variational principle involving the free energy $F = U - TS$ minimization. This principle highlights the delicate balance between the internal energy $U$ and the entropic cost $S$ at temperature $T$  for systems at thermal equilibrium. Interestingly, this framework can also be utilized in both action and perception domains. As notable instances, it not only facilitates the derivation of various decision-making frameworks, such as expected utility theory \cite{von1944theory, savage1954foundations}, but also enables the formulation of a variational principle for (approximate) Bayesian inference. This principle has been proposed as a fundamental mechanism underlying self-organizing and learning systems \cite{friston2009free, friston2010free, ortega2013thermodynamics, isomura2023experimental}.

The free energy model, drawing upon this principle, is frequently utilized to explain the behavior of agents faced with limited computational resources (bounded rationality). This is particularly relevant when they are confronted with challenges such as choosing one action from an excessive number of options within a limited time frame \cite{gottwald2019bounded, ortega2011information, hihn2019information}, or when selecting from multiple experts where a swift decision is required even if the chosen expert might not be the optimum one. In this context, the optimization problem for identifying the constrained optimal policy in state $s$ can be expressed as finding the policy $\pi^*(a \mid s)$ that minimizes the free energy $F$:

\begin{equation}\label{Free_Energy_Model}
\begin{gathered}
\pi^*(a \mid s)=\underset{\pi(a \mid s)}{\arg \min } F(s ; \pi(a \mid s)), \\
F(s ; \pi(a \mid s))=\mathbb{E}_{\pi(a \mid s)}\left[\frac{1}{\alpha} \log \frac{\pi(a \mid s)}{p_0(a \mid s)}-U(a, s)\right].
\end{gathered}
\end{equation}

Here, $F(s ; \pi(a \mid s))$ denotes the free energy of an agent in state $s$ utilizing policy $\pi(a \mid s)$, relative to a reference prior policy $p_0(a \mid s)$ (also known as the negative free energy difference). $U(a, s)$ represents the utility (or negative cost) of performing action $a$ in state $s$. The inverse temperature $\alpha$ serves as a crucial trade-off parameter.

In essence, this framework uses a physics analogy to model decision-making under constraints (bounded rationality). Minimizing free energy involves balancing the goal of maximizing expected utility ($\mathbb{E}[U(a,s)]$) against an information cost associated with deviating from a prior policy $p_0(a|s)$. This cost, represented by the KL divergence term ($\mathbb{E}[\frac{1}{\alpha} \log (\pi/p_0)]$), penalizes complex or surprising policies, which can be interpreted as the physical work needed for changing the state of the system. The parameter $\alpha$ controls this trade-off: higher $\alpha$ emphasizes utility maximization, while lower $\alpha$ enforces sticking closer to the prior policy $p_0$, reflecting stricter constraints or more uncertainty. This provides a principled way to model agents achieving good, but not necessarily optimal, outcomes due to their inherent physical limitations. This also serves as an effective tool for managing the exploration-exploitation dilemma for learning agents.

Furthermore, it can be demonstrated that for decision-making problems involving a single state, the Thompson Sampling (TS) policy emerges as the optimal decision-making strategy under this free energy model for a particular setting of the parameter $\alpha$ \cite{ortega2014generalized}.

\subsection{Individual learning method}

Our social learning method contains an individual learning algorithm. However, our social learning method is designed to be flexible and function independently of the individual bandit strategies. Thus, we do not impose any restrictions on the individual learning methods used by either SA or IAs. Therefore, any bandit method can be employed inside our SA. However, here we use  $\epsilon$-greedy \cite{SuttonRL}, UCB \cite{auer2002finite}, and TS \cite{russo2018tutorial} as individual learning methods due to their inherent benefits in the RL domain.

\section{Proposed Method}

\begin{figure}
\centerline{\includegraphics[width = 7.5cm]{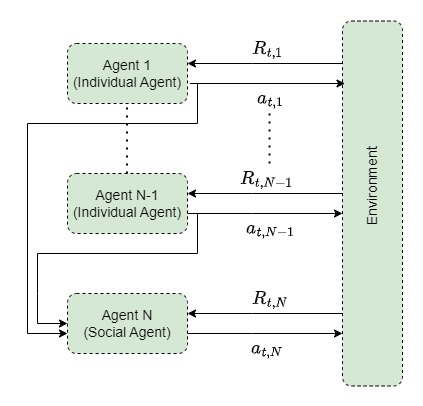}}
\caption{Social bandit learning problem setting}
\label{fig:probsetting}
\end{figure}

We propose a method for applying social learning by an SA to solve stochastic bandit problems. Our SA solves the problem by combining the information it acquires through direct interaction with the environment and passive observation of the behavior, i.e., actions only, of the other agents (IAs) interacting with the same environment;  see Fig. \ref{fig:probsetting}. 

As the agents, each IA and the SA, commonly have different goals and expertise levels, so the SA must evaluate itself and IAs in terms of suitability for solving the task. This should be done based on self-referenced evaluation by the SA, as there is no oracle or external evaluation system. However, due to the inherent, natural, and varying uncertainty in its knowledge during the learning process, the SA must incorporate this uncertainty in the evaluation strategy. We should remember that self-referenced evaluation, in general, can be inaccurate in the early stages of social learning. Because, at first, the SA does not know much about the task and this lack of knowledge can potentially cause considerable regret at the early trials. Therefore, using a combination of a proper global and task-independent measure, such as entropy, and a self-referenced evaluation measure by SA is crucial for information evaluation in effective social learning. 

\begin{figure}
\centerline{\includegraphics[width =9.5cm]{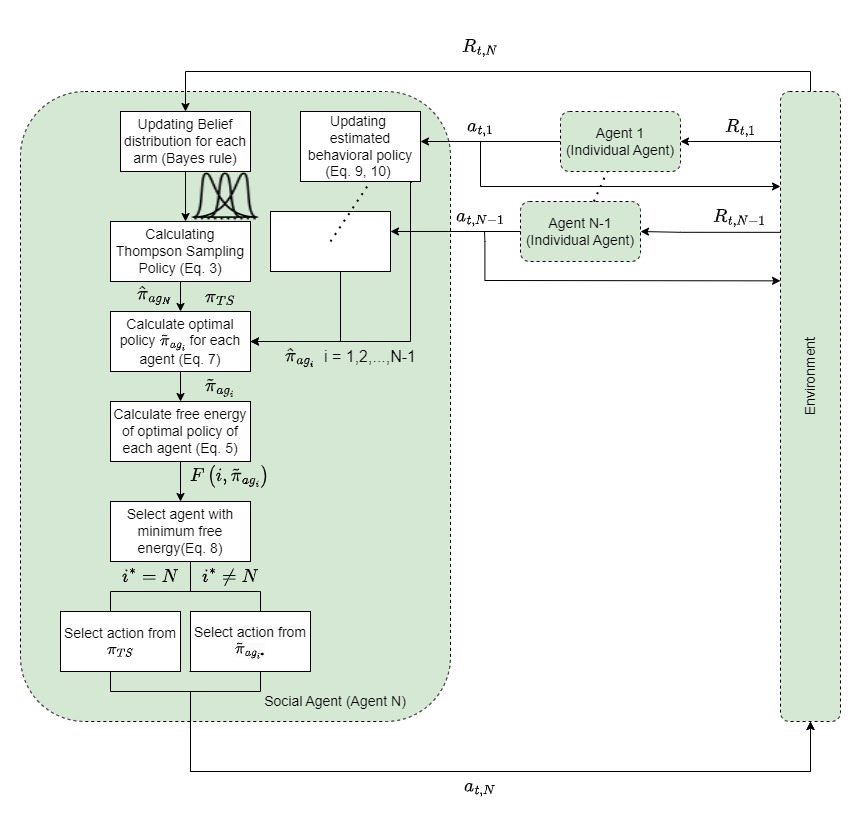}}
\caption{Informational flow diagram of the proposed method (SBL-FE)}
\label{fig:infoflow}
\end{figure} 
 
As we discussed above, the SA evaluates IAs and itself in a self-referenced manner. That is, the SA's utility forms the core of the evaluation. Nevertheless, the SA's estimation of its utility involves uncertainty. In addition, the SA observes the policy of the IAs and cannot estimate their utilities.  Therefore, our SA forms its evaluation measure in the policy space. This facilitates a common framework for integrating the SA's information regarding its uncertain expected utility estimation and the IA's experience.
Interestingly, TS policy incorporates both estimated utilities and their uncertainties. In other words, a signature of both the utility and experience of the SA are encoded in its TS policy. The same is true for the estimated policy of IAs from a rationalist point of view. In addition, the global measure of certainty, i.e., entropy, is well-defined in the policy space.  Evaluation in the policy space facilitates exploration-exploration balance for the SA as well.  

Inspired by the work of \cite{ghorbani2025learning}, our SA uses the free energy model in the policy space to identify the best behavior policy based on its uncertain experiences and observed behavior of IAs. In the first step, the SA finds a set of policies that minimize its own and each IAs' free energy:


\begin{equation} \label{Free_Energy_agents_eq}
\begin{gathered}
F\left(i, \pi\right)= c D_{K L}\left(\pi \| \pi_{TS}\right)+H\left(\pi\right)+D_{K L}\left(\pi \| \hat{\pi}_{{ag}_i}\right)   \\
\hspace{.5cm} = \mathbb{E}_{\pi(a)}\left[c \log \frac{\pi(a)}{\pi_{TS}(a)}-\log \hat{\pi}_{{ag}_i}(a)\right], \\ 
\end{gathered}
\end{equation}

\begin{equation} \label{free_energy_candidate_policy}
    \begin{gathered}
         \tilde{\pi}_{{ag}_i}=\arg \min _{\pi} F\left(i, \pi\right), \ \ i = 1,2,.., N-1,N.
    \end{gathered}
\end{equation}
 In this measure, the free energy of the $i^{th}$ agent for a given policy $\pi$   is evaluated.  The first term in \ref{Free_Energy_agents_eq} encodes SA's self-referenced evaluation, i.e. similarity of the given policy, $\pi$, and SA's expertise embedded in its TS policy, $\pi_{TS}$, where $c$ is a constant. The second term, $H(\pi)$, indicates the entropy of the given policy, i.e., the global measure of randomness, as we know the optimal policy is greedy \cite{SuttonRL}.  The last term measures the similarity of  $\hat \pi_{ag_{i}}$, the estimated policy of the target agent, $i^{th}$ agent, and the given policy $\pi$. The policy that minimizes the free energy of $i^{th}$ agent from the perspective of SA is $\tilde{\pi}_{{ag}_i}$, which can be considered as the candidate behavior policy suggested by the $i^{th}$ agent. This policy has the best balance of similarity with the SA and $i^{th}$ agent policies while having the minimum possible randomness. 

In our formulation \ref{Free_Energy_agents_eq}, for injecting  SA's knowledge, its TS policy, $\pi_{TS}$, is used as  $p_0,$ which is a reference policy in \eqref{Free_Energy_Model}. With the assumption of rationality of IAs, we use the projection of their utility in the policy space and substitute the utility term in \eqref{Free_Energy_Model} by $\log \hat{\pi}_{{ag}_i}$. In other words, the utility term can be considered as a negative informational surprise of SA's estimation of the behavior policy of the agent as a global suitability measure of other agents. Therefore, when we get the expected value of the utility with respect to an arbitrary policy, $\pi(a)$, it can be stated as the cross entropy between the estimated behavior policy of the agent and $\pi(a)$. The cross-entropy is equivalent to the entropy of the policy of $\pi(a)$ plus the KL-divergence between the estimated behavior policy of the agent and $\pi(a)$. Thus the best policy $\pi(a)$ should be close to the estimated behavior policy and have small entropy. 

Then, according to \cite{ortega2013thermodynamics} we have the below solution for \eqref{free_energy_candidate_policy}: 

\begin{equation} \label{pi_Star_eq}
\begin{aligned}
\tilde{\pi}_{{ag}_i}(a) & =\frac{1}{Z(i)} \pi_{TS}(a) e^{\frac{1}{c} \log \hat{\pi}_{{ag}_i}(a)}, \\
Z(i) & =\sum_a \pi_{TS}(a) e^{\frac{1}{c} \log \hat{\pi}_{{ag}_i}(a)}.
\end{aligned}
\end{equation}
Finally, the SA follows $\tilde{\pi}_{{ag}_i}$ that has the minimum free energy over all IAs and itself as its behavior policy: 

\begin{equation} \label{agent_opt_eq}
i^*=\underset{i}{\arg \min } \ F\left(i, \tilde{\pi}_{{ag}_i}\right), 
\pi_{{SA}_{beh}}=\begin{cases}
\pi_{TS} & \text{ if } i^{*} = N \\ 
\tilde{\pi}_{{ag}_i} & \text{ otherwise } 
\end{cases}.
\end{equation}
 Since SA updates its knowledge, and consequently its TS policy, this behavior results in an exploration-exploitation balance. Fig. \ref{fig:infoflow} shows the components of our proposed method, which we refer to as Social Bandit based on the Free Energy paradigm where we use SBL-FE as a short hand. Algorithm \ref{sociallearneralgo} illustrates our proposed method for solving stochastic bandit problems under the social learning setting.
  
\begin{algorithm}[H]
\caption{Social Bandit Learning Algorithm(SBL-FE)}
\label{sociallearneralgo}
\begin{algorithmic} 
\STATE \textbf{Input:} $K$, $N$, Environment Class, $c$, $\lambda$, $\epsilon$
\STATE Initialize $N_{ag_{i}}^{0}(a) = 1$,  $\pi_{TS}$ hyperparameters based on environment class
\STATE Calculate the candidate policy, $\tilde{\pi}_{{ag}_i}$, for each agent and its free energy \COMMENT{Eq. \ref{Free_Energy_agents_eq}, \ref{free_energy_candidate_policy}, \ref{pi_Star_eq}}
\FOR{each trial $t$}
\STATE Sample $A_t \sim \pi_{{SA}_{beh}}$ and observe reward $R_t$, and other agents' actions
\COMMENT{Eq. \ref{agent_opt_eq}}
\STATE Update $N_{ag_{i}}^{t}$ based on observations, and estimated behavior policy \COMMENT{Eq. \ref{update_observation_eq}, \ref{est_beh_policy_eq}}
\STATE Update belief distributions using Bayes rule, and TS policy \COMMENT{Eq. \ref{pi_TS_integral_eq}}
\STATE Update free energy of all agents based on updated $\tilde{\pi}_{{ag}_i}$ \COMMENT{Eq. \ref{Free_Energy_agents_eq}, \ref{free_energy_candidate_policy}, \ref{pi_Star_eq}}
\ENDFOR
\end{algorithmic}
\end{algorithm}

\subsection{Policy estimation}
The SA should estimate IA's policy  ($\hat{\pi}_{{ag}_i}$) based on observing their actions over trials. Policies are not stationary during the learning process. Thus, the SA should use an estimation method that can capture this volatility. Therefore, we consider the exponential moving average (EMA) over the observed action of the agents to estimate their behavior policy at trial $t$ as the following:

\begin{equation} \label{update_observation_eq}
    N_{ag_{i}}^{t} = (1 - \lambda) N_{ag_{i}}^{t-1} + \lambda e_i(t), 
\end{equation}
Here, $\lambda$ is a constant step size, and $e_i(t)$ is a vector where all elements are zero except the $i^{th}$ place, which is equal to one after observing that the agent takes the $i^{th}$ action. Since SA does not have any knowledge about the observed agent at the beginning, it initializes $N_{ag_{i}}^{0}$ equal to one vector for all agents in the environment,t including itself. After that, SA estimates the behavior policy for each agent as follows:
\begin{equation} \label{est_beh_policy_eq}
\hat{\pi}_{a g_i}^{t}(a)=\frac{{N}_{a g_i}^{t}(a)}{\sum_{a}  {N}_{a g_i}^{t}(a)}.
\end{equation}

For the SA, we employ the TS policy to determine its behavior. However, it should be noted that the estimated policy of the greedy agents within the society tends to converge rapidly towards a specific action, which may not necessarily align with the optimal action from SA's perspective.  To address the issue of overly sharp policies, we used a smoothing technique by linearly combining them with the uniform policy, using appropriate coefficients. 
 
\subsection{Proof of convergence}
 In Theorem 1, we investigate the conditions for algorithm convergence.

\textbf{Theorem 1.} \textit{Let $\nu$ be a stochastic bandit problem with a finite action space $\mathcal{A}$. In the proposed scheme, if a SA uses $0 < c < 1$ and we add a small constant to the TS policy and estimated behavior policies due to numerical implications, then SA finally uses its policy, $\tilde{\pi}_{{ag}_N}$, or an expert policy which is equivalent to the optimal policy. Therefore, it guarantees convergence to the optimal policy.} 

The detailed proof for Theorem 1 can be found in the supplementary section A.

\subsection{Time and Space Complexity Analysis}
Finally, we analyze the computational complexity of our algorithm in terms of time and space requirements. The time complexity of the algorithm is determined by the execution of sampling action, $A_t \sim \pi_{SA_{beh}}$, and updating belief distributions, TS policy, and free energy of all agents, each contributing to the overall runtime. The most computationally expensive operation is the updating process, which dominates the sampling, leading to a time complexity of $O(NK)$ for each trial $t$. SBL-FE also requires memory to store $N^t_{ag_i}$ $TS$ parameters. Behavior policies, belief distributions, and free energies have the same space complexity as $N^t_{ag_i}$. Therefore, the total space complexity is $O(NK)$.
\section{Experimental Results}
In order to assess the performance of our proposed social learning method in comparison to individual learning methods, we design different experiments with distinct scenarios. Initially, we examine the algorithm's potential within a society comprising agents with significant variations in abilities and goals. Subsequently, we evaluate the effectiveness of our proposed methods in identifying proper agents to learn from. We also investigate the impact of society's population sizes and problem difficulties (such as the number of arms). Finally, we assess the robustness of our algorithm against observation noise. In each section, we conduct comparative analyses between our method and the existing similar approaches.

Across all of our scenarios, the rewards are sampled from a Bernoulli distribution, where the expected reward of the optimal arm is set to 0.9. Unless explicitly mentioned, our results are based on an average of over 500 independent runs, each consisting of 2,000 trials. The reported results represent the average outcome, and in all figures, shaded areas indicate two standard deviations around the mean.


We compared our SA's performance with two alternative social learning methods that have the same assumption as our method, OUCB \cite{Zong2020SocialBL} and TUCB \cite{lupu2019leveraging}. Other agents in society are categorized as individual learners or non-learners within the environment in different scenarios. As individual learners, we are using different algorithms, including TS, UCB, and epsilon-greedy (with and without epsilon decay). Additionally, we introduce five types of non-learner agents: optimal agent, sub-optimal agent, random agent, opponent agent, and P-optimal agent. The optimal agent consistently selects the best action. The sub-optimal agent consistently selects the second-best action. The random agent chooses actions randomly for each trial, while the opponent agent selects the action with the lowest expected reward from our agent's perspective. Lastly, the P-optimal agent selects the optimal action with a probability of P and otherwise acts randomly. The value of P can remain fixed or vary throughout the trials.

We examine three bandit instances from the Bernoulli environment class, each representing different levels of uncertainty. We consider three distinct optimality gaps between the optimal arm and the second-best one; i.e., $\Delta = 0.05$, $\Delta = 0.1$, and $\Delta = 0.2$. The details of the reward distributions can be found in the supplementary section B. Prior to conducting experiments with social algorithms, we evaluate the performance of TS, UCB, and epsilon-greedy individual learning methods across these three tasks to choose the most competitive one. Unreported results show that TS consistently outperforms the other algorithms in all cases, which is consistent with the analysis reported in \cite{lattimore2020bandit}.

\subsection{The ability of social learning methods in different societies} \label{sec:ability}
To assess the performance of our social learning algorithm in comparison to alternative methods, we created various scenarios involving learning from non-learners or different types of individual learners. In this section, we compared the performance of our method, SBL-FE, with TUCB, OUCB (as social learning algorithms), TS, and UCB (as individual learning methods), using the cumulative regret criteria. We used the same hyperparameters for OUCB and TUCB as stated in their respective papers, and for all subsequent results, we employed the same hyperparameter set. Further information about these hyperparameters can be found in the supplementary section C.

\subsubsection{Learning from non-learners}
In this section, we investigate societies consisting of one social learner and one non-learner. The non-learner can belong to one of four types: optimal agent, sub-optimal agent, random agent, or opponent agent.

\begin{figure}[htbp]
  \centering

  \begin{subfigure}[t]{0.48\textwidth} 
    \centering
    \includegraphics[width=0.49\linewidth]{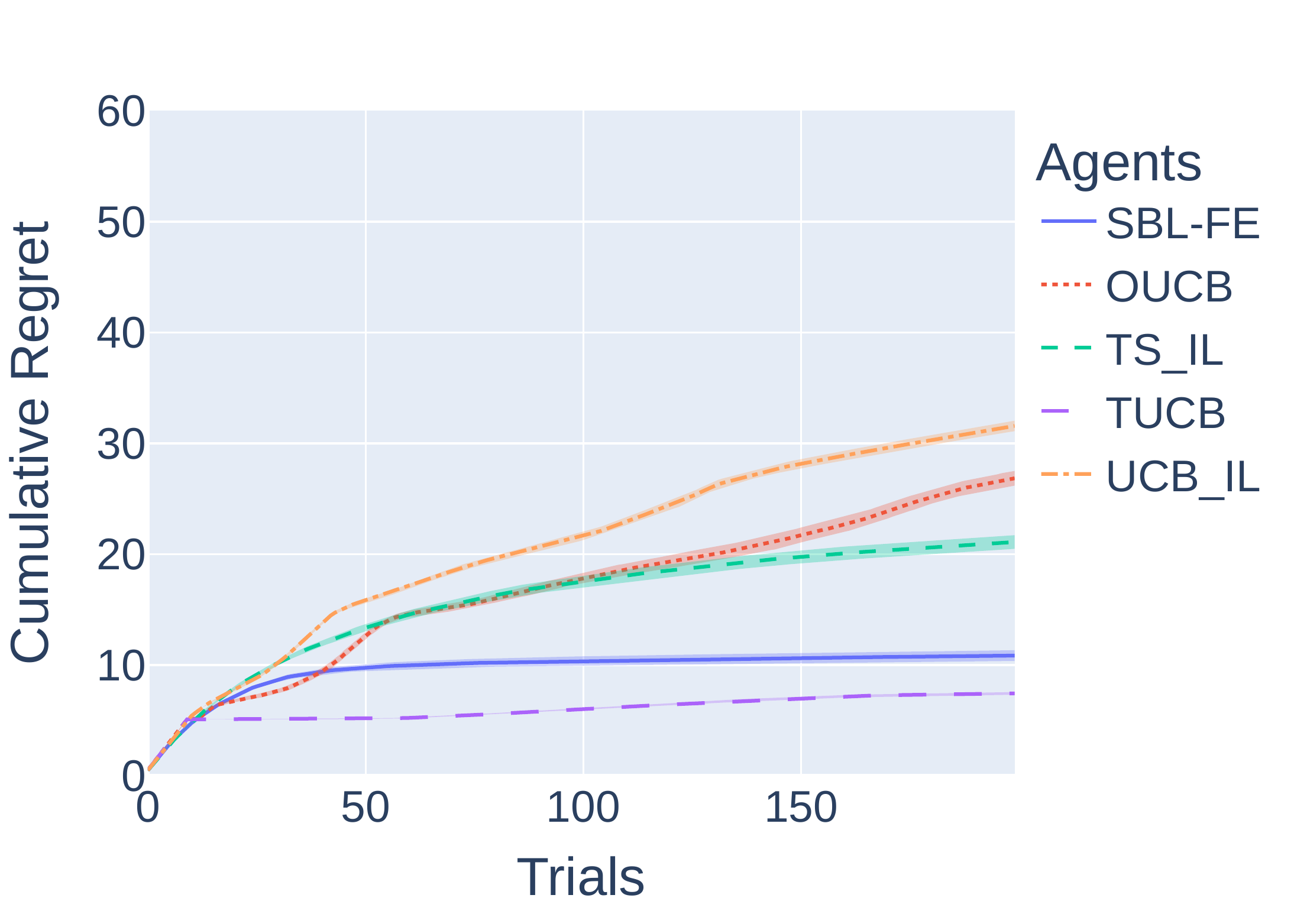}%
    \hfill
    \includegraphics[width=0.49\linewidth]{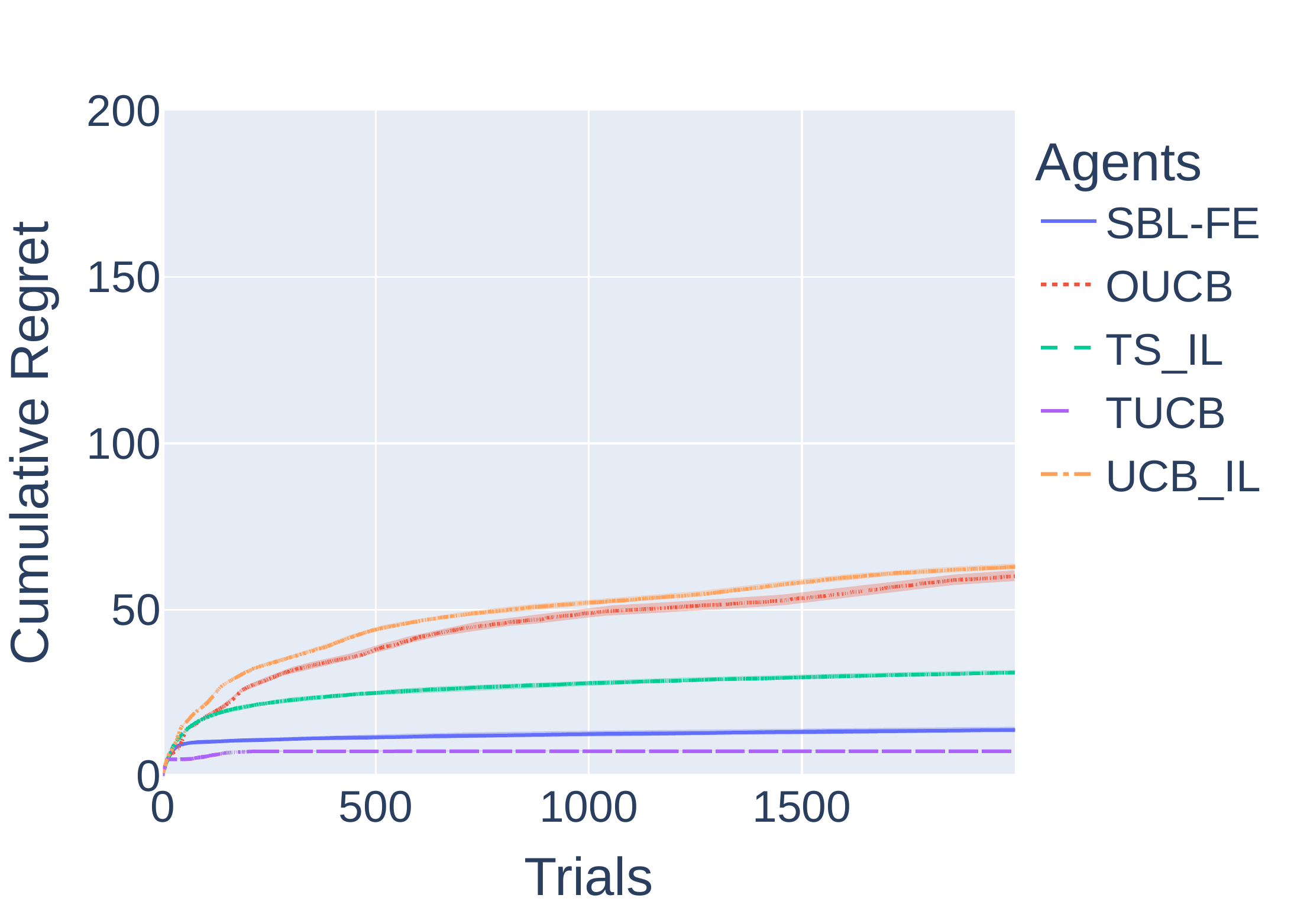}
    \subcaption{Optimal agent}
    \label{fig:nonlearners-a}
  \end{subfigure}
  \hfill
  \begin{subfigure}[t]{0.48\textwidth}
    \centering
    \includegraphics[width=0.49\linewidth]{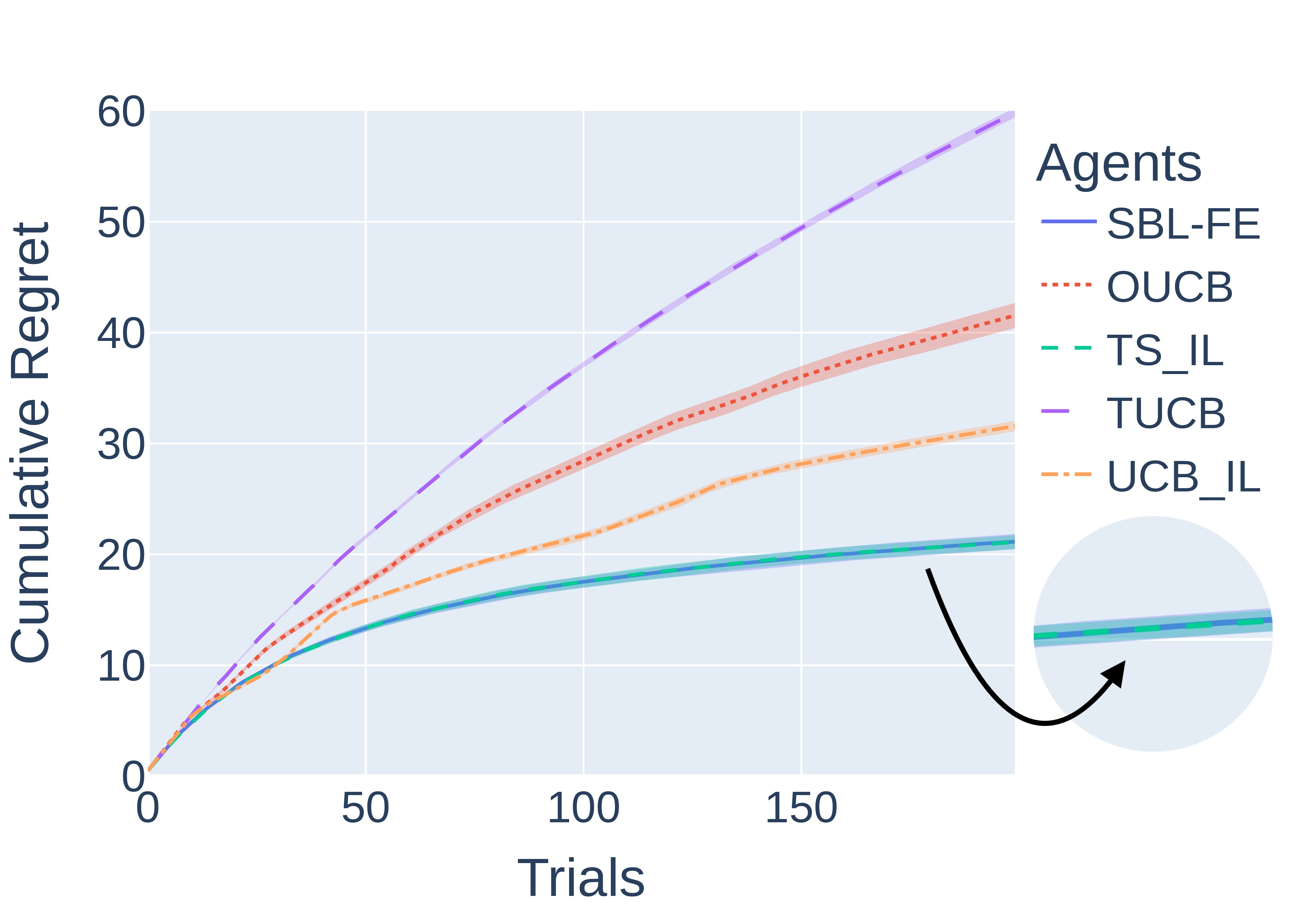}%
    \hfill
    \includegraphics[width=0.49\linewidth]{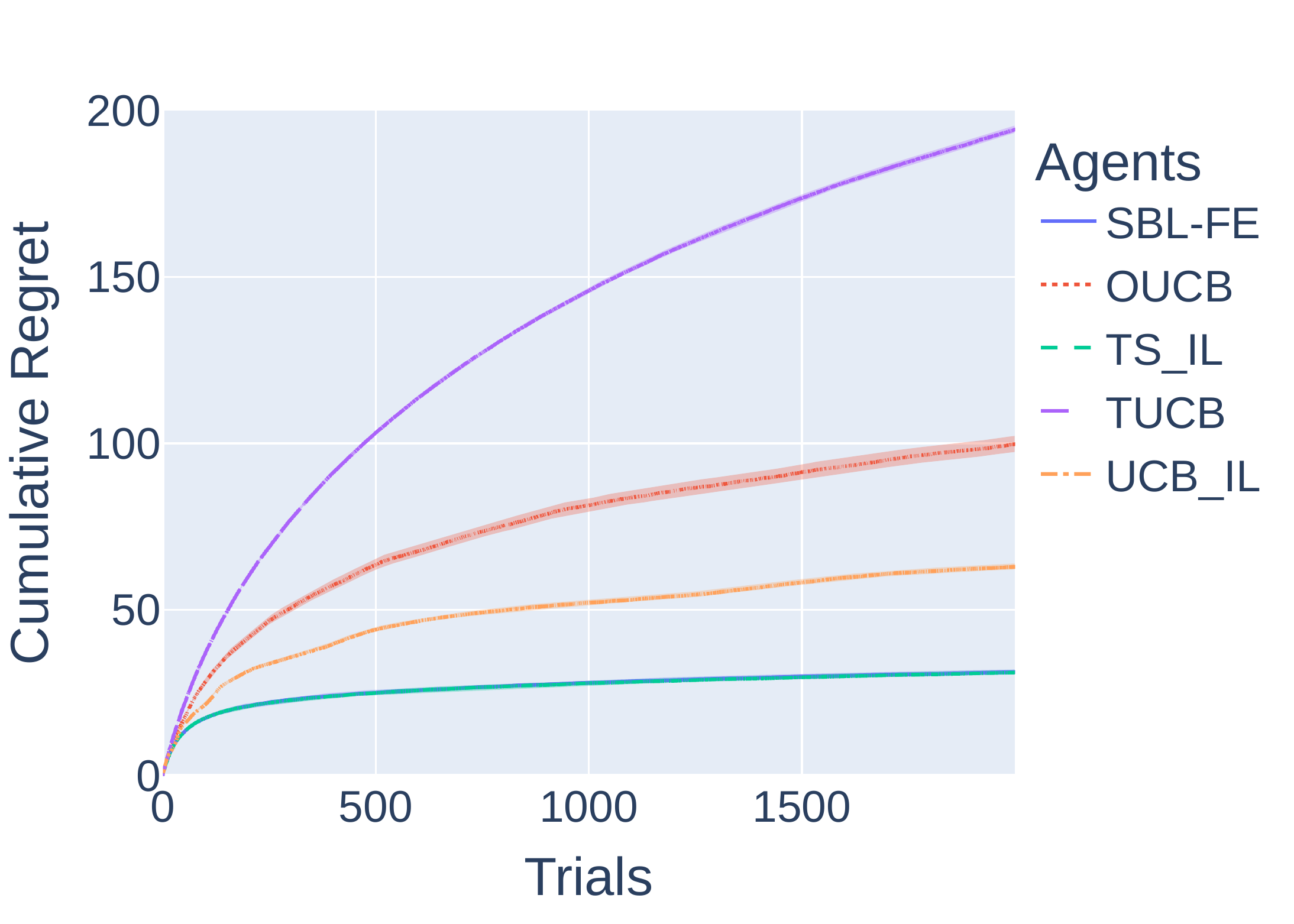}
    \subcaption{Random agent}
    \label{fig:nonlearners-b}
  \end{subfigure}

  \vspace{0.8em} 

  \begin{subfigure}[t]{0.48\textwidth}
    \centering
    \includegraphics[width=0.49\linewidth]{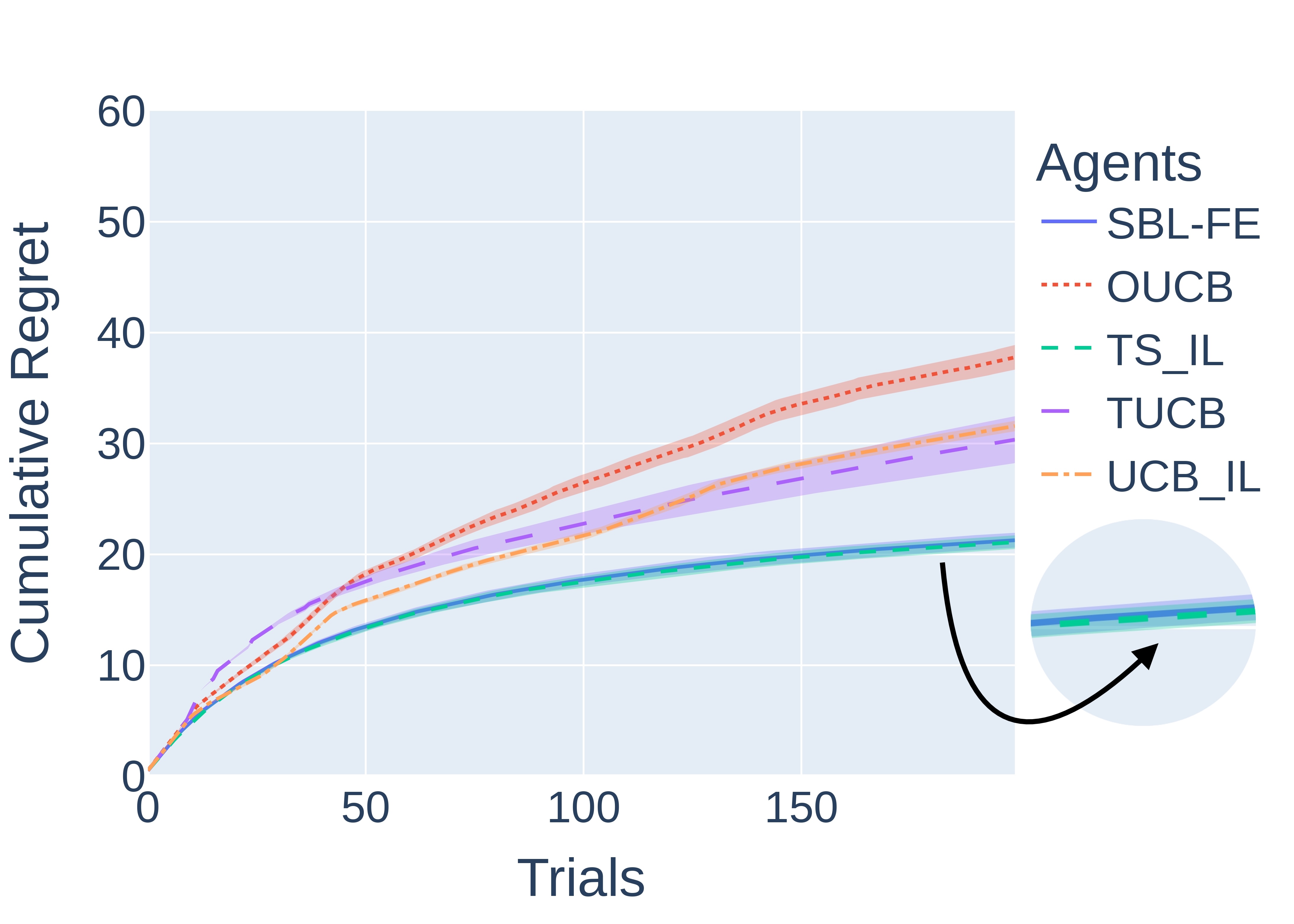}%
    \hfill
    \includegraphics[width=0.49\linewidth]{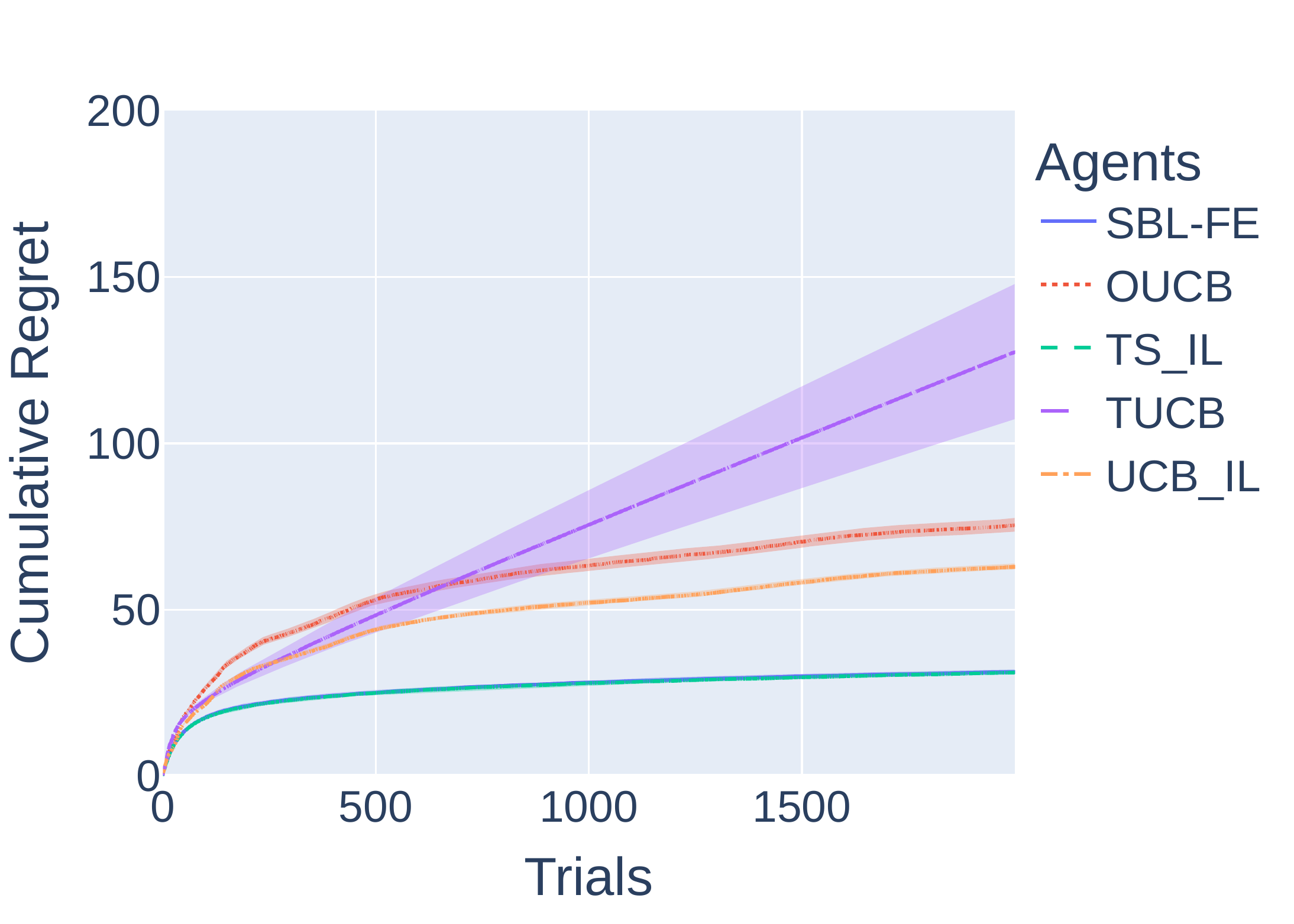}
    \subcaption{Opponent agent}
    \label{fig:nonlearners-c}
  \end{subfigure}
  \hfill
  \begin{subfigure}[t]{0.48\textwidth}
    \centering
    \includegraphics[width=0.49\linewidth]{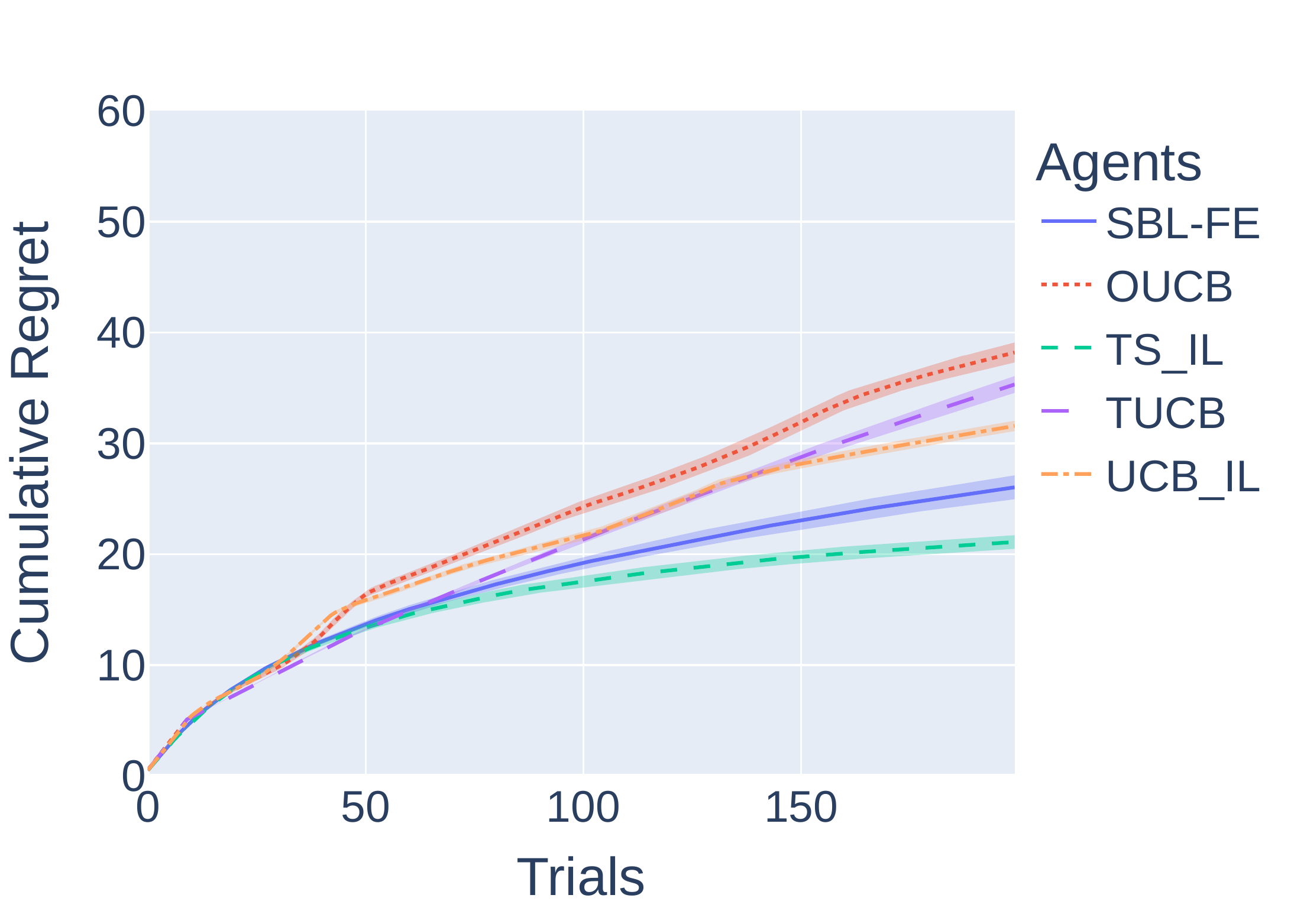}%
    \hfill
    \includegraphics[width=0.49\linewidth]{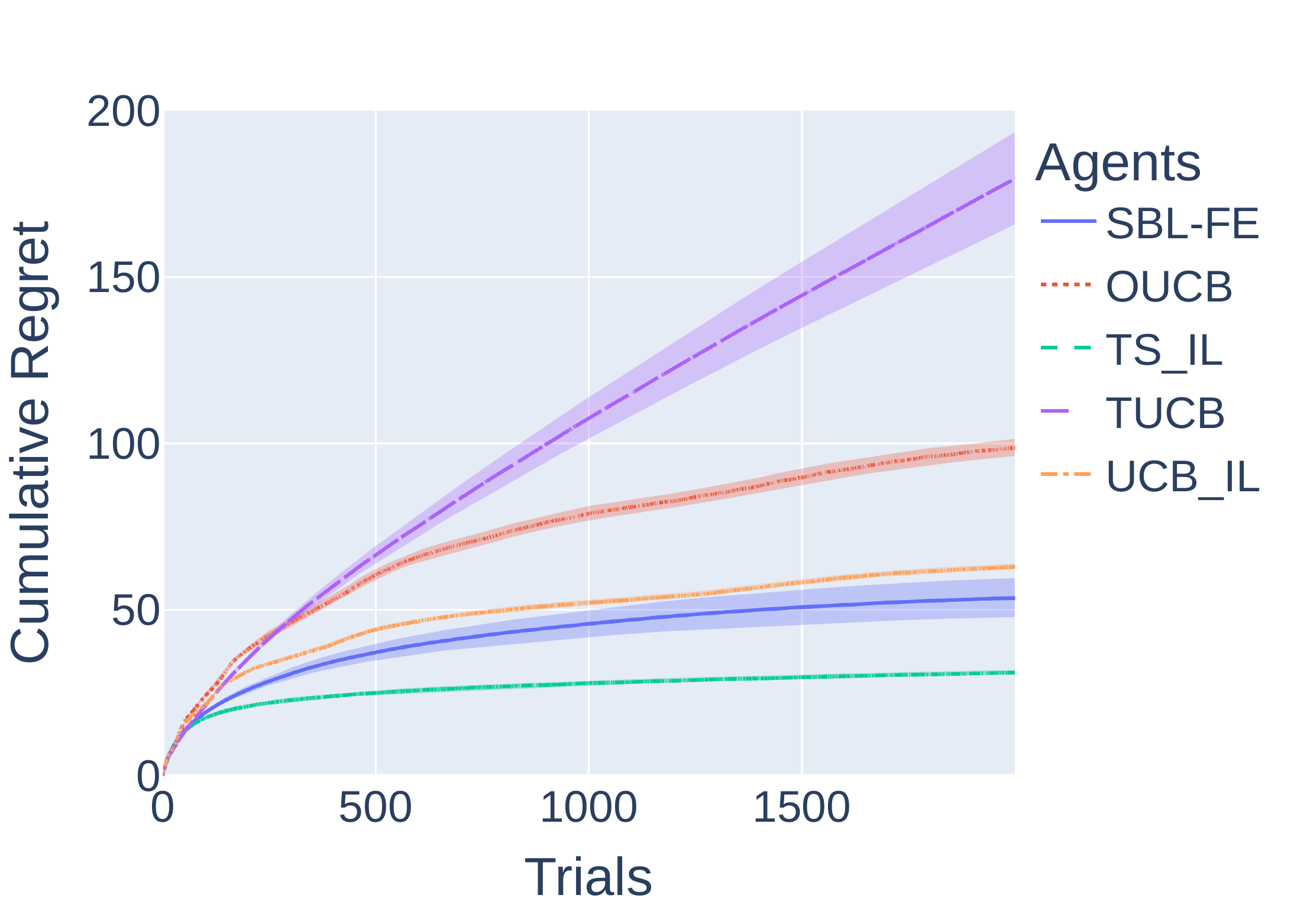}
    \subcaption{Sub-optimal agent}
    \label{fig:nonlearners-d}
  \end{subfigure}

  \caption{The cumulative regret performance of three social learning agents (OUCB, TUCB, SBL-FE) along with UCB and TS as baseline methods in societies consisting of one social learner and one non-learner. The experiments were conducted over 200 and 2000 trials for a 10-armed Bernoulli bandit problem with an optimality gap of $\Delta = 0.2$. In the zoomed-in view, we highlight that the performance of the TS method and our method are similar in some scenarios.}
  \label{fig:non-learners}
\end{figure}

Fig. \ref{fig:non-learners} compares the regret performance of these four types of societies in a Bernoulli task with $\Delta = 0.2$. The results from this figure demonstrate the excellent performance of our method under these settings. When there are no competent agents in society, our social learning method swiftly detects this and automatically switches to the TS method. In contrast, TUCB algorithm performs poorly when a relevant agent, one that consistently selects the optimal arm more than other arms, is absent in the society.

We replicated similar findings across other Bernoulli tasks with different optimality gaps. It was also observed that, in scenarios where an optimal agent exists within the society, TUCB algorithm tends to outperform our proposed methods. However, by adjusting the hyperparameters of our method, we can achieve results that are comparable to or even as good as those obtained by TUCB algorithm.

\subsubsection{Learning from learners}
\begin{figure}[htbp]
    \centering

    \begin{subfigure}[t]{0.48\textwidth}
        \centering
        \includegraphics[width=0.48\linewidth]{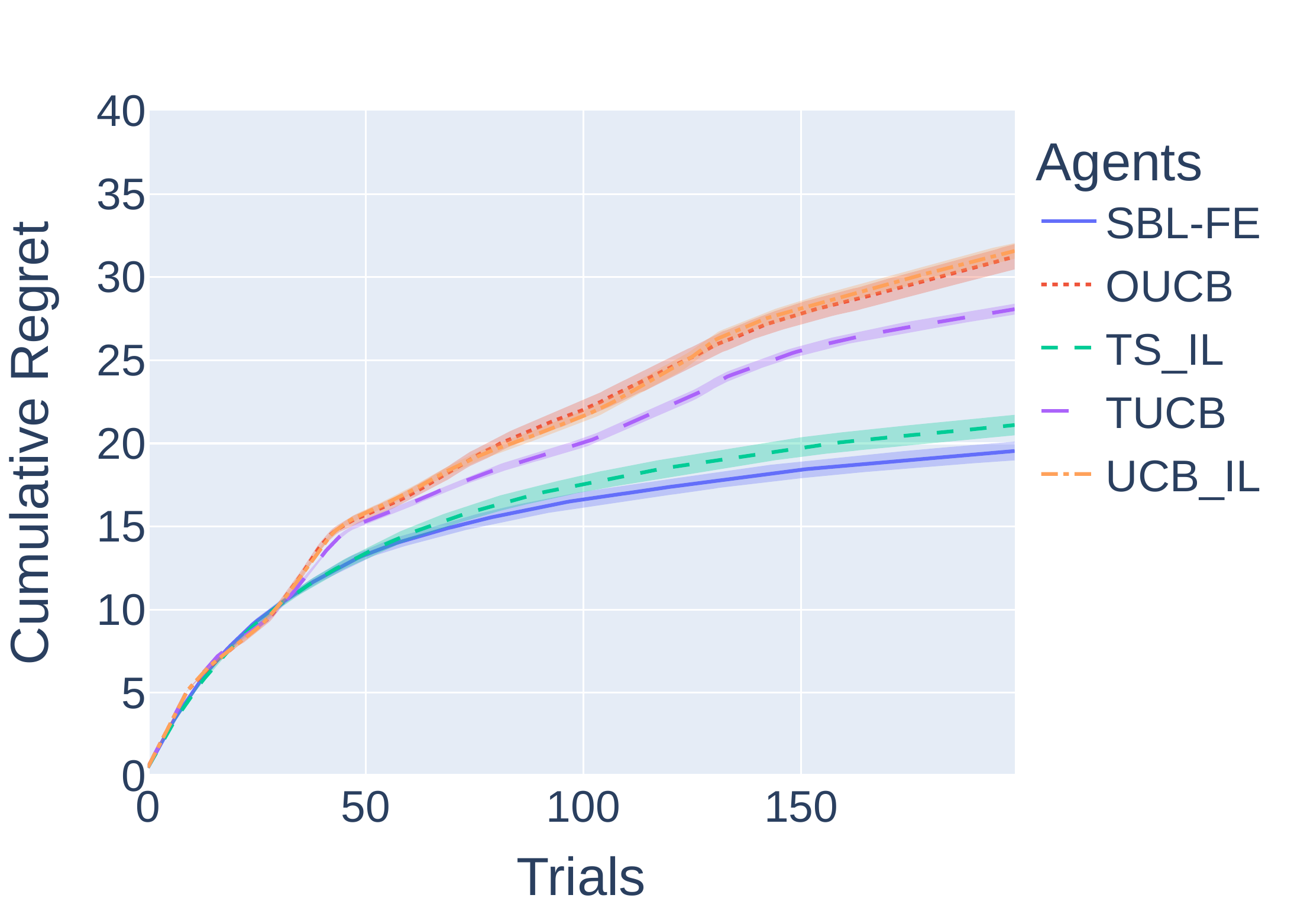}
        \includegraphics[width=0.48\linewidth]{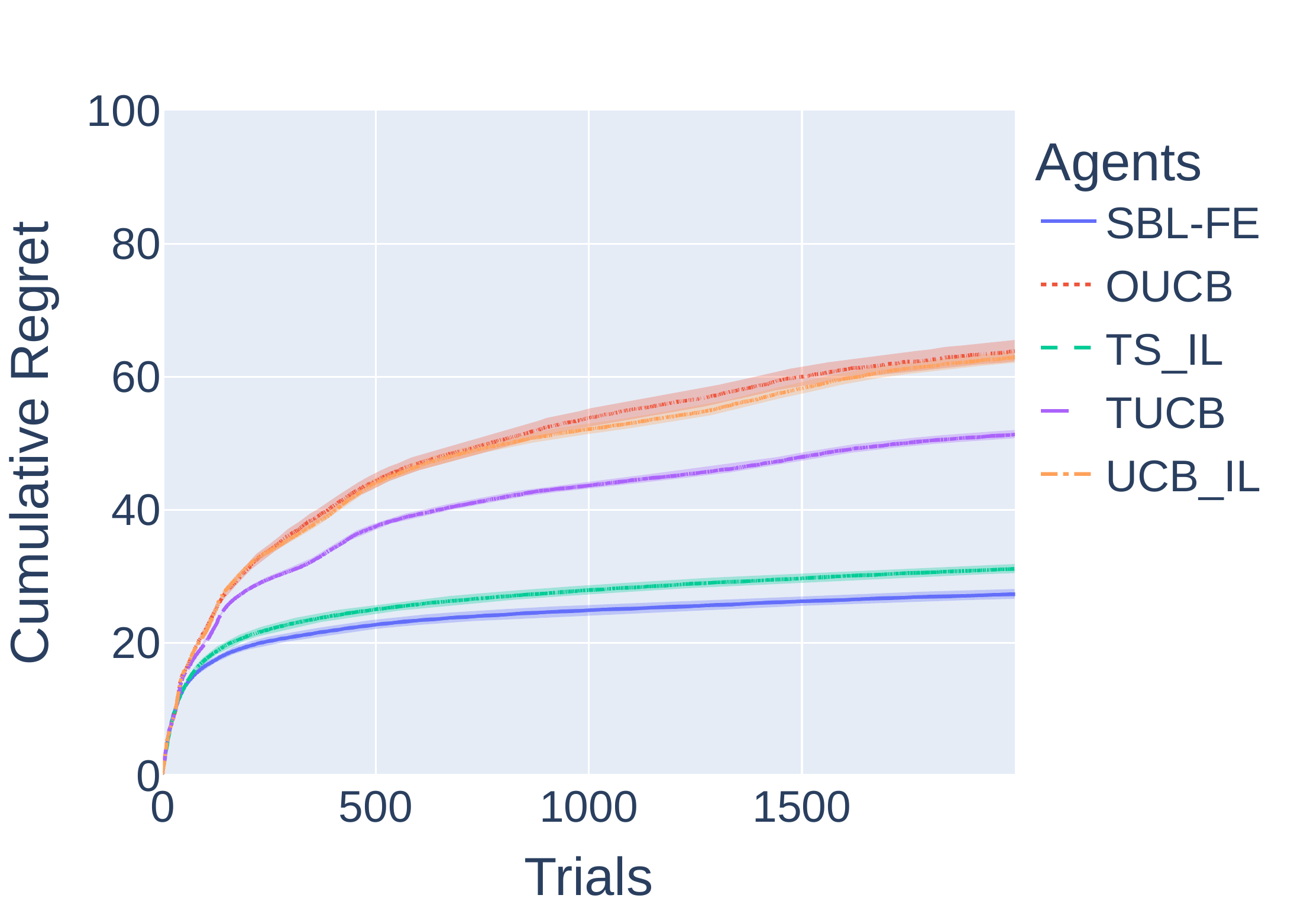}
        \subcaption*{(a) Thompson Sampling}
    \end{subfigure}
    \vspace{0.2cm}

    \begin{subfigure}[t]{0.48\textwidth}
        \centering
        \includegraphics[width=0.48\linewidth]{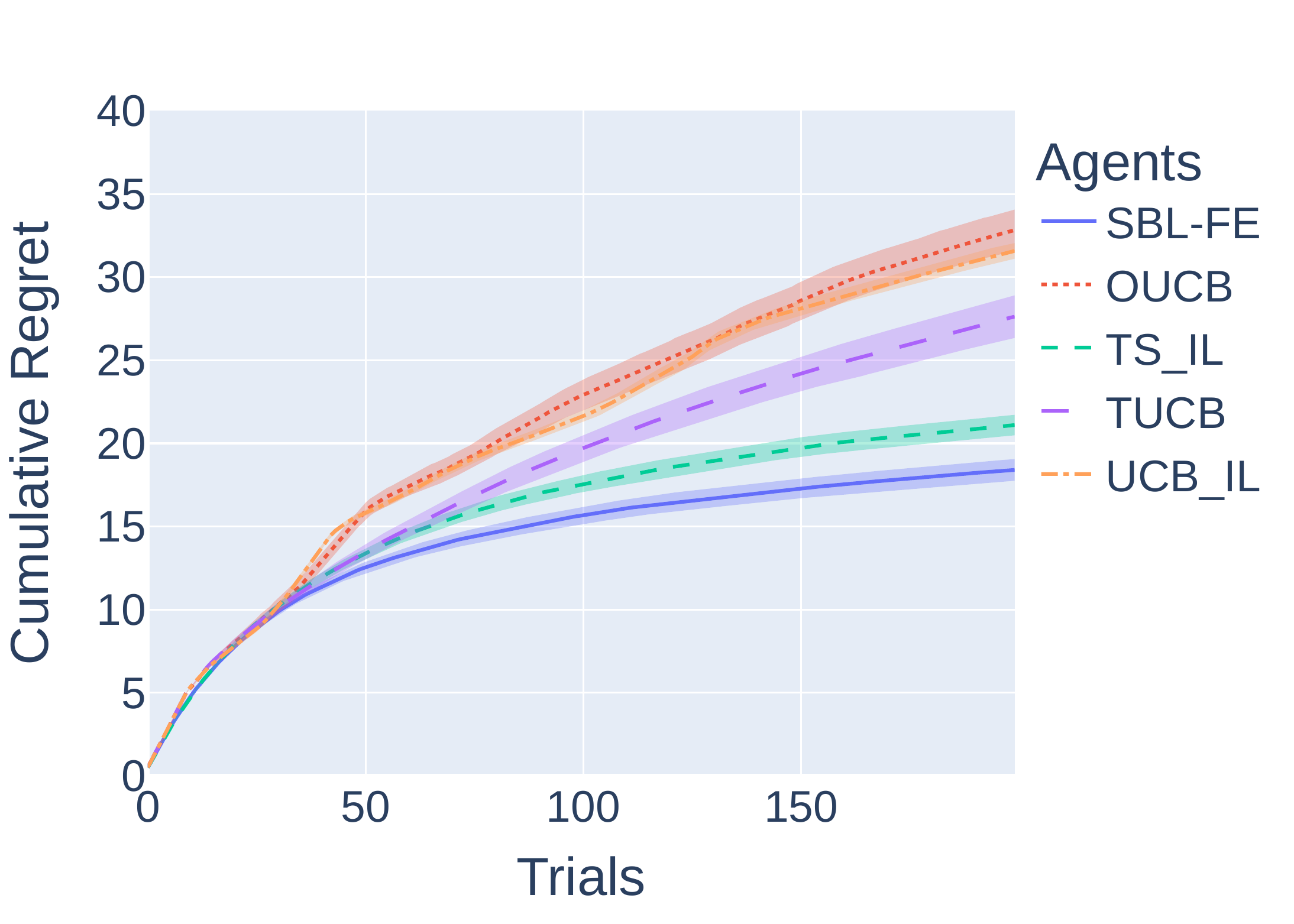}
        \includegraphics[width=0.48\linewidth]{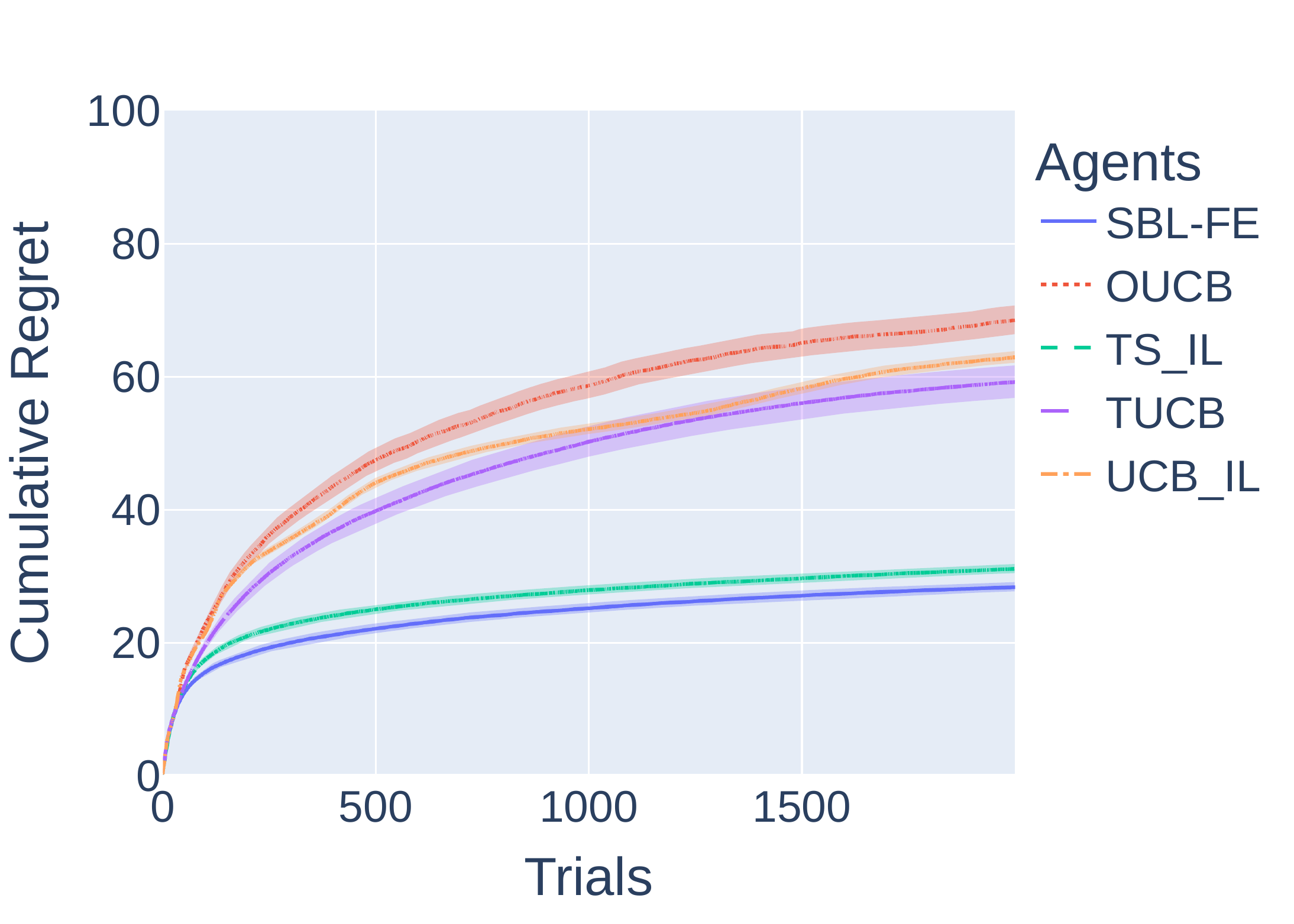}
        \subcaption*{(b) Epsilon-greedy}
    \end{subfigure}
    \vspace{0.2cm}

    \begin{subfigure}[t]{0.48\textwidth}
        \centering
        \includegraphics[width=0.48\linewidth]{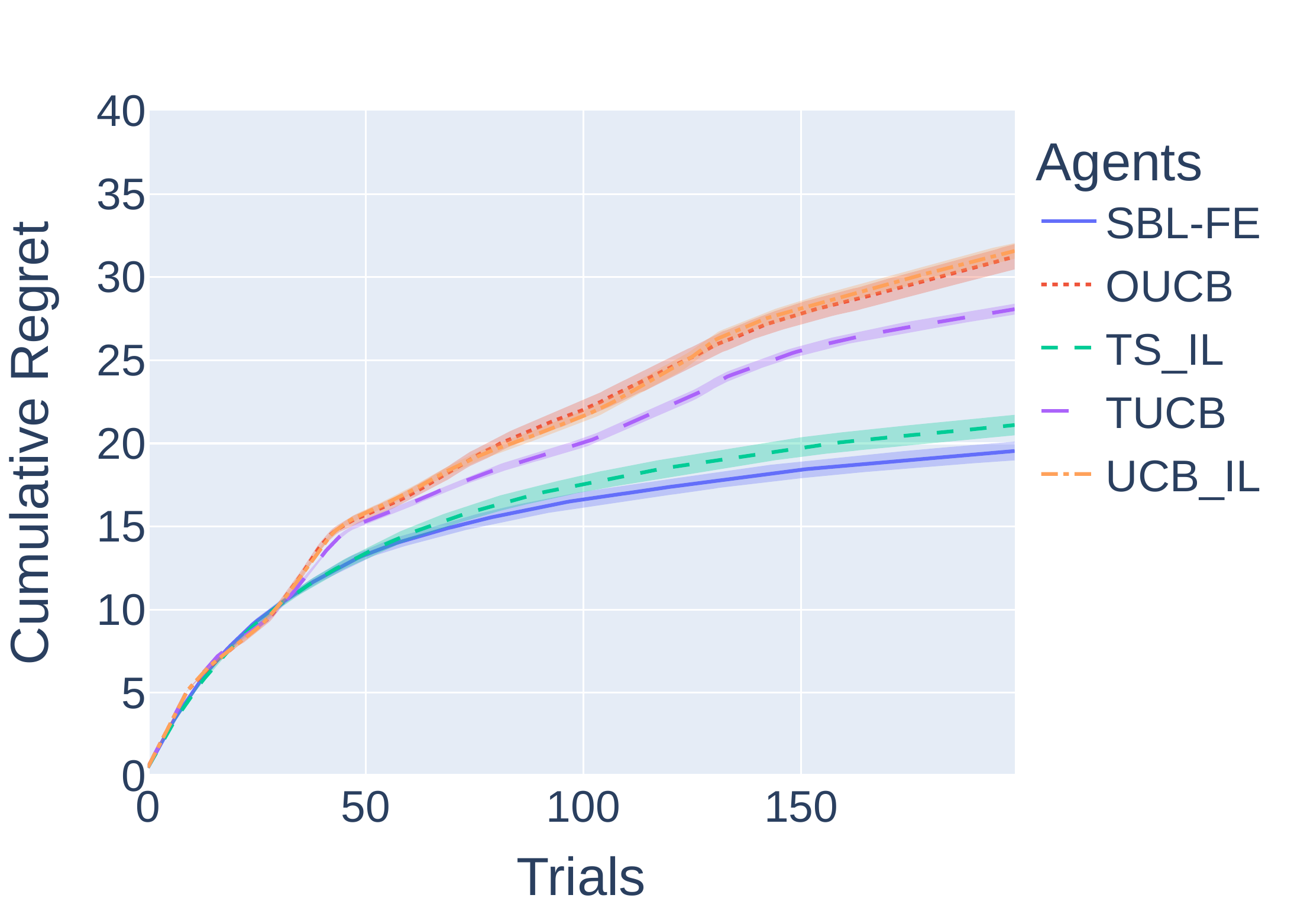}
        \includegraphics[width=0.48\linewidth]{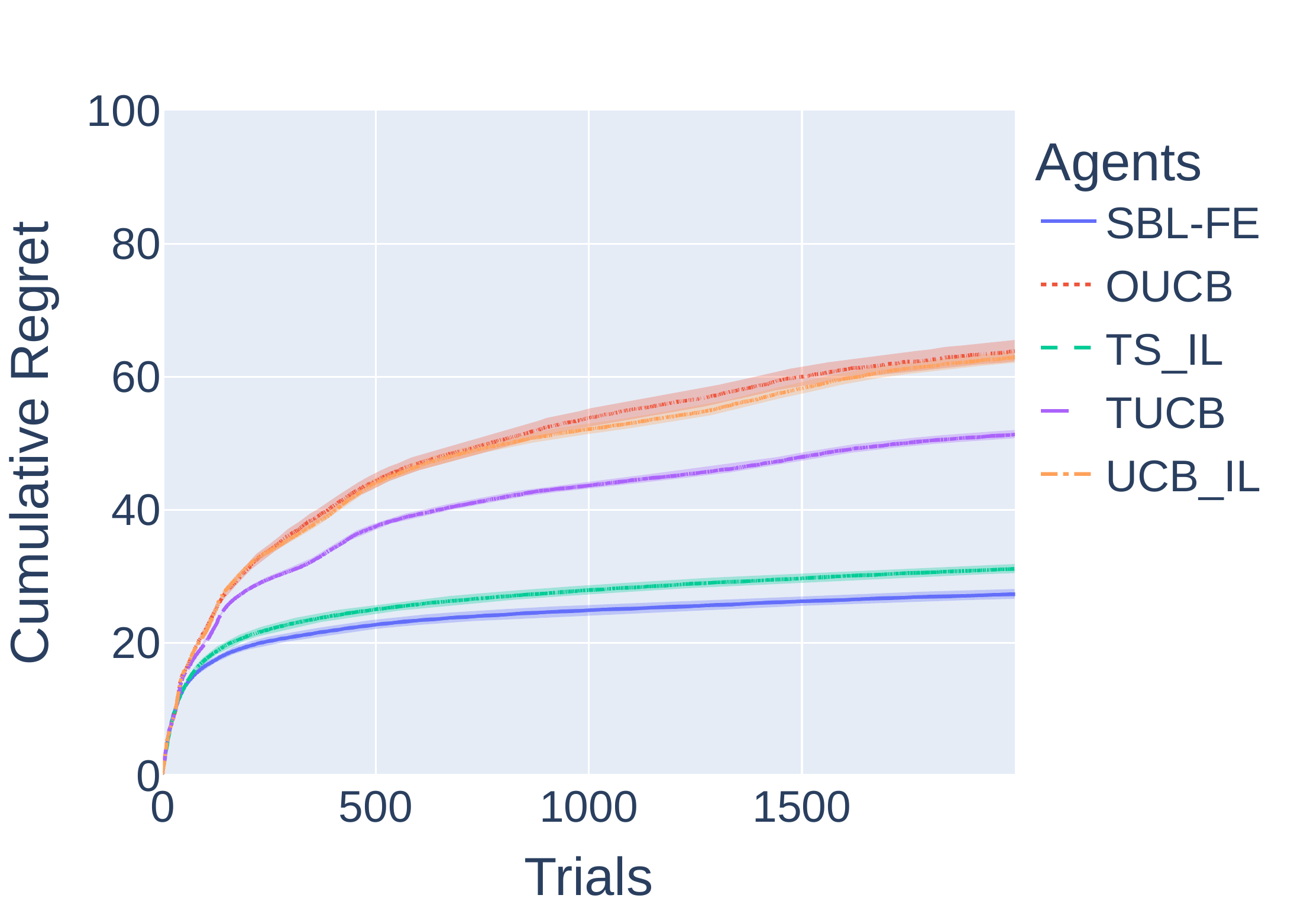}
        \subcaption*{(c) UCB}
    \end{subfigure}

    \caption{Cumulative regret performance of three social learning agents (OUCB, TUCB, SBL-FE) along with UCB and TS as baseline methods in societies consisting of one social learner and one individual learner. The experiments were conducted over 200 and 2000 trials for a 10-armed Bernoulli bandit problem with an optimality gap of $\Delta = 0.2$.}
    \label{fig:learners}
\end{figure}

In this section, we consider a society consisting of an individual learner (TS, UCB, or Epsilon-greedy) and a social agent. Fig. \ref{fig:learners} presents a comparison of different social learning algorithms, including UCB and TS as baseline methods, within this society in terms of cumulative regret. Remarkably, our method consistently outperforms an individual Thomson Sampling (TS) learner in all cases, even when it incorporates observations from learner methods that perform equally or weaker than the TS learner. This is made possible by the flexibility of our agent, which can select between two individual learners in each run. Consequently, there are instances where the weaker individual learning algorithms, namely UCB and Epsilon-greedy, outperform TS. This highlights the value of diversity in learning methods, as it enhances overall learning performance even when slower individual learning methods are utilized as demonstrators within society. On the other hand, TUCB demonstrates performance improvement only in cases when the individual learner outperforms UCB. We gained the same results across different optimality gaps. 

\subsection{The ability to detect better agents}
Given the existence of multiple suitable agents within a society, an agent must possess the capability to discern superior learning sources. This discernment becomes crucial when certain agents exhibit more favorable attributes, such as quicker learning abilities or other advantageous traits. The social agent must, therefore have the ability to identify the most advantageous agents within the pool of suitable options, enabling it to derive maximum benefit from the process of social learning.

To assess the agent's ability to identify the most suitable agents, we conducted experiments using three distinct societies in the context of a 10-armed Bernoulli bandit problem with an optimality gap of $\Delta = 0.2$. In the first society, we included one social learner alongside two non-learners (an opponent and a random agent), as well as an individual learner employing the epsilon-greedy algorithm. The second society comprised a social learning agent in conjunction with optimal, sub-optimal, and epsilon-greedy agents. The final setting involved observing the behavior of two P-optimal agents through the lens of the social agent. The first P-optimal agent commenced with an initial value of P equal to one, decreasing by 0.001 after each trial. Consequently, by trial 1000, the agent's behavior mirrored that of a random agent. The other P-optimal agent exhibited the opposite behavior of the first P-optimal agent exactly.

\begin{figure}[htbp]
    \centering
    
    \begin{subfigure}[t]{0.48\textwidth}
        \centering
        \includegraphics[width=0.48\linewidth]{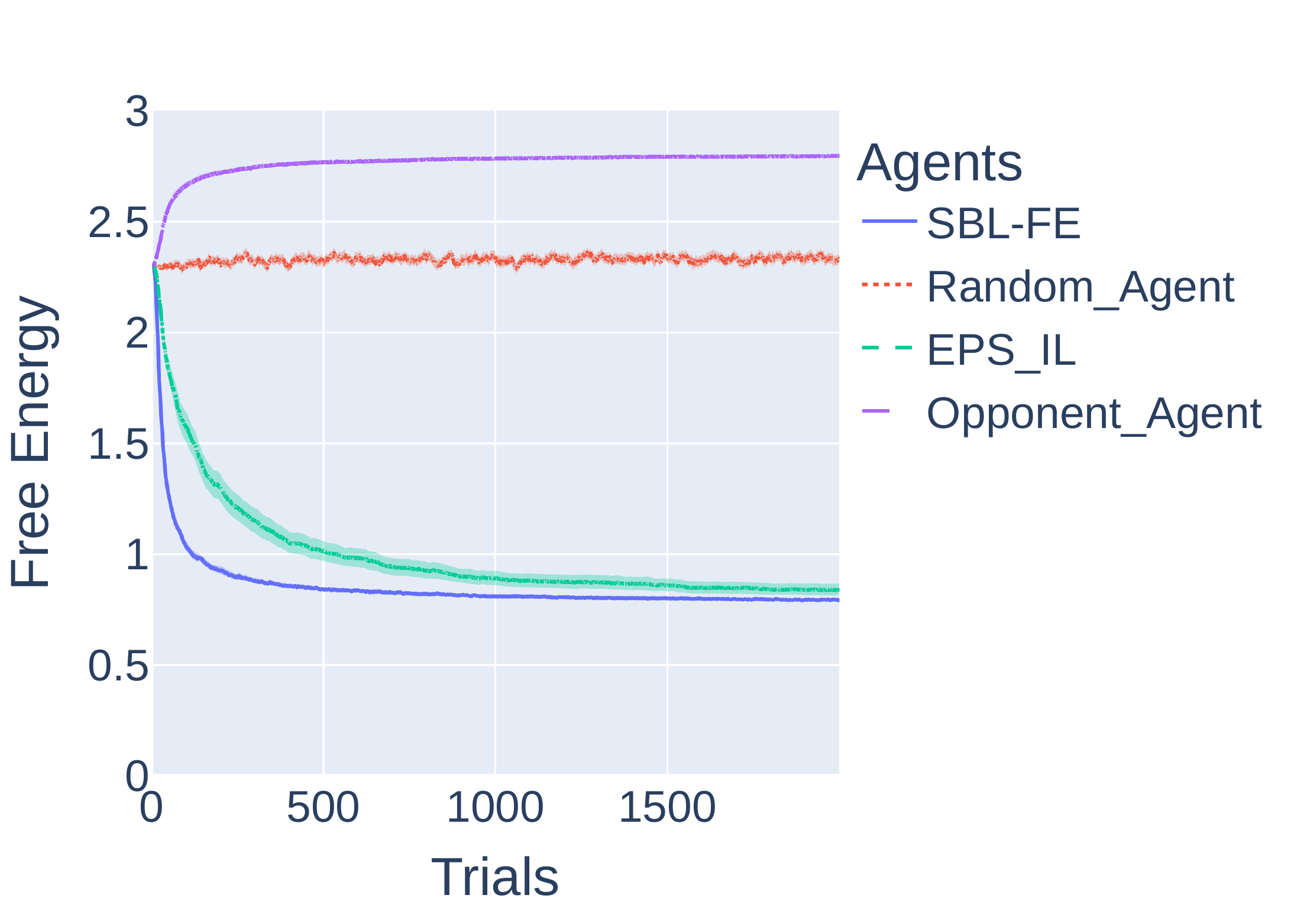}
        \hfill
        \includegraphics[width=0.48\linewidth]{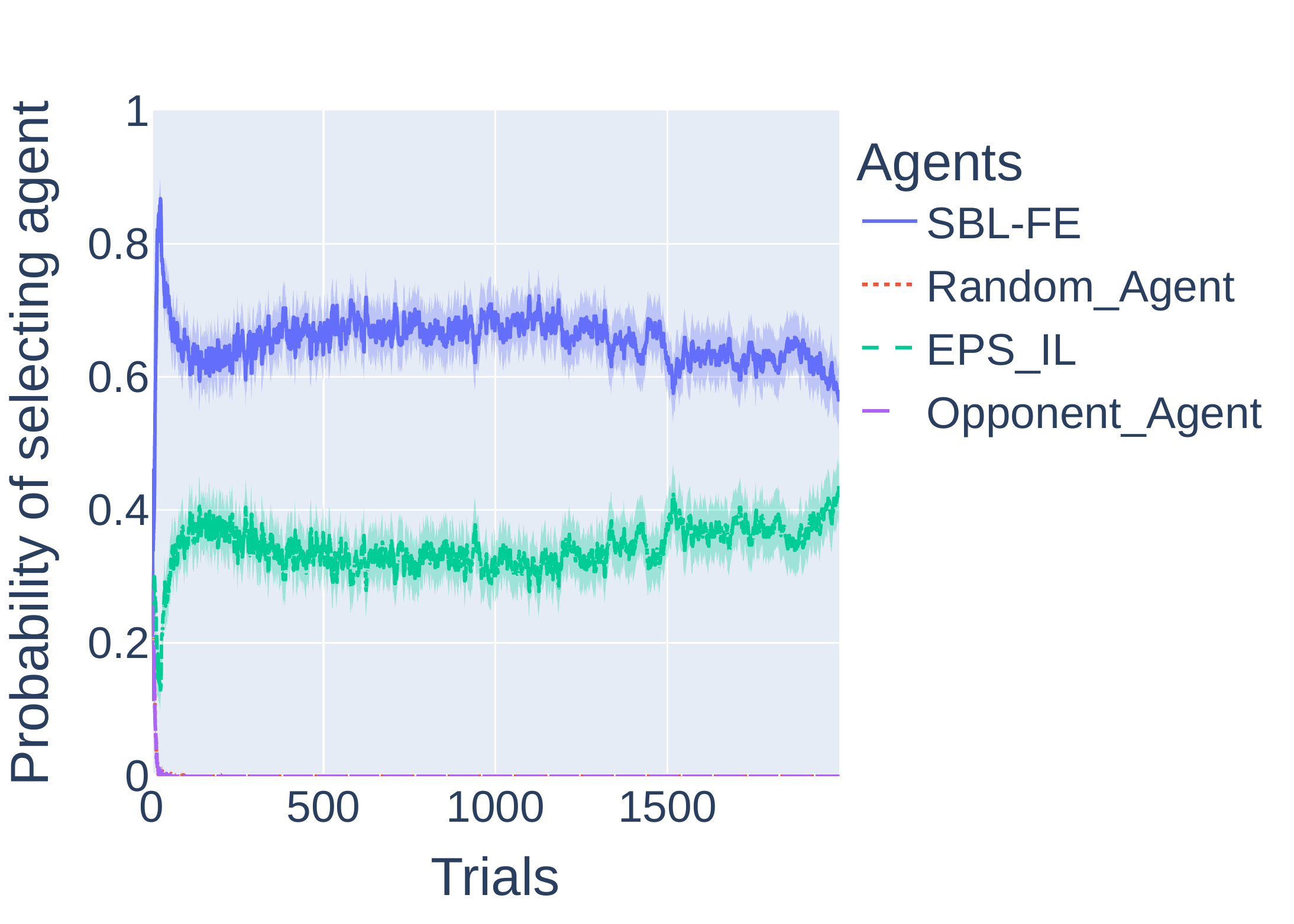}
        \subcaption*{(a) Opponent, random, and epsilon-greedy agents}
    \end{subfigure}
    \vspace{0.2cm}
    
    \begin{subfigure}[t]{0.48\textwidth}
        \centering
        \includegraphics[width=0.48\linewidth]{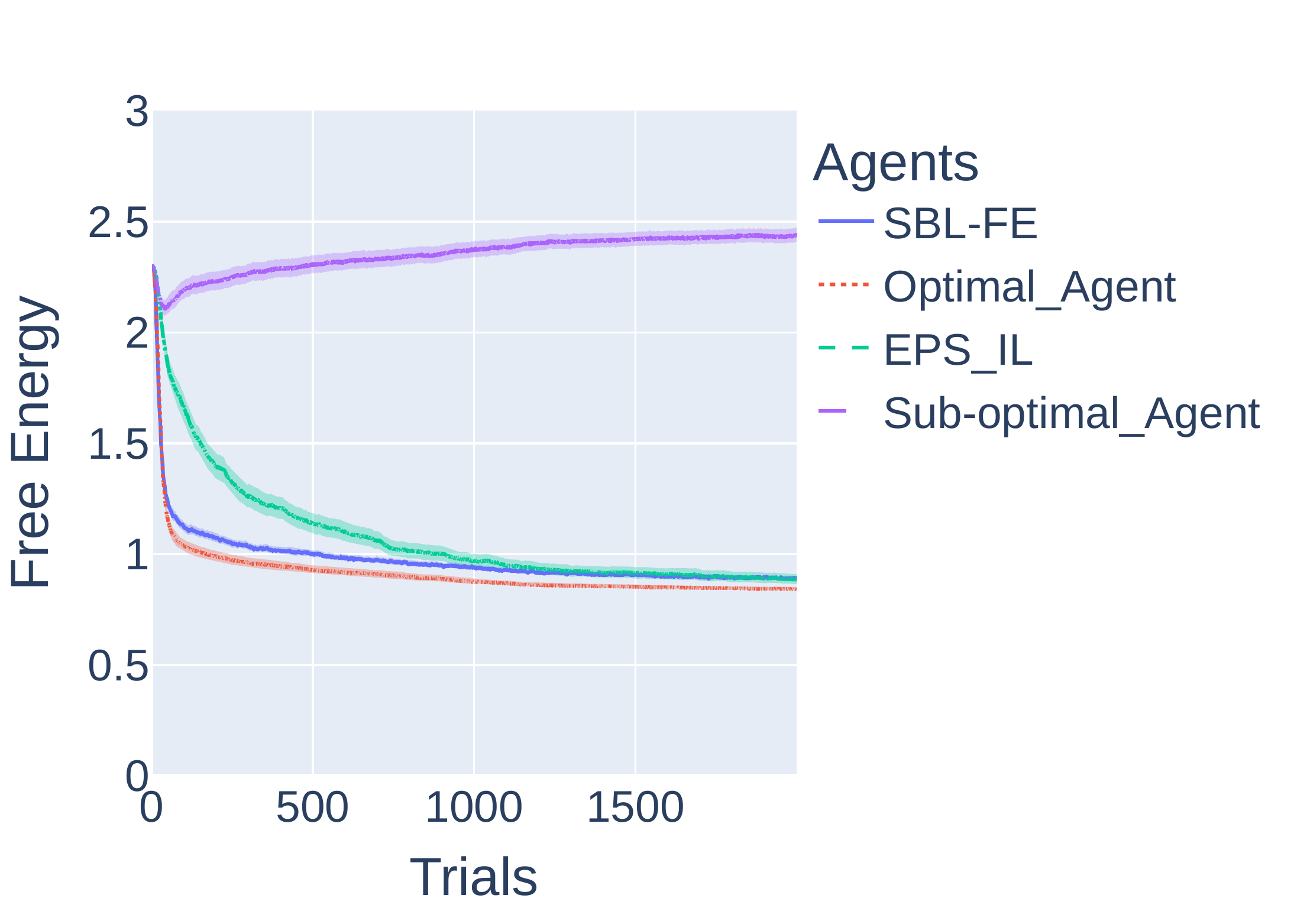}
        \hfill
        \includegraphics[width=0.48\linewidth]{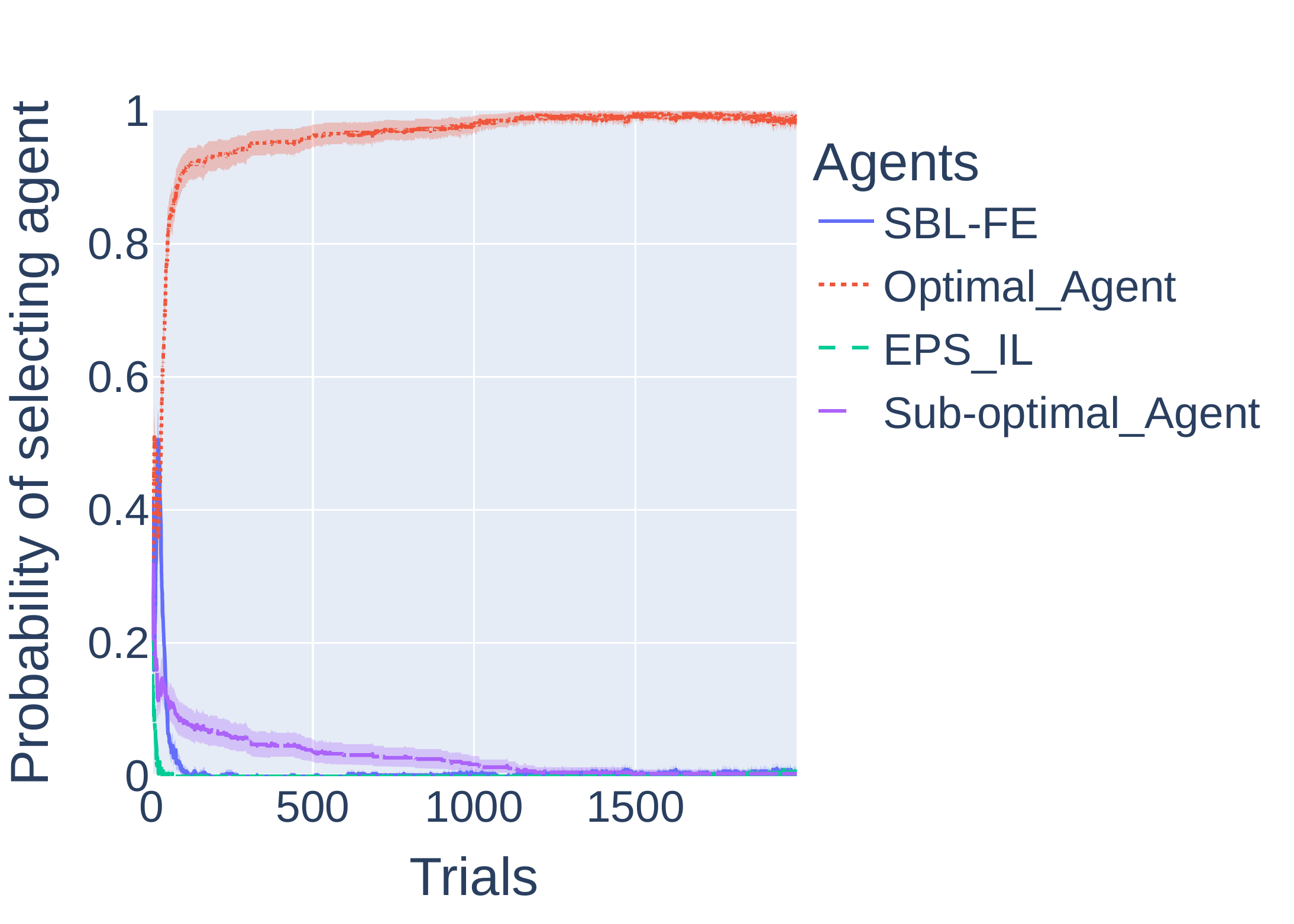}
        \subcaption*{(b) Optimal, sub-optimal, and epsilon-greedy agents}
    \end{subfigure}
    \vspace{0.2cm}
    
    \begin{subfigure}[t]{0.48\textwidth}
        \centering
        \includegraphics[width=0.48\linewidth]{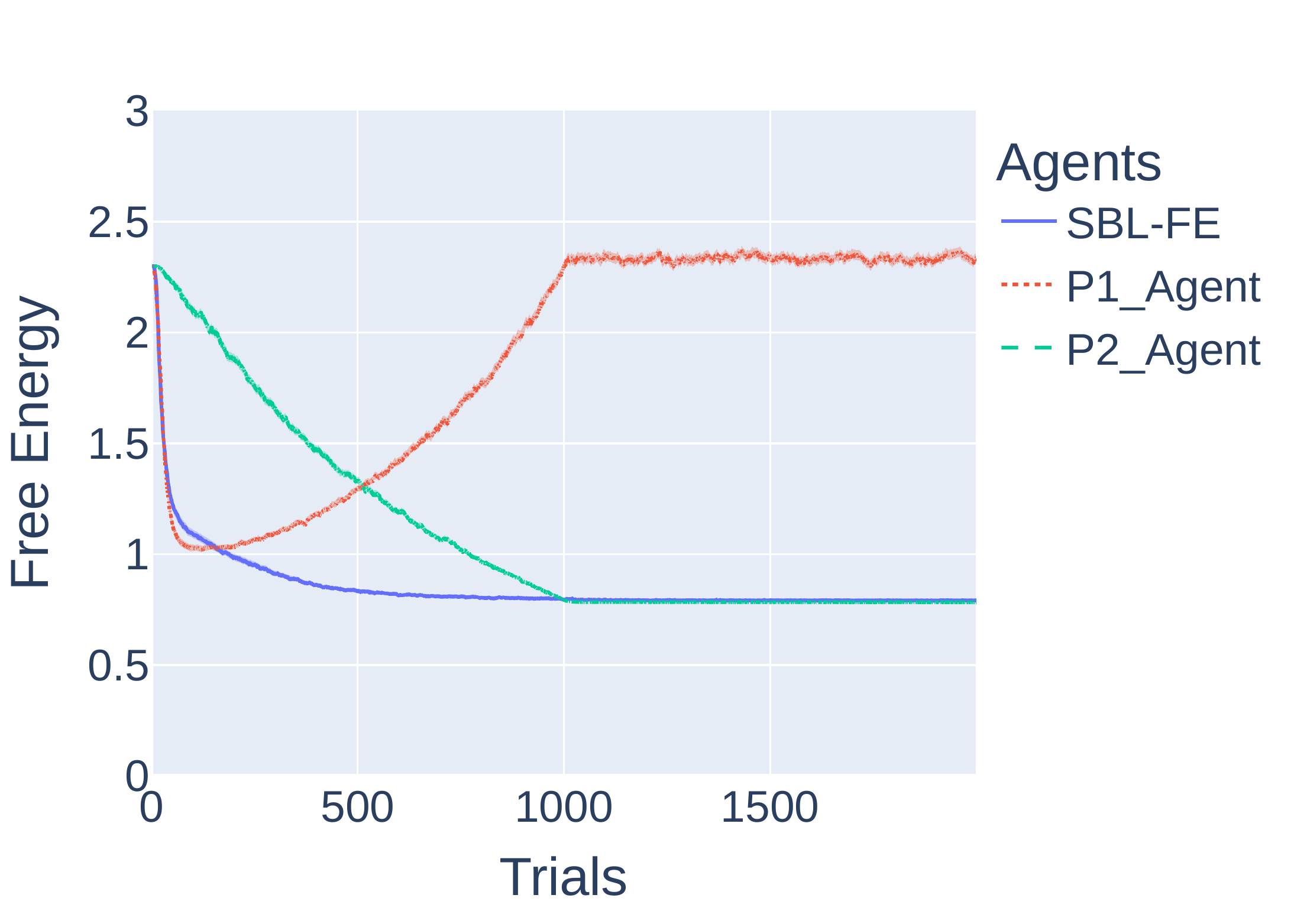}
        \hfill
        \includegraphics[width=0.48\linewidth]{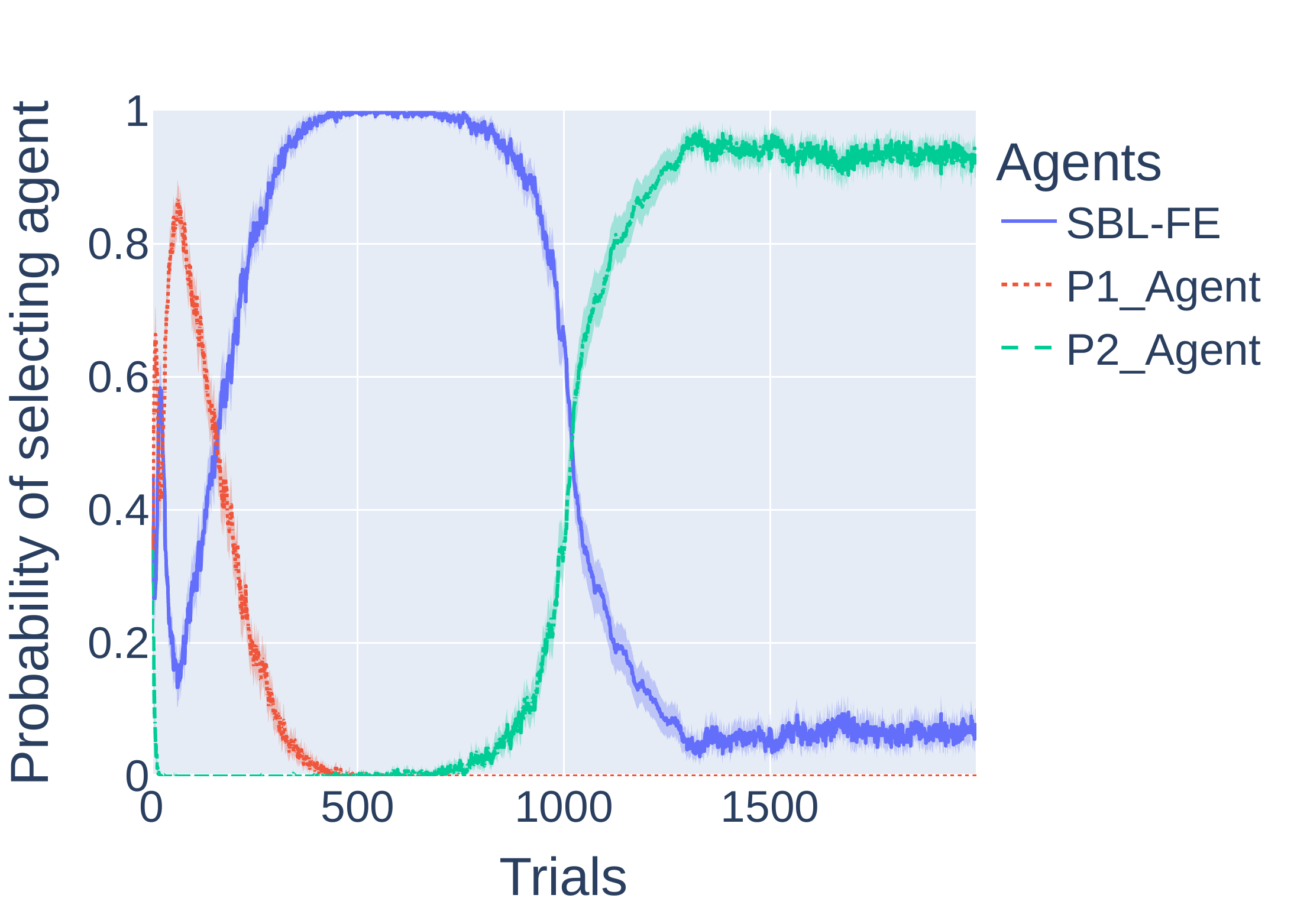}
        \subcaption*{(c) Two opposing P-optimal agents}
    \end{subfigure}
    
    \caption{Per-trial free energy and selection probability of our social agent, SBL-FE, in different societal setups. 2000 trials were conducted for a 10-armed Bernoulli bandit problem with $\Delta = 0.2$.}
    \label{fig:detecting}
\end{figure}

In Fig.~\ref{fig:detecting}, we present the free energy of $\pi^{*}_{ag_i}$ (where $ag_i$ refers to each agent) and the probability of the social agent selecting other agents (including itself) per trial during the learning process within these three societies. Initially, the selection probabilities are uniform. The findings, particularly from the third experiment, demonstrate the adaptive nature of free energy as a measure capable of accurately detecting the relevance of other agents. In the second setting, we observe a pattern where the sub-optimal agent is selected more frequently during the early trials. However, as the optimal action is confidently identified, the selection probability of the sub-optimal agent decreases to zero. We obtained the same results across different optimality gaps.

\subsection{The ability to cope with agents with different action sets}

Similar to many real-world cases, agents may have different action sets. For the sake of convenience, we assume that the individual learning agents' action sets are subsets of the SA's. However, this assumption can be relaxed easily to more generic scenarios. In that case, SA can ignore observations that do not include its actions. 

We tested our SA in a society with three epsilon-greedy learner agents with different action sets in the context of a 10-armed Bernoulli bandit problem with an optimality gap of $\Delta = 0.2$. The action sets of these three epsilon greedy agents are disjoint from each other, and each includes three actions. The optimal choice of SA and one of the agents is the same, while the other ones have different optimal actions. 

\begin{figure}[htbp]
    \centering
    \begin{minipage}[t]{0.23\textwidth}
        \centering
        \includegraphics[width=\linewidth]{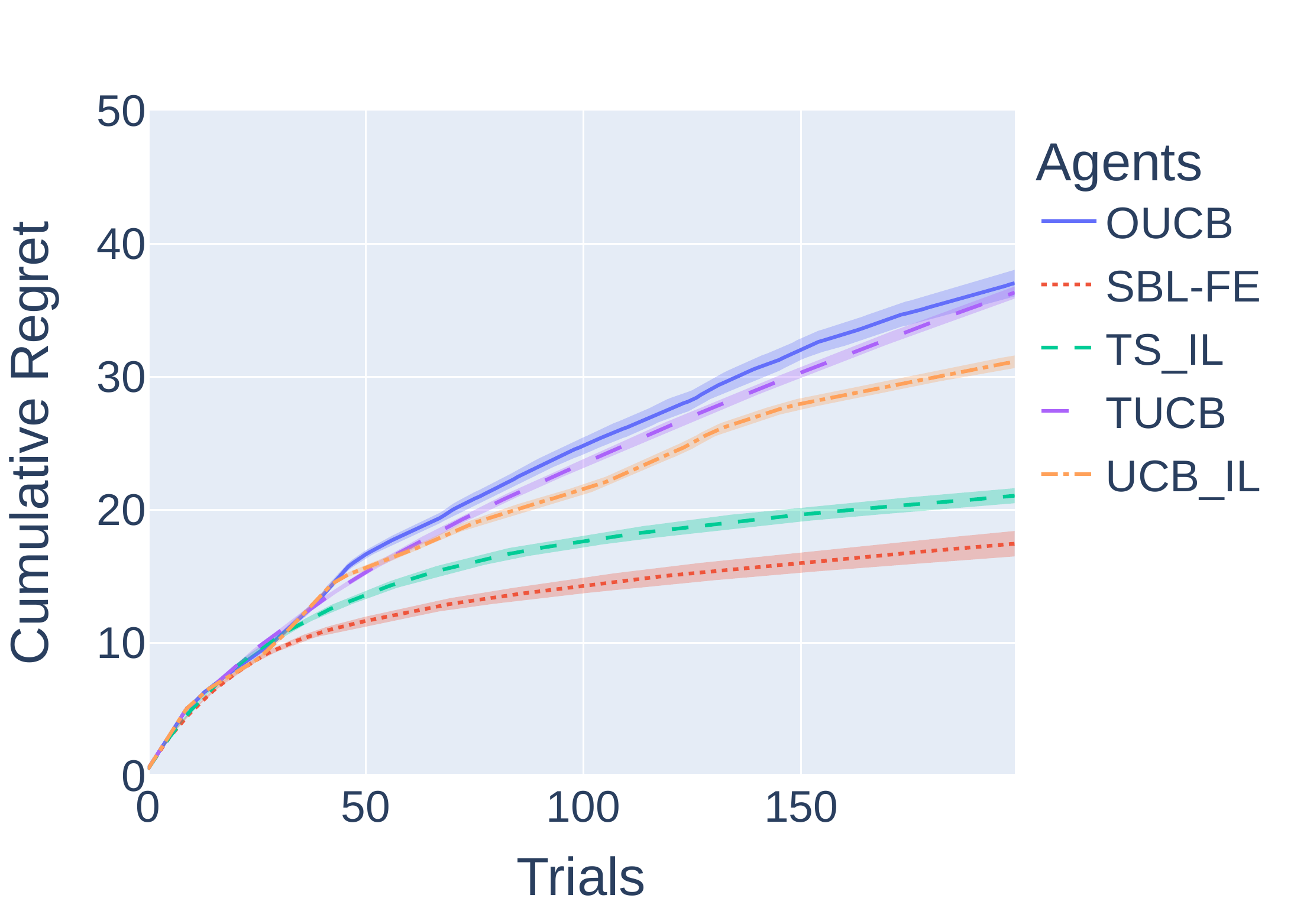}
    \end{minipage}
    \hfill
    \begin{minipage}[t]{0.23\textwidth}
        \centering
        \includegraphics[width=\linewidth]{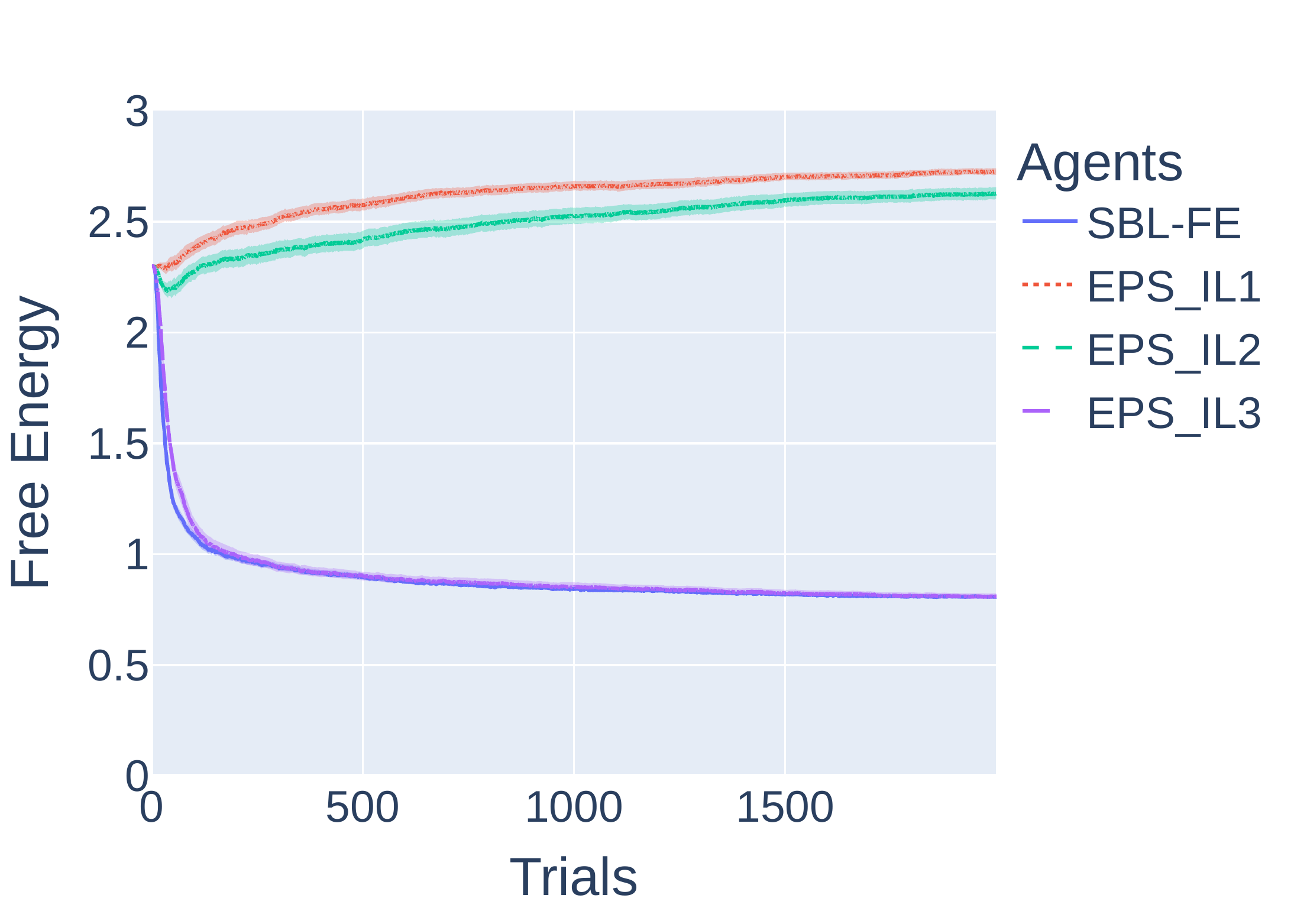}
    \end{minipage}
    \vspace{0.1cm} 
    \begin{minipage}[t]{0.23\textwidth}
        \centering
        \includegraphics[width=\linewidth]{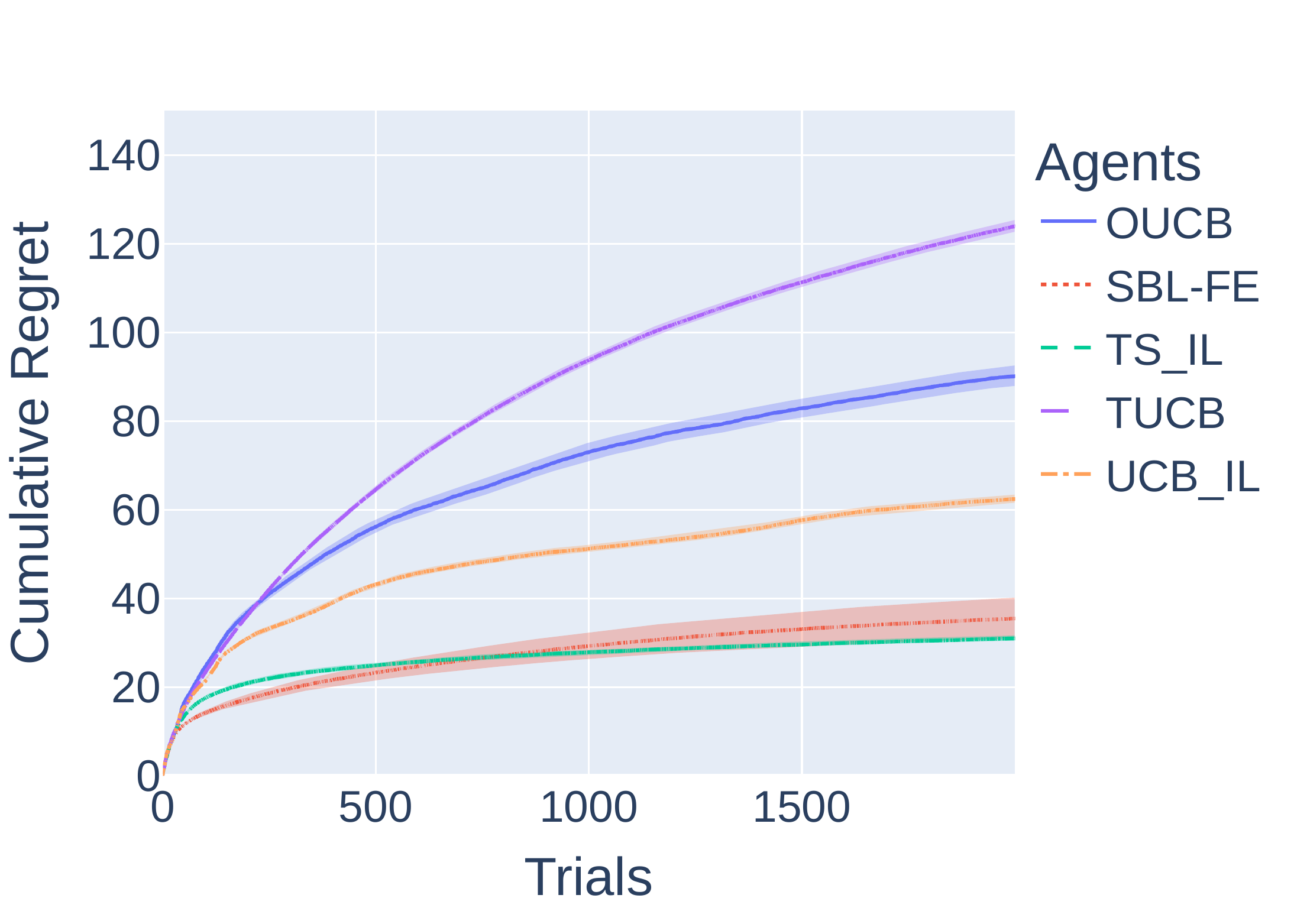}
    \end{minipage}
    \hfill
    \begin{minipage}[t]{0.23\textwidth}
        \centering
        \includegraphics[width=\linewidth]{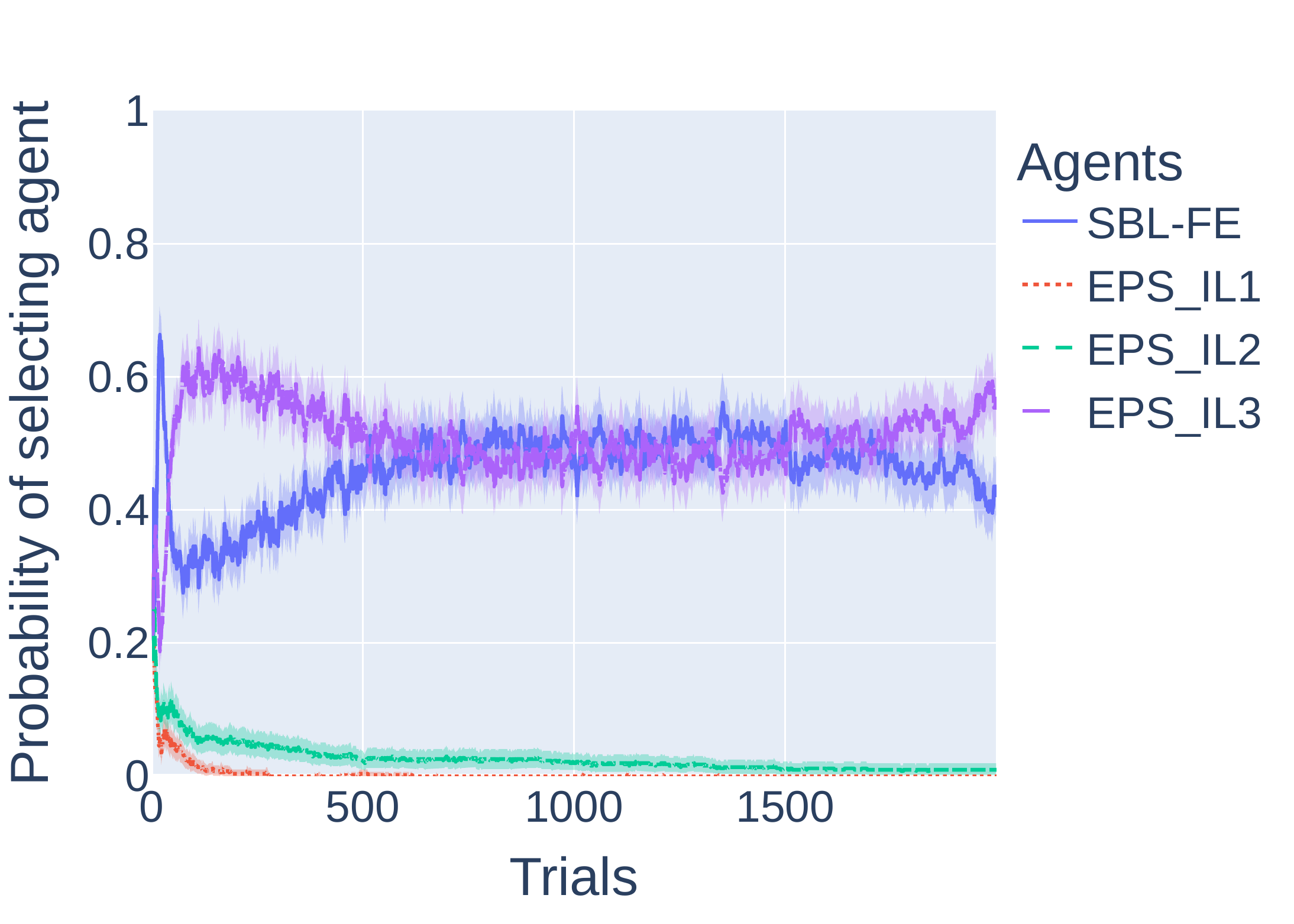}
    \end{minipage}
    
    \caption{The Per-trial free energy, selection probability, and cumulative regret of three social learning agents (OUCB, TUCB, SBL-FE) along with UCB and TS as baseline methods in a society consisting of one social learner and three epsilon-greedy agents. The experiments were conducted over 200 and 2000 trials for a 10-armed Bernoulli bandit problem with an optimality gap of $\Delta = 0.2$. The action sets of these three epsilon greedy agents are disjoint from each other, and each includes three actions.     
    $EPS\_IL3$ and SA share the same optimal action, while the other two individual agents have different optimal decisions.}
    \label{fig:subset}
\end{figure}

In Fig.~\ref{fig:subset}, we present the free energy of $\pi^{*}_{ag_i}$ (where $ag_i$ refers to each agent) and the probability of the SA selecting other agents (including itself) per trial during the learning process. Additionally, we compared the cumulative regret of different social learning algorithms, including UCB and TS as baseline methods throughout the learning process. The results show that, unlike OUCB and TUCB, SBL-FE significantly benefits from observing the individual learners during the early 200  trials, which is very important in budgeted learning. The reason behind this is related to that OUCB and TUCB look optimistically toward other agents in society and assume that other agents learn the same task. In contrast, the SBL-FE algorithm just needs to observe relevant actions to enhance its learning or otherwise just ignore them. In addition, the probability of the SA selecting the best epsilon-greedy agent is more than the experiment of Fig. \ref{fig:detecting}. This is because here the action set of the individual learners is smaller and they learn faster. We got the same results at different optimality gaps. 

\subsection{The influence of society’s population and problem difficulty}

It is necessary for a social learning method to effectively identify relevant agents, even in densely populated societies. Therefore, in this section, our objective is to demonstrate the robustness of our method towards different societal sizes. We conducted experiments using three distinct societies within the framework of a 10-armed Bernoulli bandit problem with an optimality gap of $\Delta = 0.2$. In order to analyze the susceptibility of our social learner to irrelevant agents, we introduced three opponent agents and two random agents alongside a social learner, in addition to either an optimal agent or an epsilon-greedy agent. Although random agents act differently in each trial, they lack any relevant information. Hence, a robust social learning method should ignore their influence. Similarly, in the presence of an optimal agent, we expect the performance of our social agent to remain unaffected by the addition of other agents, including learners.

\begin{figure}[htbp]
  \centering

  \begin{subfigure}[t]{0.49\textwidth} 
    \centering
    \includegraphics[width=0.49\linewidth]{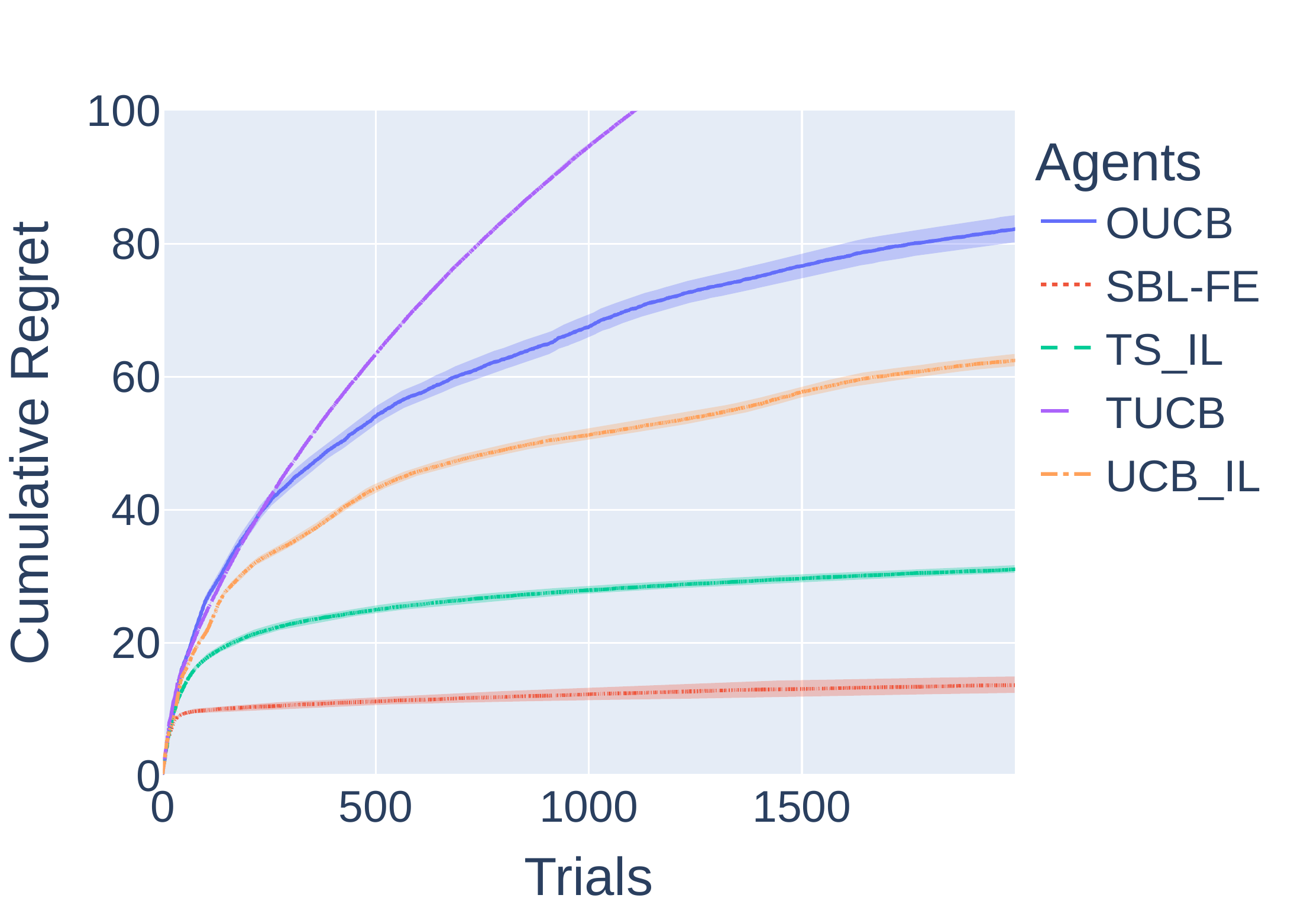}%
    \hfill
    \includegraphics[width=0.49\linewidth]{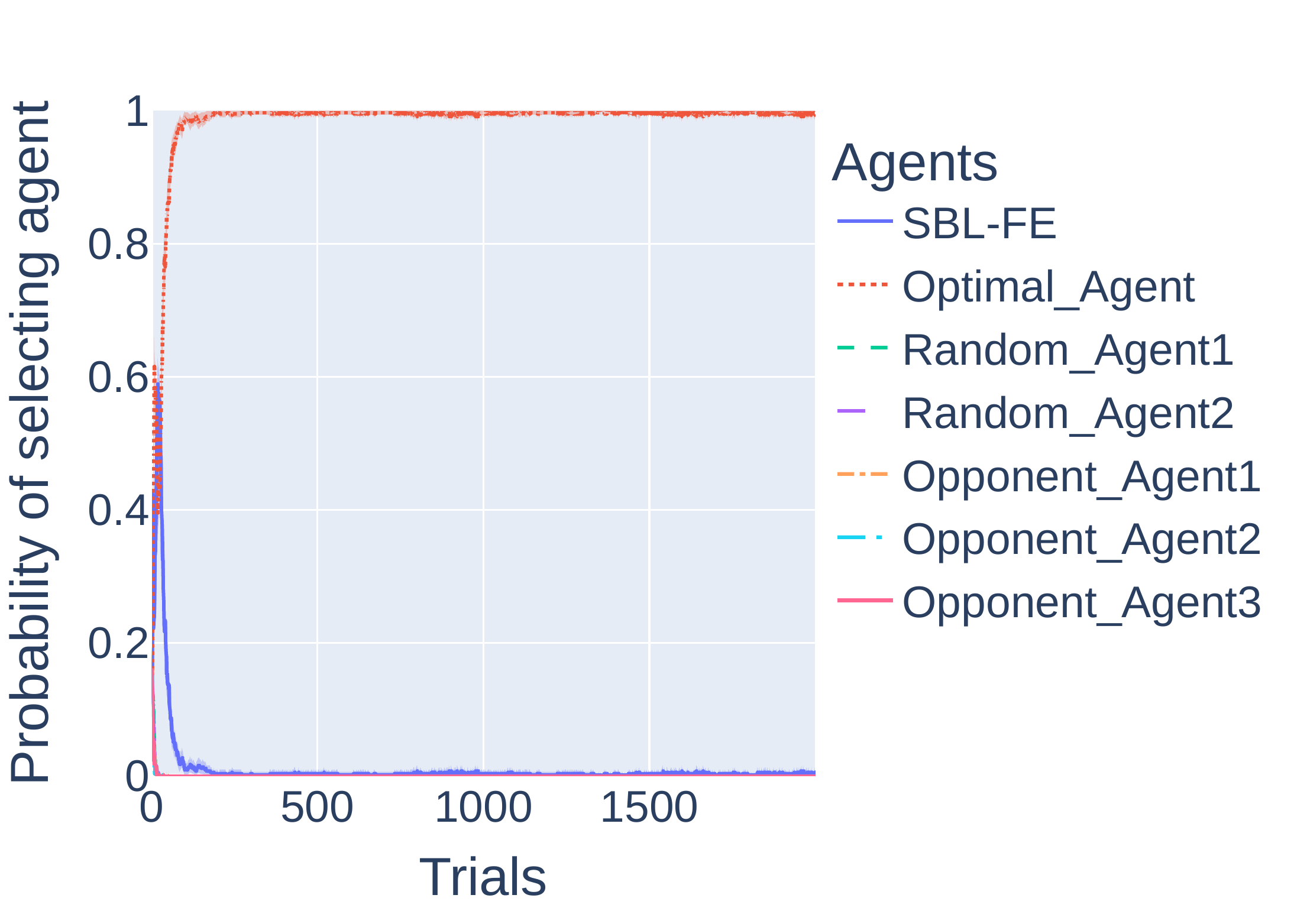}
    \subcaption{Optimal agent with three opponent and two random agents}
    \label{fig:group-a}
  \end{subfigure}
  \hfill
  \begin{subfigure}[t]{0.49\textwidth}
    \centering
    \includegraphics[width=0.49\linewidth]{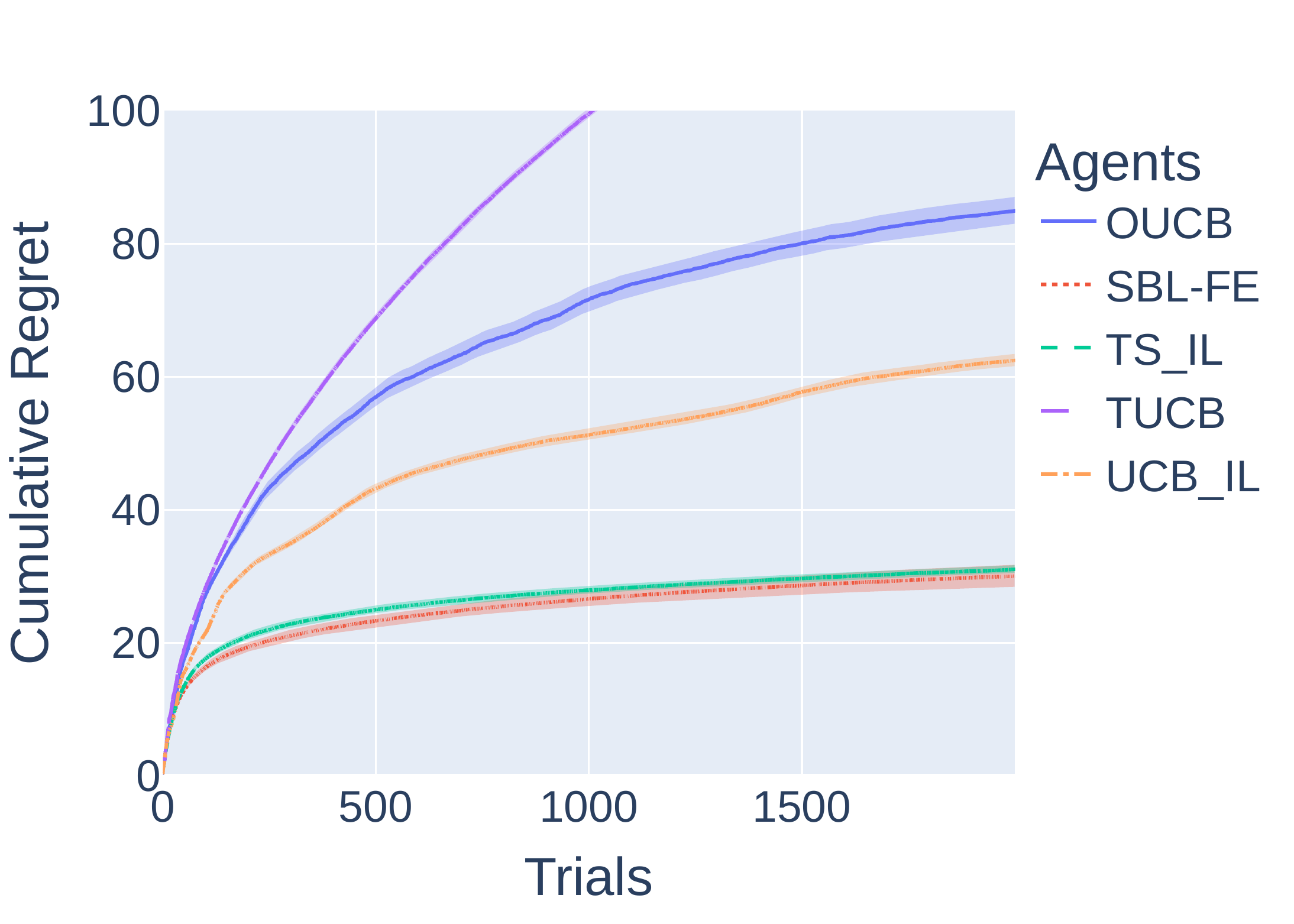}%
    \hfill
    \includegraphics[width=0.49\linewidth]{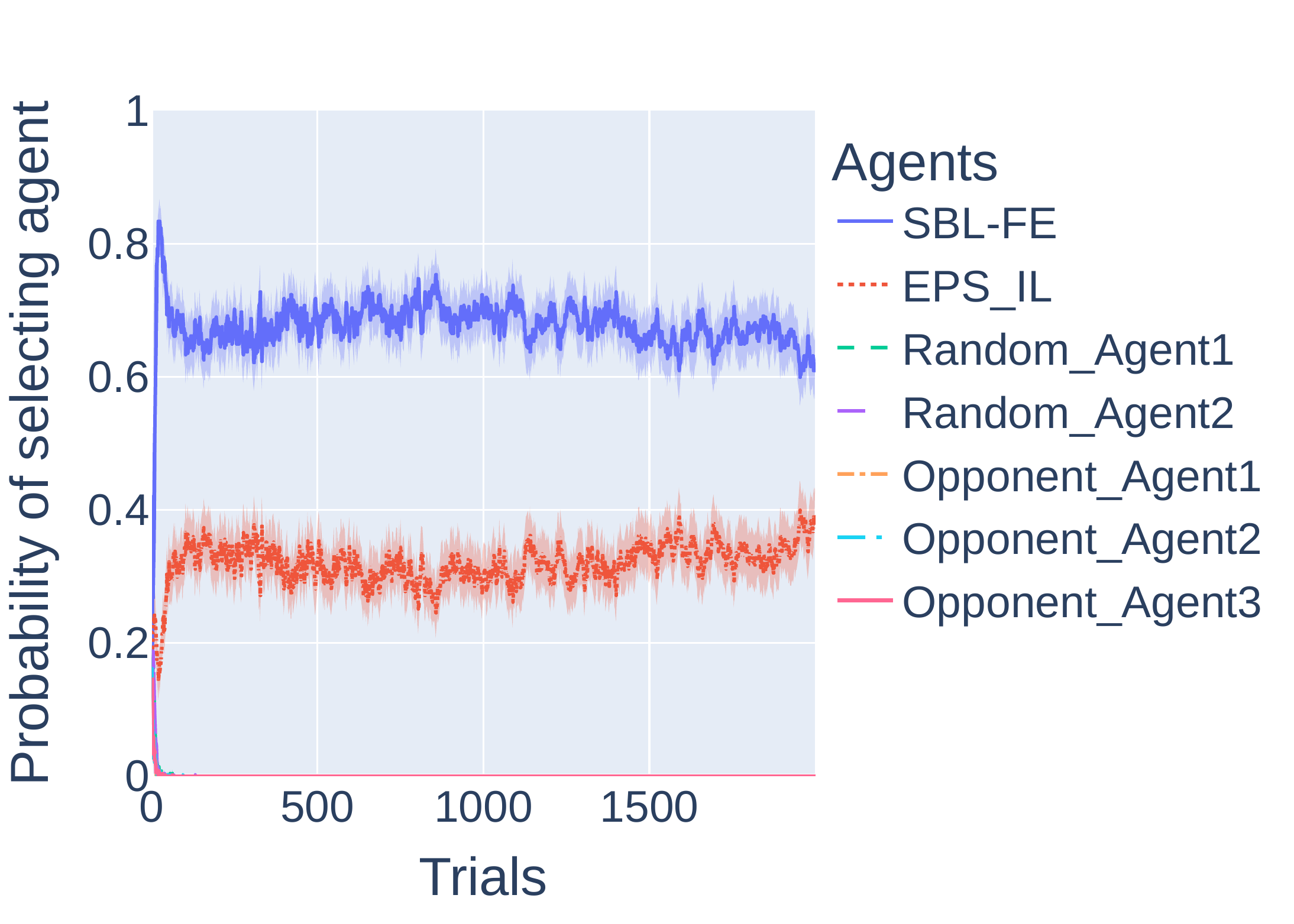}
    \subcaption{Epsilon-greedy agent with three opponent and two random agents}
    \label{fig:group-b}
  \end{subfigure}
  \hfill
  \begin{subfigure}[t]{0.49\textwidth}
    \centering
    \includegraphics[width=0.49\linewidth]{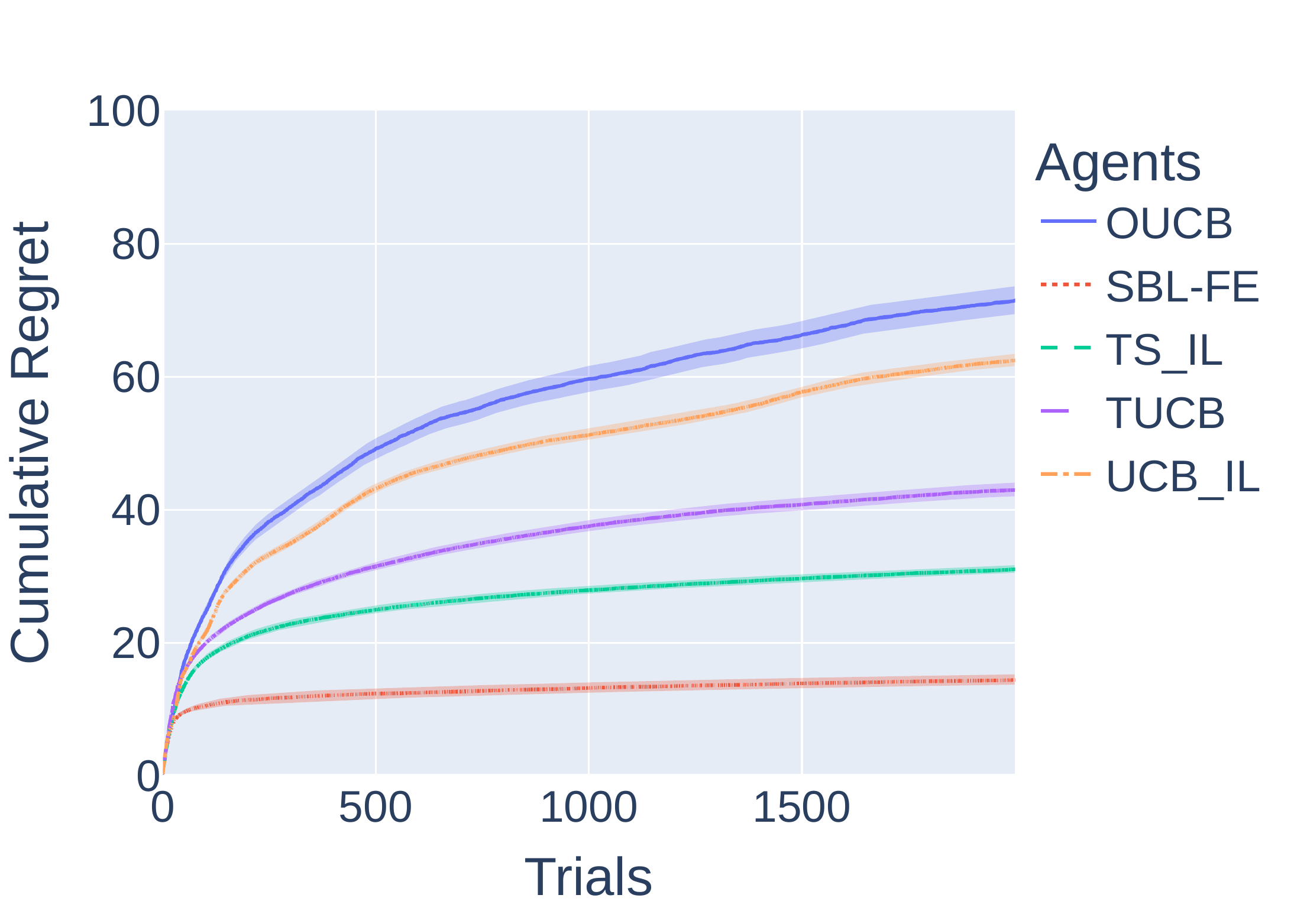}%
    \hfill
    \includegraphics[width=0.49\linewidth]{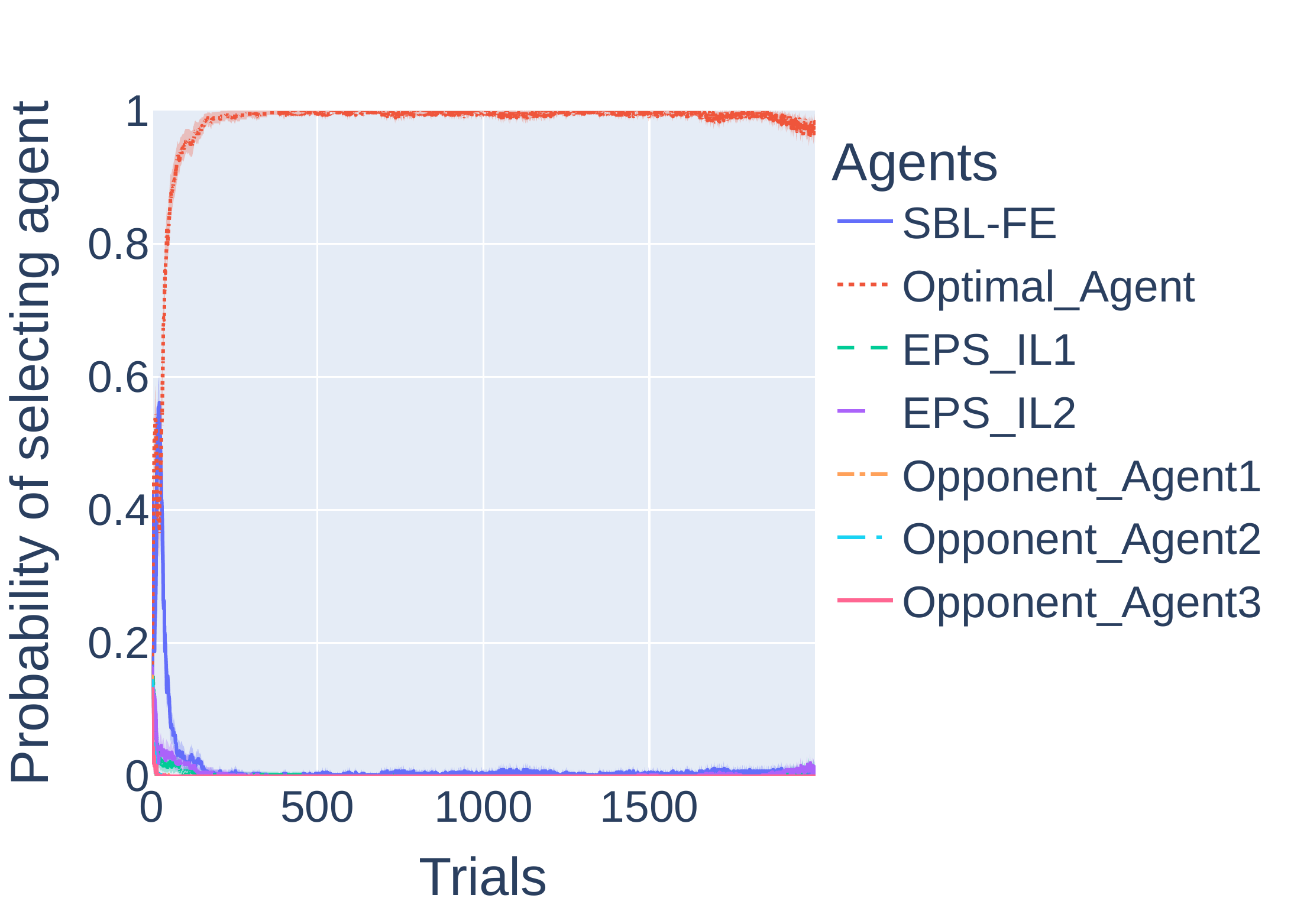}
    \subcaption{Optimal agent with two epsilon-greedy and three opponent agents}
    \label{fig:group-c}
  \end{subfigure}

  \vspace{0.6em} 
  \caption{The cumulative regret performance of three social learning agents (OUCB, TUCB, SBL-FE) along with UCB and TS as baseline methods and per-trial selection probability of SBL-FE, our social agent, in different societal setups. Conducted over 2000 trials for a 10-armed Bernoulli bandit problem with $\Delta = 0.2$.}
  \label{fig:worst_random_best_eps}
\end{figure}

Fig.~\ref{fig:worst_random_best_eps} illustrates the per-trial probability of the social agent selecting other agents (including itself)  and the cumulative regret of different social learning algorithms, including UCB and TS as baseline methods throughout the learning process. As anticipated, the introduction of additional random, opponent, or epsilon-greedy agents in the presence of an optimal agent does not distract our social learning method, as the SA did not choose to learn from them at all. This is attributed to our selection criterion of choosing the agent with the minimum free energy. 
When there is only one relevant agent (either optimal or epsilon-greedy) among multiple opponents and random agents, both OUCB and TUCB algorithms exhibit notably poor performance. Naturally, in real-world scenarios, it is common that the majority of agents within a society lack relevant information or do not share our task perspective. In such cases, our algorithm demonstrates the ability to efficiently identify the appropriate agent. We gained the same results at other optimality gaps as well.

We conducted a series of experiments, as described in Section \ref{sec:ability}, using a multi-armed bandit with two arms to investigate how the algorithm's behavior is influenced by variations in problem difficulty. In this context, we utilized a 2-armed Bernoulli bandit problem, where one arm had an expected reward of 0.5, while the expected value of the second arm was modified to manipulate the problem's difficulty.

\begin{figure}[htbp]
    \centering
    
    \begin{subfigure}[t]{0.48\textwidth}
        \centering
        \includegraphics[width=0.48\linewidth]{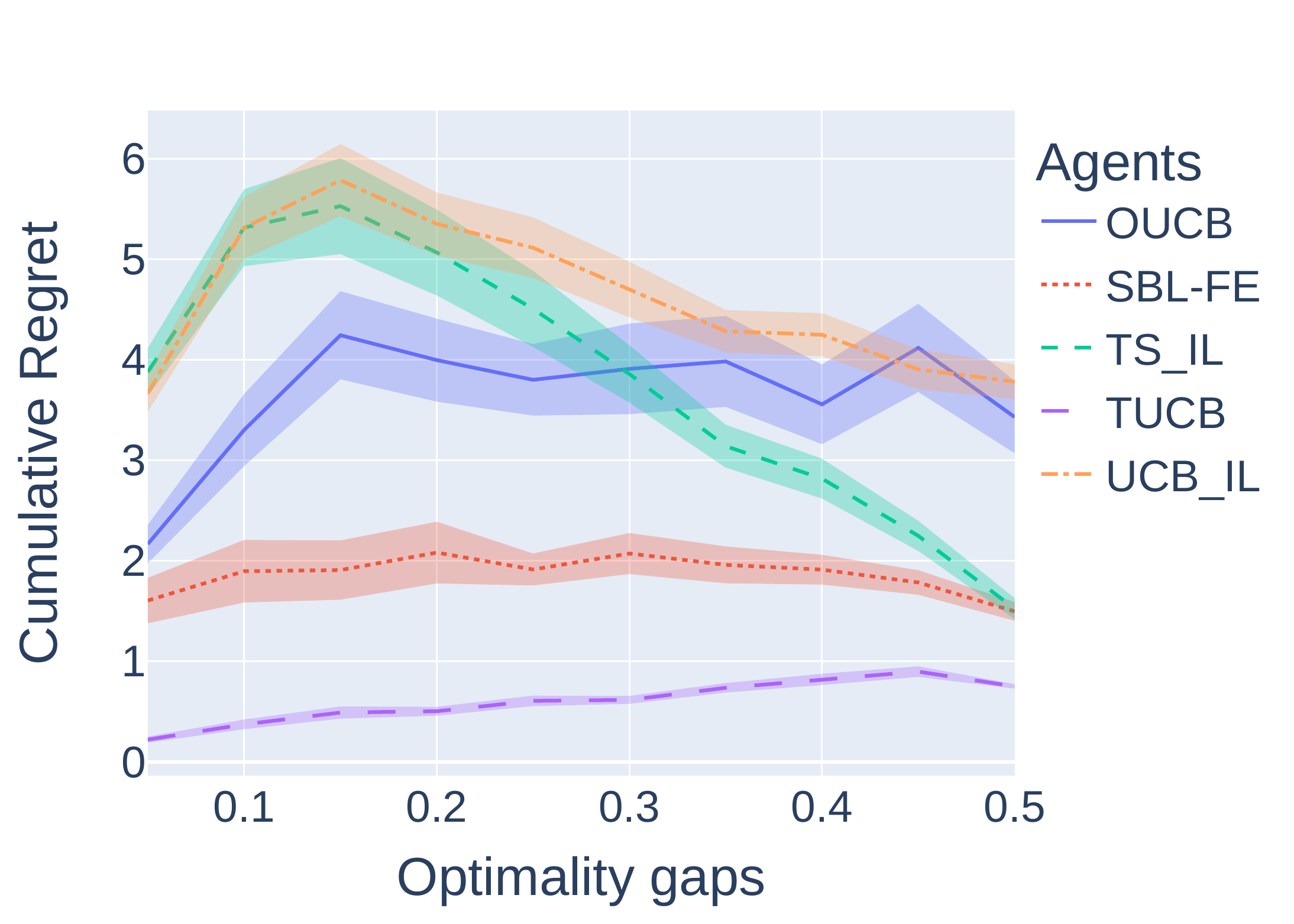}
        \hfill
        \includegraphics[width=0.48\linewidth]{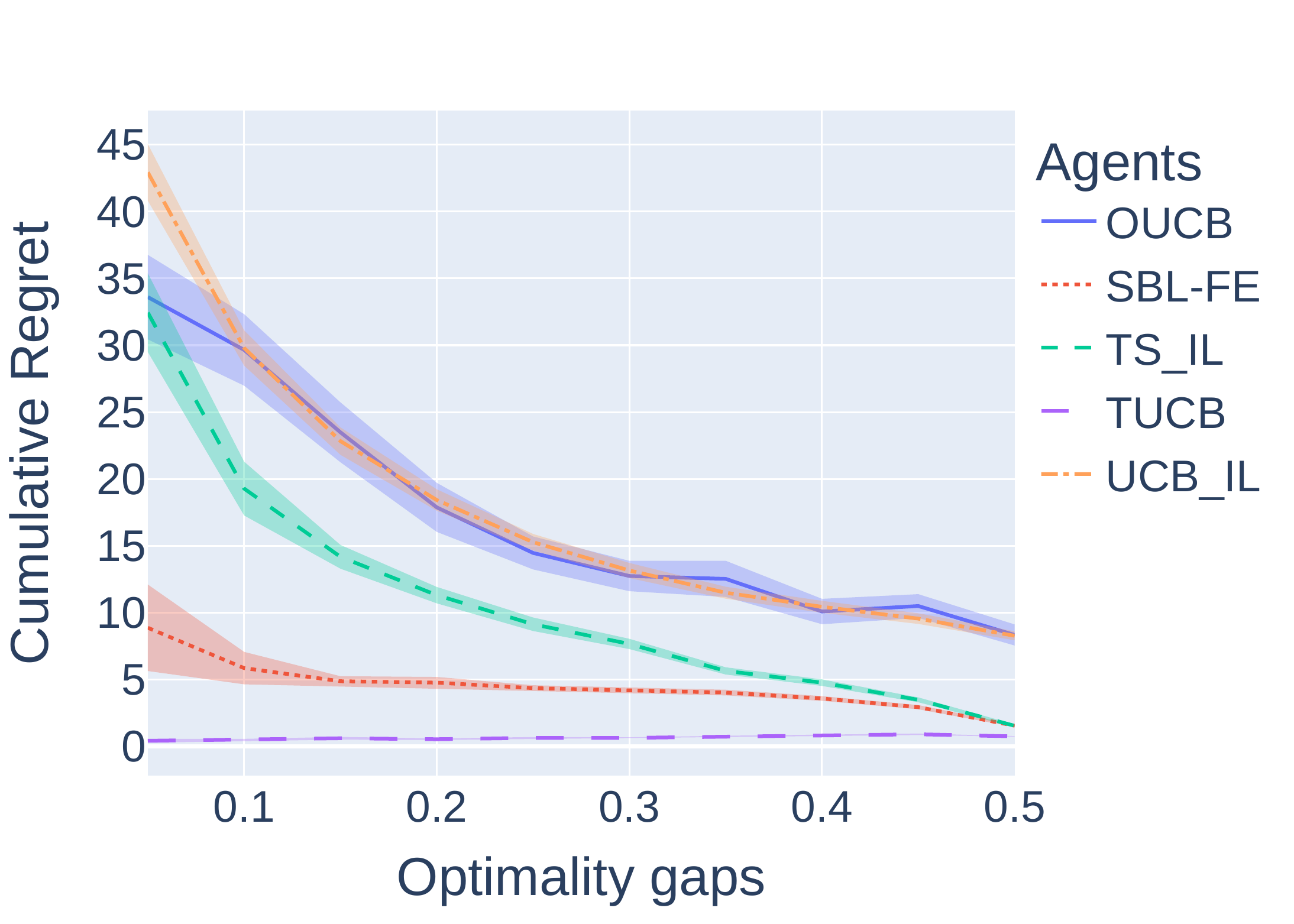}
        \subcaption*{(a) Optimal agent}
    \end{subfigure}
    \vspace{0.2cm}
    
    \begin{subfigure}[t]{0.48\textwidth}
        \centering
        \includegraphics[width=0.48\linewidth]{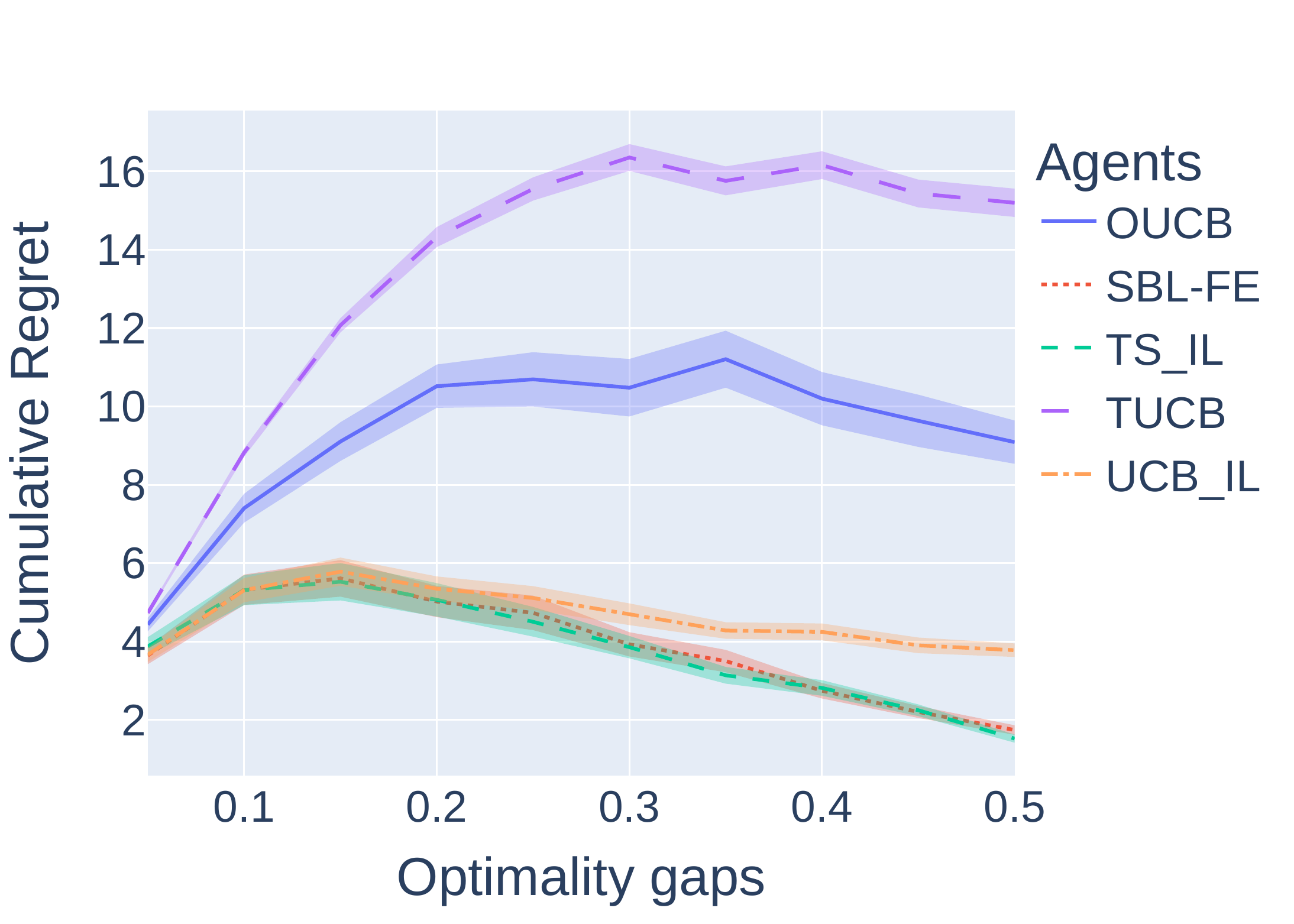}
        \hfill
        \includegraphics[width=0.48\linewidth]{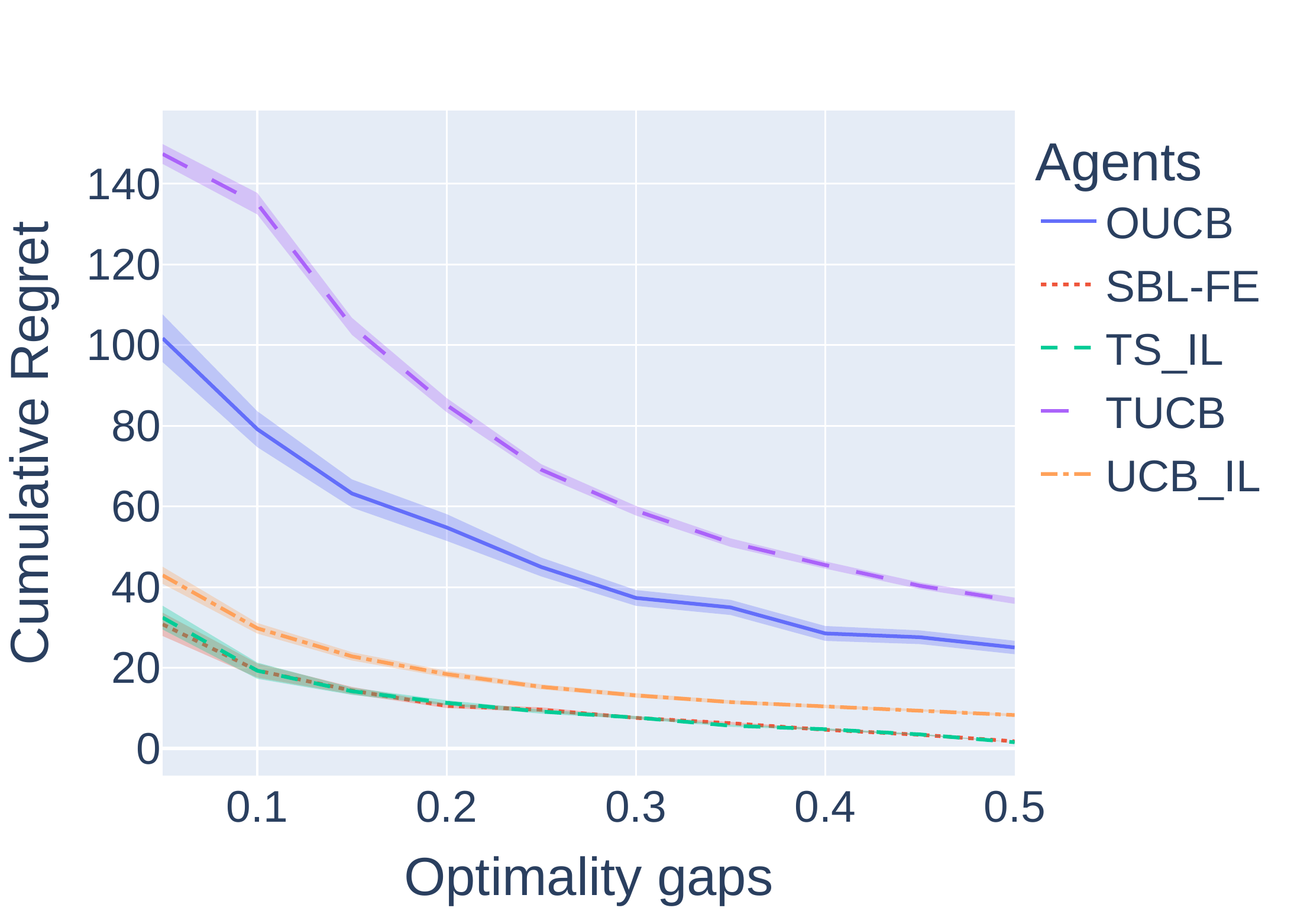}
        \subcaption*{(b) Random agent}
    \end{subfigure}
    \vspace{0.2cm}
    
    \begin{subfigure}[t]{0.48\textwidth}
        \centering
        \includegraphics[width=0.48\linewidth]{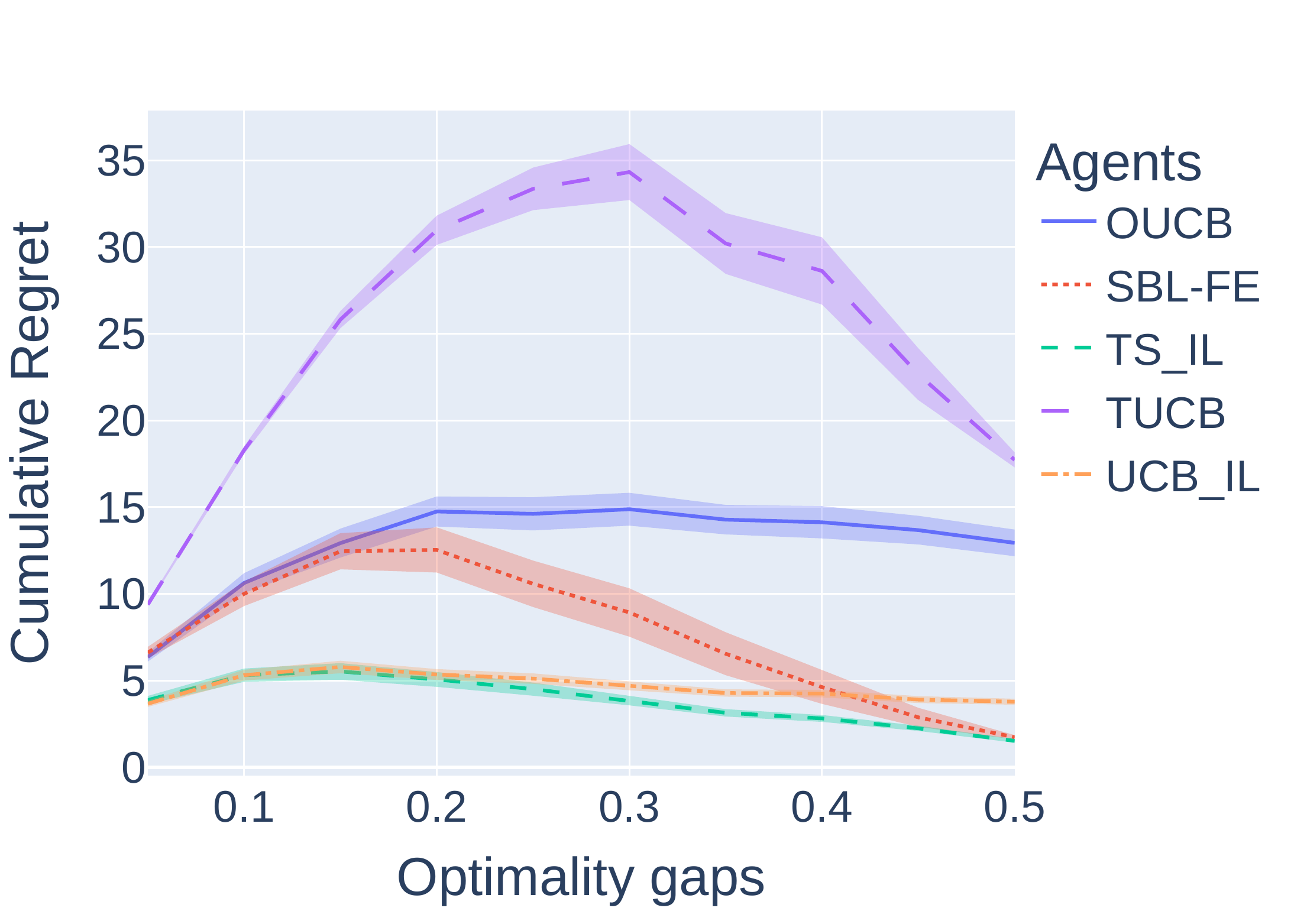}
        \hfill
        \includegraphics[width=0.48\linewidth]{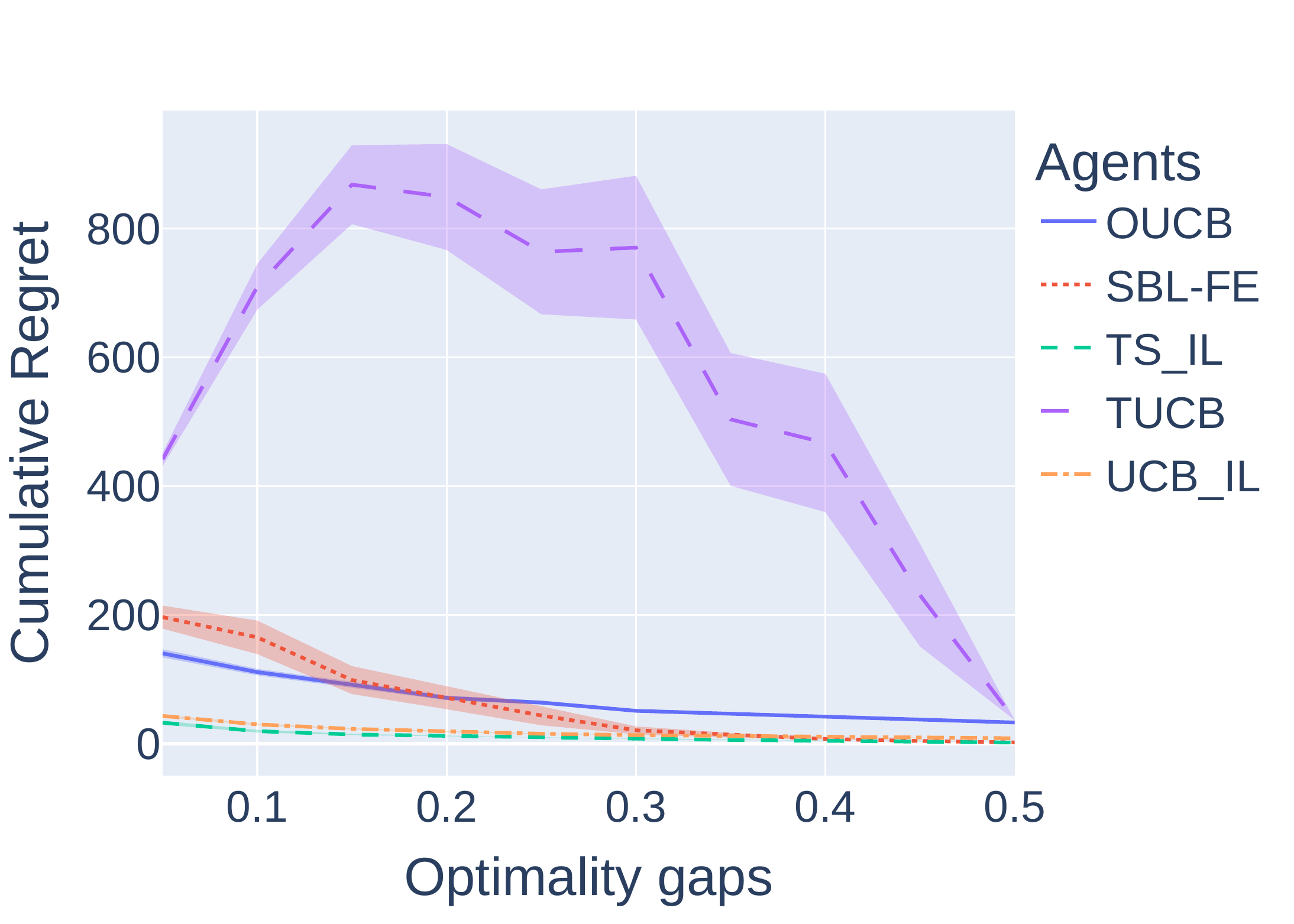}
        \subcaption*{(c) Opponent agent}
    \end{subfigure}
    \vspace{0.2cm}
    
    \begin{subfigure}[t]{0.48\textwidth}
        \centering
        \includegraphics[width=0.48\linewidth]{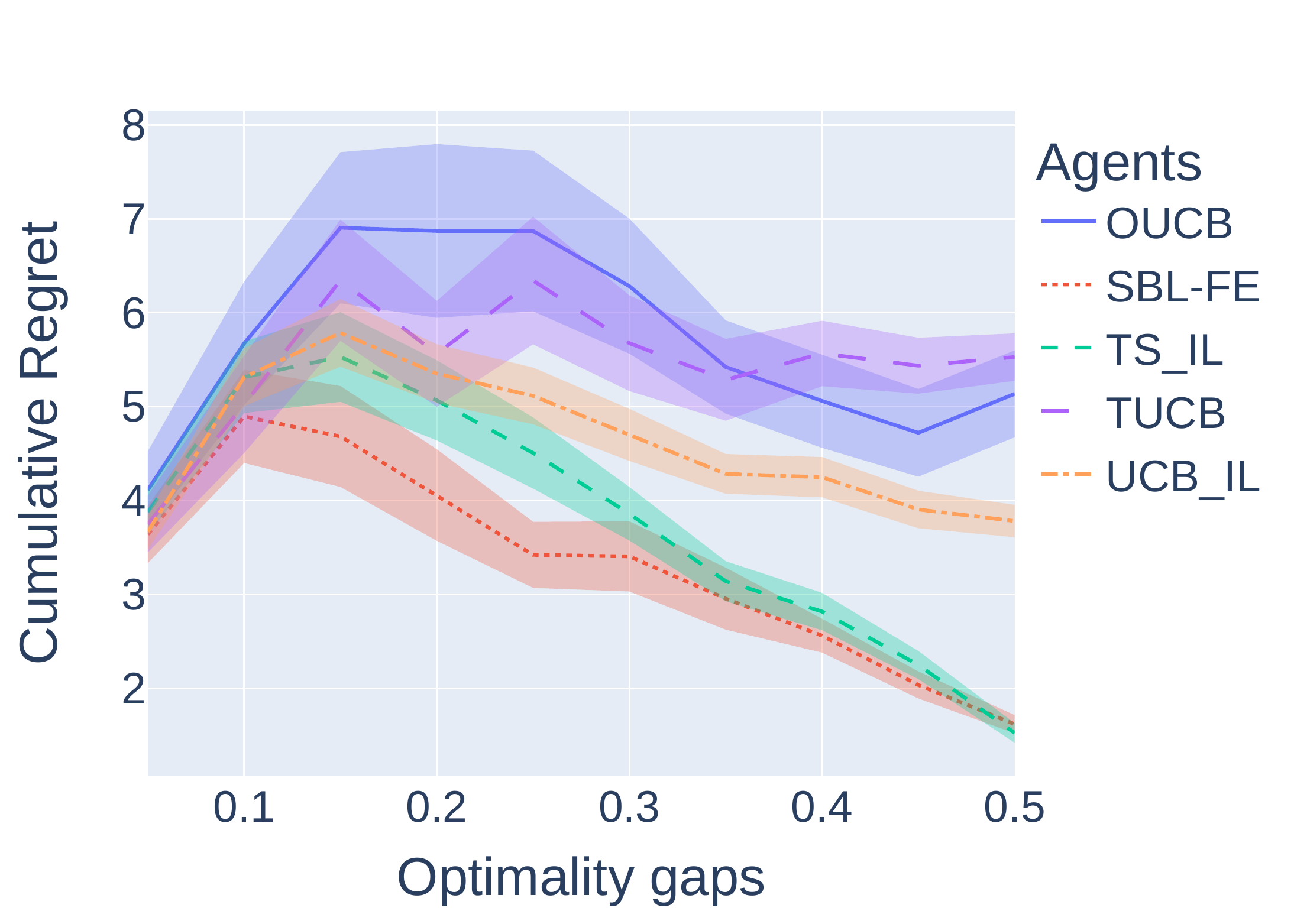}
        \hfill
        \includegraphics[width=0.48\linewidth]{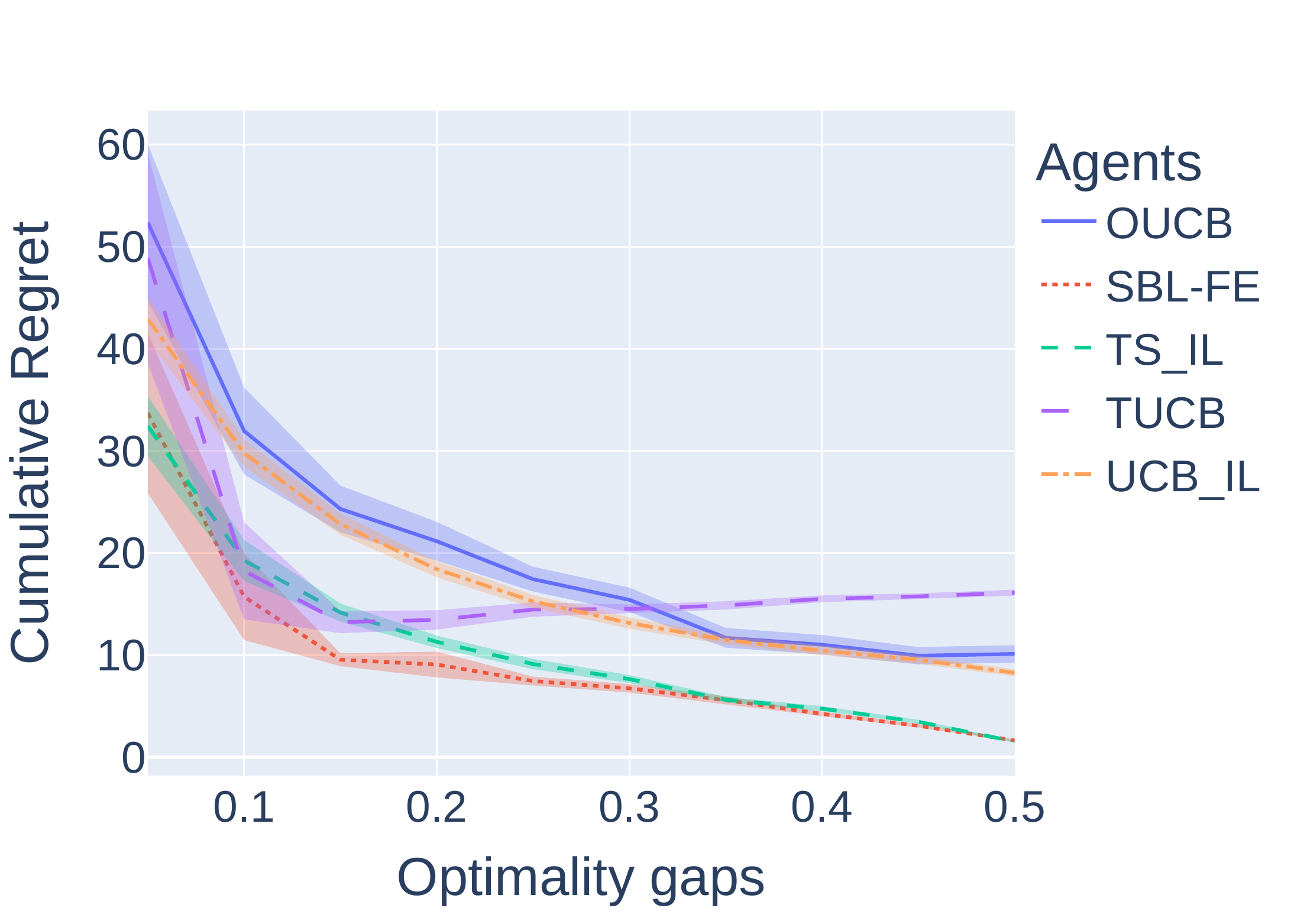}
        \subcaption*{(d) Epsilon-greedy agent}
    \end{subfigure}

    \caption{Cumulative regret performance of social learning agents (OUCB, TUCB, SBL-FE) compared to UCB and TS baselines in societies with one social learner and one non-learner or epsilon-greedy agent. Experiments were conducted for horizons 200 (left column) and 10000 (right column) in a 2-armed Bernoulli bandit problem, varying optimality gaps from 0 to 0.5.}
    \label{fig:2armed}
\end{figure}

By adjusting the number of actions and for different optimality gaps, Fig. \ref{fig:2armed} compares the cumulative regret of various social learning algorithms (OUCB, TUCB, and SBL-FE) against baseline methods such as UCB and TS. These comparisons were made within societies comprising one social learner and either a non-learner or an epsilon-greedy agent. In 2-armed multi-armed bandits, the sub-optimal agent performs the same as the opponent agent, thus, we just mention the opponent agent in the figure. Notably, we observed that when a relevant agent is present in society, social learning methods can enhance the performance of individual learning methods. Conversely, when no relevant agent exists, our social learning method performs exceptionally well in both scenarios, despite other social learning algorithms. It is worth mentioning that for small optimality gaps, the difference in cumulative regret among different methods is most prominent at a horizon of 10k, but this distinction diminishes for a horizon of 200. TUCB algorithm has some optimism to actions that are done by other agents, and when there is only an opponent agent in the environment it performs drastically poorly. 

\subsection{The robustness of our algorithm to the noise}
In this section, the robustness of our algorithm to the noise in the observation is evaluated. We assume that with probability p, the observed behavior of other agents changes randomly with uniform probability with another action. We conducted a series of experiments, as described in Section \ref{sec:ability}, using a multi-armed bandit with 10 arms to investigate the robustness of our method to the noise.  

\begin{figure}[htbp]
    \centering

    \begin{subfigure}[t]{0.48\textwidth}
        \centering
        \includegraphics[width=0.48\linewidth]{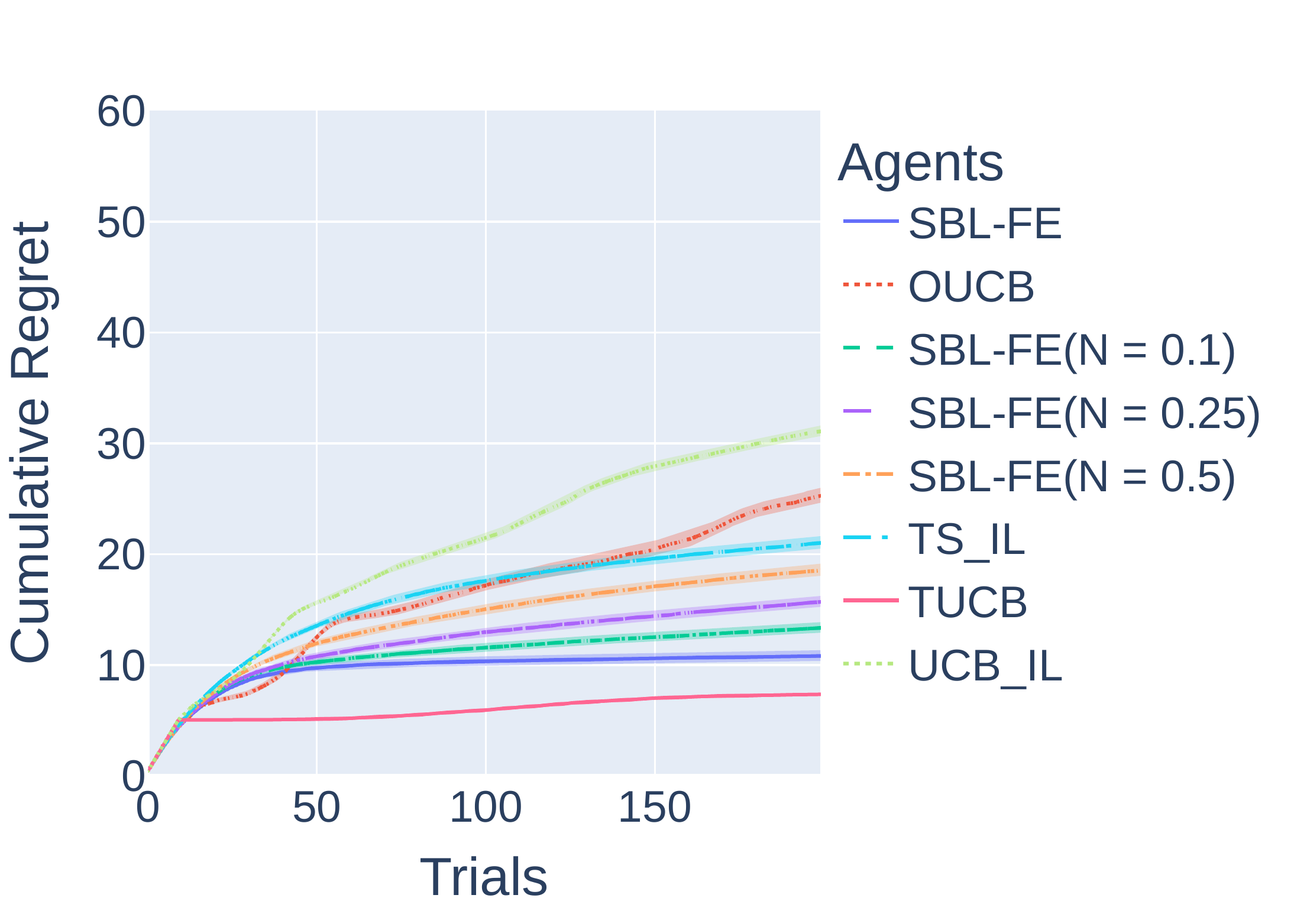}
        \hfill
        \includegraphics[width=0.48\linewidth]{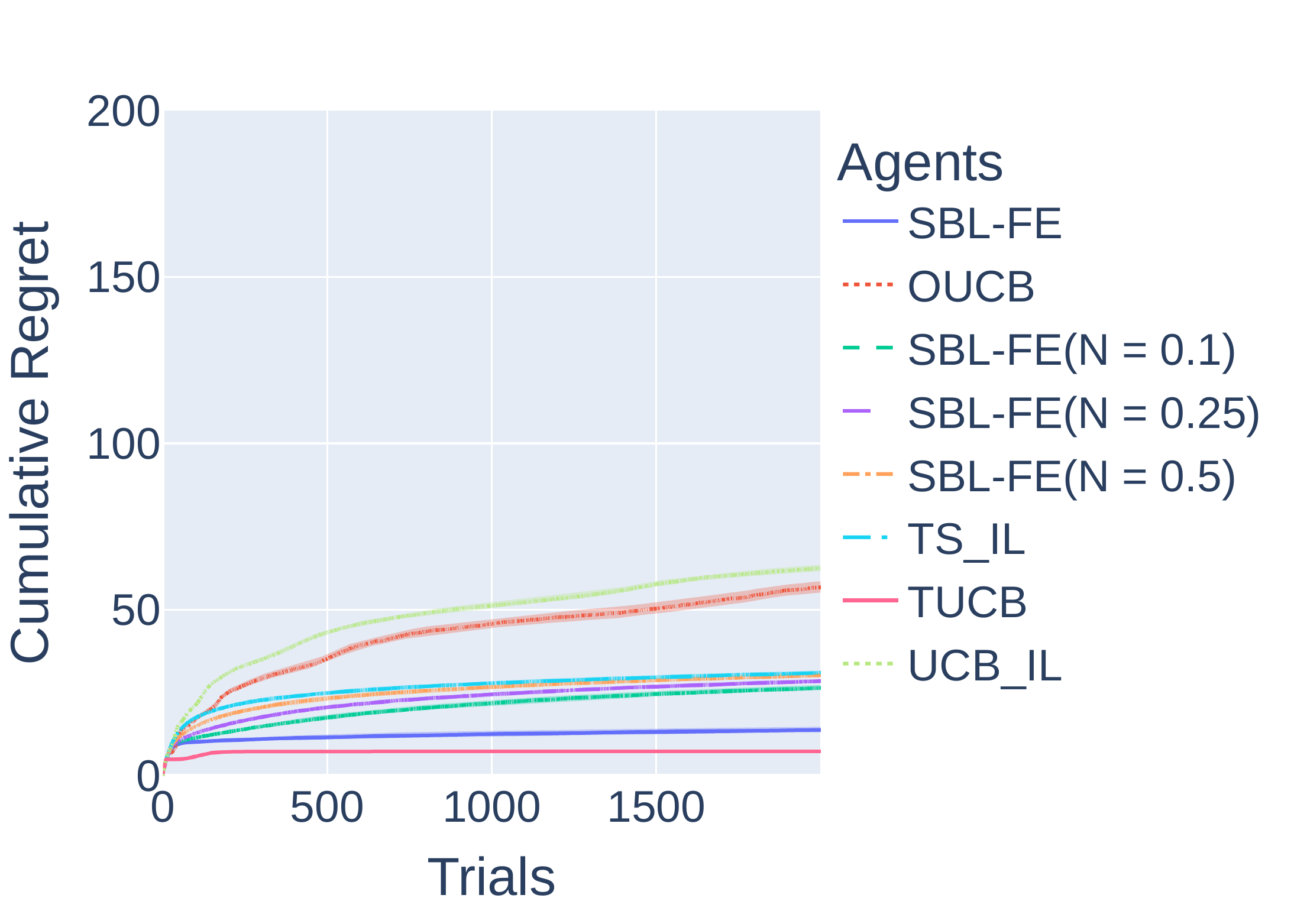}
        \subcaption*{(a) Optimal agent}
    \end{subfigure}
    \vspace{0.2cm}

    \begin{subfigure}[t]{0.48\textwidth}
        \centering
        \includegraphics[width=0.48\linewidth]{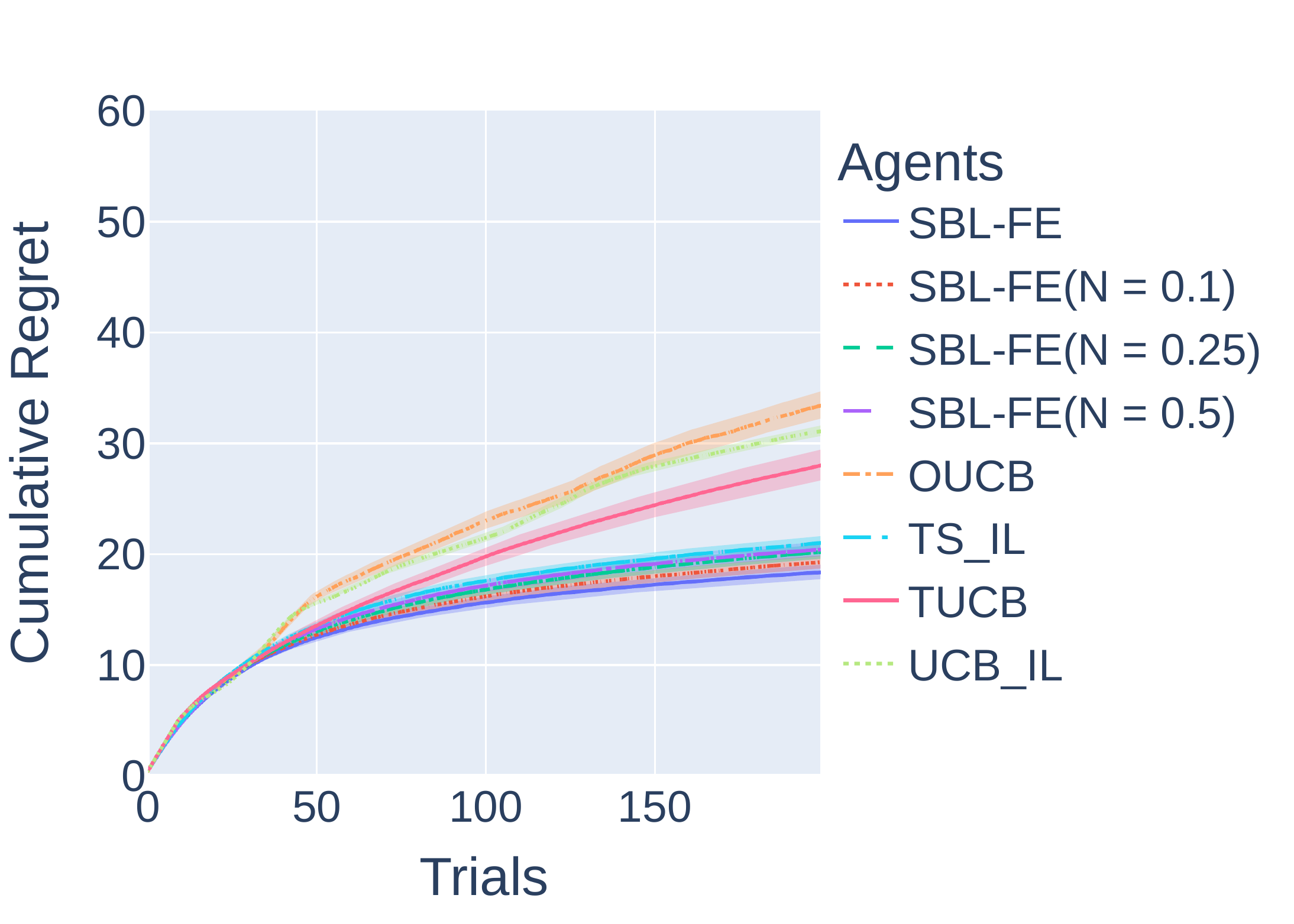}
        \hfill
        \includegraphics[width=0.48\linewidth]{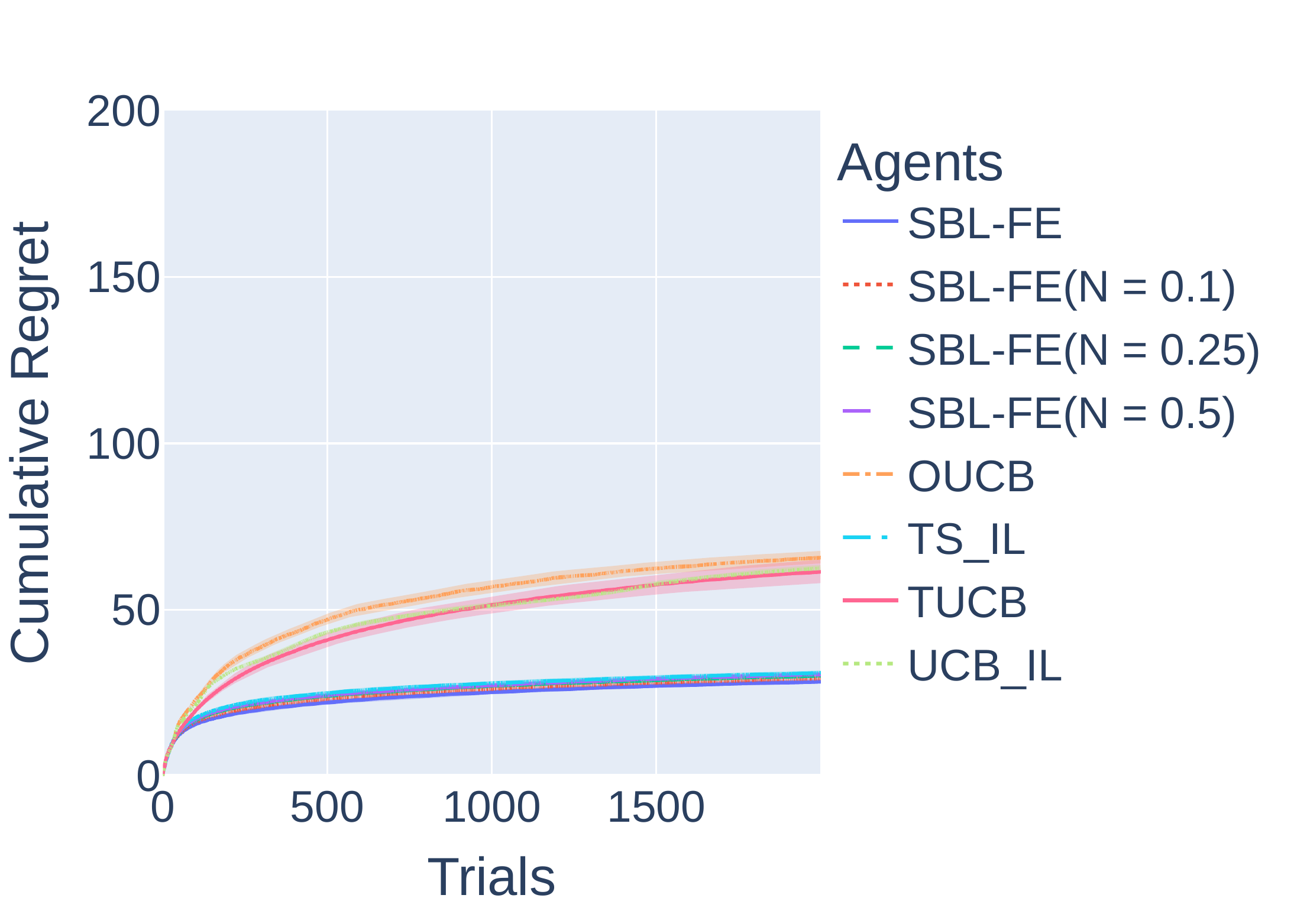}
        \subcaption*{(b) Epsilon-greedy agent}
    \end{subfigure}
    \vspace{0.2cm}

    \begin{subfigure}[t]{0.48\textwidth}
        \centering
        \includegraphics[width=0.48\linewidth]{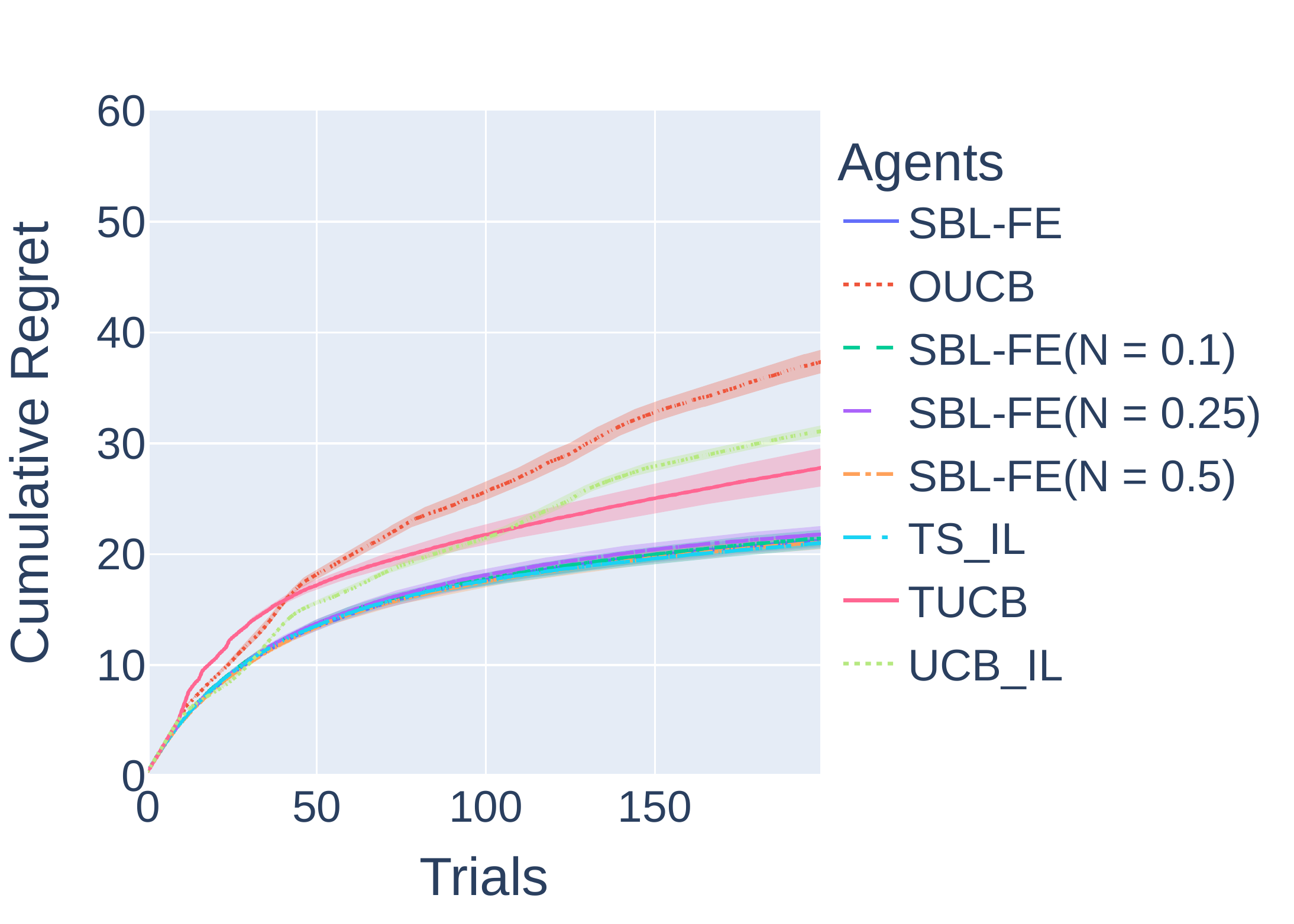}
        \hfill
        \includegraphics[width=0.48\linewidth]{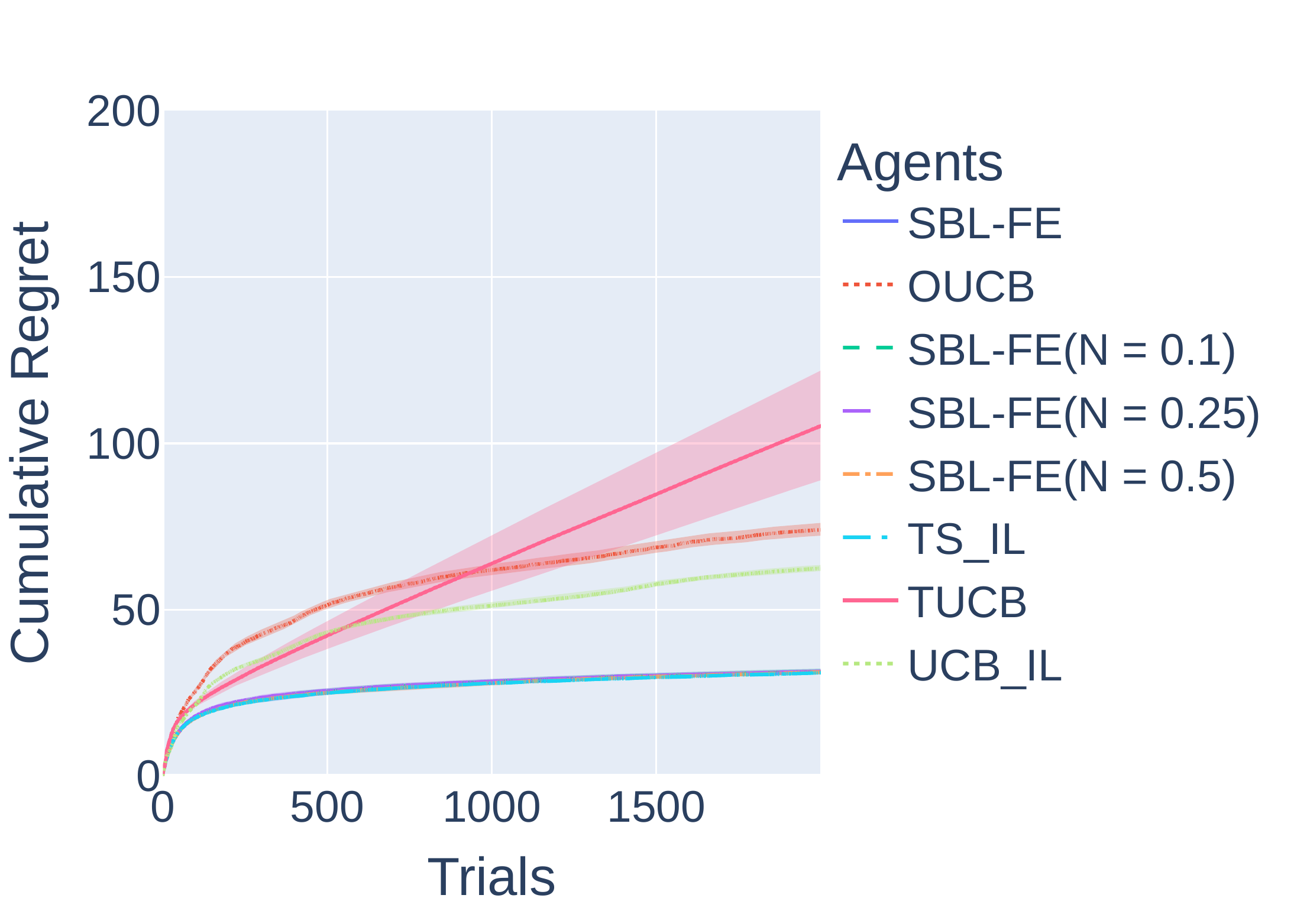}
        \subcaption*{(c) Opponent agent}
    \end{subfigure}
    \vspace{0.2cm}

    \begin{subfigure}[t]{0.48\textwidth}
        \centering
        \includegraphics[width=0.48\linewidth]{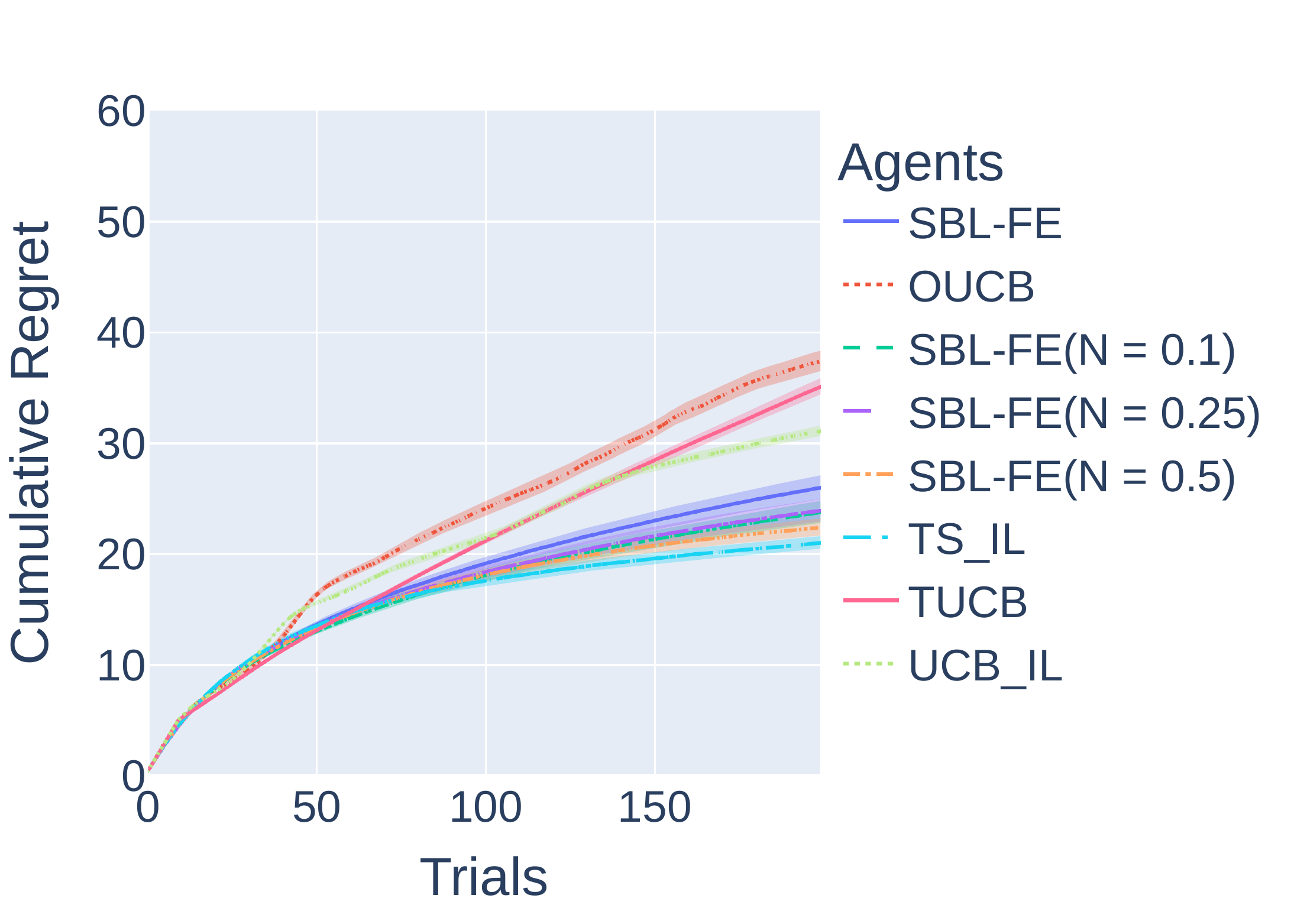}
        \hfill
        \includegraphics[width=0.48\linewidth]{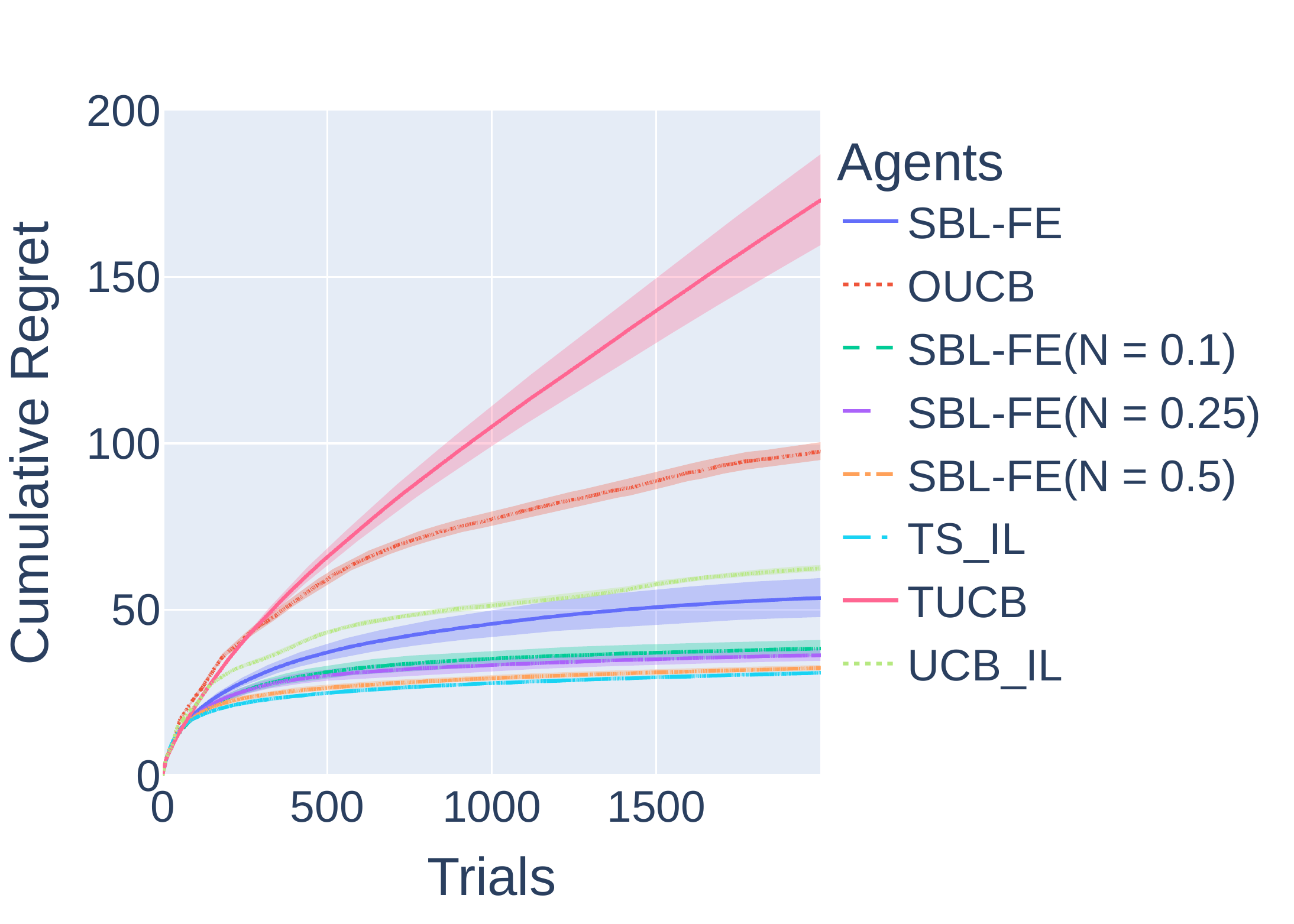}
        \subcaption*{(d) Sub-optimal agent}
    \end{subfigure}

    \caption{The cumulative regret performance of three social learning agents (OUCB, TUCB, SBL-FE with different levels of noise) along with UCB and TS as baseline methods in societies consisting of one social learner and one non-learner or epsilon-greedy. The experiments were conducted over 200 and 2000 trials for a 10-armed Bernoulli bandit problem with an optimality gap of $\Delta = 0.2$.}
    \label{fig:noise}
\end{figure}

Fig. \ref{fig:noise} shows the cumulative regret of various social learning algorithms, including UCB and TS as baseline methods, throughout the learning process over 200 and 2000 trials. The comparisons presented here involve societies consisting of one social learner alongside either a non-learner or an epsilon-greedy agent. Notably, we excluded the results when the non-learner is a random agent, as adding random noise to a random agent's action does not yield any meaningful difference. The insights drawn from this figure indicate that our method exhibits a remarkable level of robustness in the presence of noise and performs exceptionally well in diverse scenarios. Similar results are obtained with different instances of Bernoulli bandits.

\section{Discussions and Conclusion }

This research contributes to advancing social bandit learning and underscores the untapped potential of leveraging non-experts and diversity in optimizing interactive learning algorithms. In fact, regret reduction in stochastic bandit tasks is the most persuasive goal in the  community due to its practical importance in real-world applications. Parallel to human and animal societies and inspired by research in brain and cognitive sciences, in this paper, we presented a social RL method for stochastic bandit problems. As in the well-studied social animals, our method keeps the core of bandit learning intact while providing some means to benefit from observing the decisions of other agents performing the same or similar tasks in society.  Here, by society, we mean a group of artificial agents performing or learning a range of bandit tasks while their actions are observable by others, and their reward and task definitions are their private data.  To manage the confounding factors involved in investigating the proposed method in the current setting, one of the agents, called the social agent, SA, or social learner, employs social learning, and the rest learn individually. 

Benefiting from observing other agents is well studied in extreme cases, where a fully learned agent, like a mentor, is identified or where target agents, on average, select the optimal decision more often than chance. Such situations are limited to sharing the same task and prior identification of expert agents. Our method is free from such assumptions and works well in both homogeneous and heterogeneous societies composed of learning and non-fully expert agents with diverse tasks. Nevertheless, it benefits the most if an expert agent is present in the society. One of the main properties of our method is handling all such different cases in a unified form.

Our method does not need to be informed of the other agents' tasks and expertness a prior. Instead, our social learner starts learning and, along with its learning process, forms its behavior policy. That policy is based on SA's estimation of its action values, updated according to its personal experience and reward in a normal RL format, and the estimated policy of the other agents. The social agent's behavior policy is the one that has minimum free energy among all agents, including itself. That is, free energy is the measure of suitability of an agent's behavior to exploit. The suitability measure, i.e., energy, is a weighted sum of three pillars; two pillars are relative, and the other one is absolute. The first one is the similarity of the candidate behavior policy with SA's Thompson policy. This relative element enforces the centricity of SA's personal experience in forming its behavior policy. The second pillar is the similarity of candidate behavior policy with the estimated policy of a target agent. This element encodes the following target agent's behavior. The absolute pillar is the entropy of the candidate policy as the more greedy ones are preferred. Nevertheless, the agent doesn't know which deterministic policy is the optimal one. 

The proposed free-energy criterion is defined in the policy space, not in the utility or the reward space. Because the SA's information about the other agents is limited to its estimation of their policies, the agents can be heterogeneous in different dimensions, including tasks, utility functions, and action sets. The SA's original experiences are saved in the action values. However, our measure maps those experiences to the policy space through  Thompson sampling. The good point about Thompson sampling's policy is to carry SA's uncertainties to the policy space. The more SA is uncertain the less it exploits others' behavior. This point is crucial for our method as the SA's experience plays a central role in evaluating other agents because it has no other source of reliable reference. For example, in the early learning trials, our method prefers closer-to-random, not fully random, behavior policies even in the presence of greedy agents in the environment, as the greedy agents and the SA may have different tasks. But, after some trials, where SA's Thompson policy slightly distances from random, SA's behavior policy is inclined more toward the greedy agent that shares the same task with SA while avoiding other greedy policies. As a result, our method can not follow a full expert from the beginning as it has no idea about the expertise of others related to its task until it has some uncertain estimation of its action values. This property helps our SA agent to wisely exploit the experiences of related agents and to avoid following non-related ones.  In addition, the free-energy criterion provides an uncertainty-aware means for our SA agent to flexibly benefit from diverse agents, ranging from partially expert and learning agents to fully expert ones, in a unified manner. Nevertheless, in case SA is being informed about the presence of a related expert or a semi-expert agent in the society, it can mitigate the role of its own experience in forming the behavior policy by reducing constant $C$ in \eqref{free_energy_candidate_policy}.

We rigorously analyzed the agent's ability to assess and select other agents by conducting experiments in a variety of societal configurations. Specifically, we considered cases where the societies consisted of multiple heterogeneous agents with varying levels of expertise. Through empirical evaluations, we demonstrated that our method consistently outperforms alternative methods, especially when non-expert yet relevant agents are present in the societies. Noteworthy is our method's ability to cleverly identify the relevant agents, regardless of their expertise, and leverage their knowledge in scenarios where a substantial number of non-relevant agents are present. In striking contrast, alternative methods stumble in this regard. Furthermore, we highlight the robustness of our social learning method by investigating its performance across different factors, such as the number of arms, problem difficulty, and observation noise. In addition, we showed that our method is robust and works excellently in societies including learning agents with subsets of SA's decisions. 

In line with the reported results, some complementary experiments showed that our method acts strongly better than its alternatives even where all individual agents' preferred actions are sub-optimal for the SA; no individual agent and the SA have the same optimal action. Nevertheless, for example, when one of the individuals' optimal decisions is the SA's second best action, the SA tunes its behavior policy towards that agent's policy for some trials before gaining sufficient, but still limited, experience to exploit its own Thompson sampling policy more. This behavior is rooted in SA-centered expertness evaluation that helps the SA detect proper agents and experts in the absence of any external evaluation. However, our method needs some improvements to perform better in societies where other optimal decisions are sub-optimal for the SA. One solution could be using a soft method to select the SA's behavior policy from the set of free-energy minimum policies instead of greedy selection; see \eqref{est_beh_policy_eq}.   

Our free-energy measure has the capacity to include soft avoidance from a set of given policies. Extending our method to social-constrained and social-safe learning is among our future works. In that direction, having another measure, like the SA-centered expected utility-based method, will help detect the hazardous behavior of others to avoid them. In addition, we will extend our method to non-stationary tasks. Moreover, testing our method for knowledge transfer from one task to a novel one seems promising. Given the complexity involved, we did not explore societies with multiple social learners in this paper, leaving it as a potential avenue for further investigation.

We successfully tackled the main problem in RL, i.e., regret reduction. Nevertheless, it does not come for free. We have computation costs due to Thompson sampling and updating free energy. However, our results show that the target agent for social learning does not vary quickly. Therefore we can reduce the computation cost by reducing the frequency of free-energy calculation for all agents. As a future research direction, we propose integrating the cost of observations into the free energy framework as well. Our empirical results indicate that our social learner switches to individual learning when other agents fail to provide helpful information that facilitates accelerated learning. However, it is worth exploring how these agents might possess valuable insights into actions that we should avoid during the learning process. Although our current method does not utilize this information, it presents an avenue for extension and enhancement.

Unreported primary results on extending our social method to off-policy learning in MDPs are promising. However, more research is needed to fit the method to MDP tasks. Applications involving many learners,  such as autonomous driving and robotic tasks in human-AI interactions or human-in-the-loop settings, would greatly benefit from social learning in MDP tasks. These domains often involve numerous humans doing diverse tasks, including those performing the related ones, where individual exploration by RL agents can be costly, inefficient, error-prone, or even hazardous. Social learning provides a means to leverage cues from these experts, bypassing sub-optimal or resource-intensive exploration~\cite{ndousse2021emergent}. Looking ahead, one insightful direction for future work involves applying the social learning method to Markov Decision Process (MDP) problems to learn complex behavior in long-horizon tasks.

\bibliographystyle{IEEEtran}
\bibliography{bibtxt}

@book{bandura1977social,
  title={Social learning theory},
  author={Bandura, Albert and Walters, Richard H},
  volume={1},
  year={1977},
  publisher={Englewood cliffs Prentice Hall}
}

@Book{SuttonRL,
  author =       "Richard S Sutton and Andrew G Barto",
  title =        "Reinforcement learning:An introduction",
  publisher =    "the MIT Press",
  edition =      "2nd",
  year =         "2018"
}

@article{SocialAINature,  
author = { Duéñez-Guzmán, Edgar A. and  Sadedin, Suzanne and  Wang, Jane X. and  McKee, Kevin R. and  Leibo, Joel Z.},
year = {2023},
month = {03},
pages = {1181--1188},
title = {A social path to human-like artificial intelligence},
journal={Nature Machine Intelligence}
}

@article{chai2020human,
  title={Human-in-the-loop Techniques in Machine Learning.},
  author={Chai, Chengliang and Li, Guoliang},
  journal={IEEE Data Eng. Bull.},
  volume={43},
  number={3},
  pages={37--52},
  year={2020}
}

@article{tsiakas2024unpacking,
  title={Unpacking Human-AI interactions: From interaction primitives to a design space},
  author={Tsiakas, Konstantinos and Murray-Rust, Dave},
  journal={ACM Transactions on Interactive Intelligent Systems},
  volume={14},
  number={3},
  pages={1--51},
  year={2024},
  publisher={ACM New York, NY}
}

@article{laland2004social,
  title={Social learning strategies},
  author={Laland, Kevin N},
  journal={Animal Learning \& Behavior},
  volume={32},
  number={1},
  pages={4--14},
  year={2004},
  publisher={Springer}
}

@article{boyd2011cultural,
  title={The cultural niche: Why social learning is essential for human adaptation},
  author={Boyd, Robert and Richerson, Peter J and Henrich, Joseph},
  journal={Proceedings of the National Academy of Sciences},
  volume={108},
  number={Supplement 2},
  pages={10918--10925},
  year={2011},
  publisher={National Acad Sciences}
}

@inproceedings{ndousse2021emergent,
  title={Emergent social learning via multi-agent reinforcement learning},
  author={Ndousse, Kamal K and Eck, Douglas and Levine, Sergey and Jaques, Natasha},
  booktitle={International Conference on Machine Learning},
  pages={7991--8004},
  year={2021},
  organization={PMLR}
}

@article{borsa2017observational,
  title={Observational learning by reinforcement learning},
  author={Borsa, Diana and Piot, Bilal and Munos, R{\'e}mi and Pietquin, Olivier},
  journal={arXiv preprint arXiv:1706.06617},
  year={2017}
}

@phdthesis{jaques2019social,
  title={Social and affective machine learning},
  author={Jaques, Natasha},
  year={2019},
  school={Massachusetts Institute of Technology}
}

@article{hua2021learning,
  title={Learning for a robot: Deep reinforcement learning, imitation learning, transfer learning},
  author={Hua, Jiang and Zeng, Liangcai and Li, Gongfa and Ju, Zhaojie},
  journal={Sensors},
  volume={21},
  number={4},
  pages={1278},
  year={2021},
  publisher={Multidisciplinary Digital Publishing Institute}
}

@article{najar2020actions,
  title={The actions of others act as a pseudo-reward to drive imitation in the context of social reinforcement learning},
  author={Najar, Anis and Bonnet, Emmanuelle and Bahrami, Bahador and Palminteri, Stefano},
  journal={PLoS biology},
  volume={18},
  number={12},
  pages={e3001028},
  year={2020},
  publisher={Public Library of Science San Francisco, CA USA}
}

@Book{laland2017,
  author =       "Kevin N. Laland",
  title =        "Darwin's Unfinished Symphony: How Culture Made the Human Mind",
  publisher =    "Princeton University Press",
  edition =      "1st",
  year =         "2017"
}

@Book{SecretOfSuccess,
  author =       "Joseph Henrich",
  title =        "The Secret of Our Success: How Culture Is Driving Human Evolution, Domesticating Our Species, and Making Us Smarter",
  publisher =    "Princeton University Press",
  edition =      "1st",
  year =         "2015"
}

@inproceedings{duan2017one,
  title={One-Shot Imitation Learning},
  author={Duan, Yan and Andrychowicz, Marcin and Stadie, Bradly C and Ho, Jonathan and Schneider, Jonas and Sutskever, Ilya and Abbeel, Pieter and Zaremba, Wojciech},
  booktitle={NIPS},
  year={2017}
}

@article{auer2002finite,
  title={Finite-time analysis of the multiarmed bandit problem},
  author={Auer, Peter and Cesa-Bianchi, Nicolo and Fischer, Paul},
  journal={Machine learning},
  volume={47},
  number={2},
  pages={235-256},
  year={2002},
  publisher={Springer}
}

@inproceedings{lupu2019leveraging,
  title={Leveraging observations in bandits: Between risks and benefits},
  author={Lupu, Andrei and Durand, Audrey and Precup, Doina},
  booktitle={Proceedings of the AAAI Conference on Artificial Intelligence},
  volume={33},
  number={01},
  pages={6112--6119},
  year={2019}
}

@article{Zong2020SocialBL,
  title={Social Bandit Learning: Strangers Can Help},
  author={Jun Zong and Ting Liu and Zhaowei Zhu and Xiliang Luo and Hua Lin Qian},
  journal={2020 International Conference on Wireless Communications and Signal Processing (WCSP)},
  year={2020},
  pages={239-244}
}

@article{ortega2013thermodynamics,
  title={Thermodynamics as a theory of decision-making with information-processing costs},
  author={Ortega, Pedro A and Braun, Daniel A},
  journal={Proceedings of the Royal Society A: Mathematical, Physical and Engineering Sciences},
  volume={469},
  number={2153},
  pages={20120683},
  year={2013},
  publisher={The Royal Society Publishing}
}

@inproceedings{lerer2019learning,
  title={Learning existing social conventions via observationally augmented self-play},
  author={Lerer, Adam and Peysakhovich, Alexander},
  booktitle={Proceedings of the 2019 AAAI/ACM Conference on AI, Ethics, and Society},
  pages={107--114},
  year={2019}
}

@article{cheng2020policy,
  title={Policy improvement via imitation of multiple oracles},
  author={Cheng, Ching-An and Kolobov, Andrey and Agarwal, Alekh},
  journal={Advances in Neural Information Processing Systems},
  volume={33},
  pages={5587--5598},
  year={2020}
}

@inproceedings{hihn2019information,
  title={An information-theoretic on-line learning principle for specialization in hierarchical decision-making systems},
  author={Hihn, Heinke and Gottwald, Sebastian and Braun, Daniel A},
  booktitle={2019 IEEE 58th conference on decision and control (CDC)},
  pages={3677--3684},
  year={2019},
  organization={IEEE}
}

@inproceedings{ortega2011information,
  title={Information, utility and bounded rationality},
  author={Ortega, Daniel Alexander and Braun, Pedro Alejandro},
  booktitle={Artificial General Intelligence: 4th International Conference, AGI 2011, Mountain View, CA, USA, August 3-6, 2011. Proceedings 4},
  pages={269--274},
  year={2011},
  organization={Springer}
}

@article{gottwald2019bounded,
  title={Bounded rational decision-making from elementary computations that reduce uncertainty},
  author={Gottwald, Sebastian and Braun, Daniel A},
  journal={Entropy},
  volume={21},
  number={4},
  pages={375},
  year={2019},
  publisher={MDPI}
}

@article{yaman2022meta,
  title={Meta-control of social learning strategies},
  author={Yaman, Anil and Bredeche, Nicolas and {\c{C}}aylak, Onur and Leibo, Joel Z and Lee, Sang Wan},
  journal={PLoS Computational Biology},
  volume={18},
  number={2},
  pages={e1009882},
  year={2022},
  publisher={Public Library of Science San Francisco, CA USA}
}

@inproceedings{NEURIPS2023_212b143b,
 author = {Banihashem, Kiarash and Hajiaghayi, MohammadTaghi and Shin, Suho and Slivkins, Aleksandrs},
 booktitle = {Advances in Neural Information Processing Systems},
 pages = {10385--10411},
 title = {Bandit Social Learning under Myopic Behavior},
 volume = {36},
 year = {2023}
}

@article{russo2018tutorial,
  title={A tutorial on thompson sampling},
  author={Russo, Daniel J and Van Roy, Benjamin and Kazerouni, Abbas and Osband, Ian and Wen, Zheng and others},
  journal={Foundations and Trends{\textregistered} in Machine Learning},
  volume={11},
  number={1},
  pages={1--96},
  year={2018},
  publisher={Now Publishers, Inc.}
}

@book{lattimore2020bandit,
  title={Bandit algorithms},
  author={Lattimore, Tor and Szepesv{\'a}ri, Csaba},
  year={2020},
  publisher={Cambridge University Press}
}

@article{ortega2014generalized,
  title={Generalized Thompson sampling for sequential decision-making and causal inference},
  author={Ortega, Pedro A and Braun, Daniel A},
  journal={Complex Adaptive Systems Modeling},
  volume={2},
  number={1},
  pages={1--23},
  year={2014},
  publisher={Springer}
}

@article{thompson1933likelihood,
  title={On the likelihood that one unknown probability exceeds another in view of the evidence of two samples},
  author={Thompson, William R},
  journal={Biometrika},
  volume={25},
  number={3-4},
  pages={285--294},
  year={1933},
  publisher={Oxford University Press}
}

@article{von1944theory,
  title={Theory of games and economic behavior Princeton},
  author={Von Neumann, John and Morgenstern, Oskar},
  journal={Princeton University Press},
  volume={1947},
  pages={1953},
  year={1944}
}

@article{savage1954foundations,
  title={The foundations of statistics; jon wiley and sons},
  author={Savage, Leonard J},
  journal={Inc.: New York, NY, USA},
  year={1954}
}

@article{friston2009free,
  title={The free-energy principle: a rough guide to the brain?},
  author={Friston, Karl},
  journal={Trends in cognitive sciences},
  volume={13},
  number={7},
  pages={293--301},
  year={2009},
  publisher={Elsevier}
}

@article{friston2010free,
  title={The free-energy principle: a unified brain theory?},
  author={Friston, Karl},
  journal={Nature reviews neuroscience},
  volume={11},
  number={2},
  pages={127--138},
  year={2010},
  publisher={Nature publishing group}
}

@book{thorndike2017animal,
  title={Animal intelligence: Experimental studies},
  author={Thorndike, Edward},
  year={2017},
  publisher={Routledge}
}

@article{schultz1997neural,
  title={A neural substrate of prediction and reward},
  author={Schultz, Wolfram and Dayan, Peter and Montague, P Read},
  journal={Science},
  volume={275},
  number={5306},
  pages={1593--1599},
  year={1997},
  publisher={American Association for the Advancement of Science}
}

@article{mosqueira2023human,
  title={Human-in-the-loop machine learning: A state of the art},
  author={Mosqueira-Rey, Eduardo and Hern{\'a}ndez-Pereira, Elena and Alonso-R{\'\i}os, David and Bobes-Bascar{\'a}n, Jos{\'e} and Fern{\'a}ndez-Leal, {\'A}ngel},
  journal={Artificial Intelligence Review},
  volume={56},
  number={4},
  pages={3005--3054},
  year={2023},
  publisher={Springer}
}

@inproceedings{landgren2016distributed,
  title={Distributed cooperative decision-making in multiarmed bandits: Frequentist and bayesian algorithms},
  author={Landgren, Peter and Srivastava, Vaibhav and Leonard, Naomi Ehrich},
  booktitle={2016 IEEE 55th Conference on Decision and Control (CDC)},
  pages={167--172},
  year={2016},
  organization={IEEE}
}

@article{rendell2010copy,
  title={Why copy others? Insights from the social learning strategies tournament},
  author={Rendell, Luke and Boyd, Robert and Cownden, Daniel and Enquist, Marquist and Eriksson, Kimmo and Feldman, Marc W and Fogarty, Laurel and Ghirlanda, Stefano and Lillicrap, Timothy and Laland, Kevin N},
  journal={Science},
  volume={328},
  number={5975},
  pages={208--213},
  year={2010},
  publisher={American Association for the Advancement of Science}
}

@article{isomura2023experimental,
  title={Experimental validation of the free-energy principle with in vitro neural networks},
  author={Isomura, Takuya and Kotani, Kiyoshi and Jimbo, Yasuhiko and Friston, Karl J},
  journal={Nature Communications},
  volume={14},
  number={1},
  pages={4547},
  year={2023},
  publisher={Nature Publishing Group UK London}
}

@ARTICLE{979961,
  author={Ahmadabadi, M.N. and Asadpour, M.},
  journal={IEEE Transactions on Systems, Man, and Cybernetics, Part B (Cybernetics)}, 
  title={Expertness based cooperative Q-learning}, 
  year={2002},
  volume={32},
  number={1},
  pages={66-76},
  doi={10.1109/3477.979961}}

@article{zhu2023transfer,
  title={Transfer learning in deep reinforcement learning: A survey},
  author={Zhu, Zhuangdi and Lin, Kaixiang and Jain, Anil K and Zhou, Jiayu},
  journal={IEEE Transactions on Pattern Analysis and Machine Intelligence},
  year={2023},
  publisher={IEEE}
}

@article{filos2021psiphilearning,
	title={PsiPhi-Learning: Reinforcement Learning with Demonstrations using Successor Features and Inverse Temporal Difference Learning},
	author={Filos, A. and Lyle, C. and Gal, Y. and Levine, S. and Jaques, N. and Farquhar, G.},
	journal={International Conference on Machine Learning (ICML)},
	year={2021},
}

@article{ghorbani2025learning,
  title={Learning from different perspectives for regret reduction in reinforcement learning: A free energy approach},
  author={Ghorbani, Milad and Hosseini, Reshad and Shariatpanahi, Seyed Pooya and Ahmadabadi, Majid Nili},
  journal={Neurocomputing},
  volume={614},
  pages={128797},
  year={2025},
  publisher={Elsevier}
}

@inproceedings{ha2023social,
  title={Social learning spontaneously emerges by searching optimal heuristics with deep reinforcement learning},
  author={Ha, Seungwoong and Jeong, Hawoong},
  booktitle={International Conference on Machine Learning},
  pages={12319--12338},
  year={2023},
  organization={PMLR}
}

@article{hawkins2023flexible,
  title={Flexible social inference facilitates targeted social learning when rewards are not observable},
  author={Hawkins, Robert D and Berdahl, Andrew M and Pentland, Alex Sandy and Tenenbaum, Joshua B and Goodman, Noah D and Krafft, PM},
  journal={Nature Human Behaviour},
  volume={7},
  number={10},
  pages={1767--1776},
  year={2023},
  publisher={Nature Publishing Group UK London}
}

@article{witt2024humans,
  title={Humans flexibly integrate social information despite interindividual differences in reward},
  author={Witt, Alexandra and Toyokawa, Wataru and Lala, Kevin N and Gaissmaier, Wolfgang and Wu, Charley M},
  journal={Proceedings of the National Academy of Sciences},
  volume={121},
  number={39},
  pages={e2404928121},
  year={2024},
  publisher={National Academy of Sciences}
}
\clearpage
\pagenumbering{arabic}

\newpage

\appendices

\section{Proof of Theorem 1.}\label{proof}

If we substitute the \eqref{pi_Star_eq} to the free energy model (\ref{Free_Energy_agents_eq}), then after algebraic simplification we have: 

\begin{equation}\label{free_energy_star_eq}
    F\left(i, \tilde{\pi}_{{ag}_i}\right) = - c \log(Z(i)).
\end{equation}
Due to numerical implications, we add a small constant, $\xi > 0$, to the TS policy and estimated behavior policies so all the logarithms are bounded, and then normalize them. Therefore, there exists a positive constant $\epsilon = \frac{\xi}{1 + K. \xi} > 0$ such that:

\begin{equation*}
\pi_{TS}(a) e^{\frac{1}{c} \log \hat{\pi}_{a g_i}(a)} \geq \epsilon e^{\frac{1}{c} \log \epsilon} = \epsilon^{\frac{c+1}{c}} > 0.
\end{equation*}
In addition, we know that $\log \hat{\pi}_{a g_i}(a) \leq 0$, so $e^{\frac{1}{c} \log \hat{\pi}_{a g_i}(a)} \leq 1$ for all actions. As a result:

\begin{equation*}
    0 < Z(i) =\sum_a \pi_{TS}(a) e^{\frac{1}{c} \log \hat{\pi}_{a g_i}(a)} \leq \sum_a \pi_{TS}(a) = 1.
\end{equation*}
By considering all of these inequalities, we can say that $\tilde \pi_{a g_i}(a) > 0$ for all actions. Thus, if following the $\tilde \pi_{a g_i}(a)$ or $\pi_{TS}(a)$, the SA does not stop the exploration.

Now, we should analyze which behavior policy among all the policies minimizes the free energy function when there are infinite samples. For this purpose, without losing the generality, we presume that the first action is the optimal action. As the sample size approaches infinity, the confidence interval of estimation of the expected reward of each arm reaches zero. As a result, the limit of the Thompson sampling policy after normalization would be as follows: 

\begin{equation*}
    \pi_{TS}^{\infty} = \lim_{t \to \infty} \pi_{TS} = [\frac{1+ \xi}{1 + K.\xi}, \frac{\xi}{1 + K.\xi}, ..., \frac{\xi}{1 + K.\xi}].
\end{equation*}

From \ref{free_energy_star_eq} we know the minimum value of $ F\left(i, \tilde{\pi}_{{ag}_i}\right)$ is when the $Z(i)$ has its maximum value. Now, we should show that the above policy maximizes the $Z(i)$ when it is substituted as the estimated behavior policy. Because the $Z(i)$ only depends on the value of the estimated behavior policy, we take the derivative of $Z(i)$ with respect to the probability of actions in the estimated policy. Thus, the following result is taken: 

\begin{equation*}
    \frac{\partial Z(i)}{\partial \hat \pi_{{ag}_{i}}(a)} = \frac{1}{c} \pi_{TS}(a) e^{\frac{1}{c} \log \hat{\pi}_{a g_i}(a)} = \frac{1}{c} \pi_{TS}(a) (\hat \pi_{{ag}_{i}}(a))^{\frac{1}{c}} > 0.
\end{equation*}

We saw that the gradient of $Z(i)$ with respect to the probability of actions in the estimated policy is positive. In addition, each element of behavior policy is bounded between $\frac{\xi}{1 + K.\xi}$ and $\frac{1 + \xi}{1 + K.\xi}$, and their summation should be equal to 1. Thus, the extremum value of $Z(i)$ is on the one of $K$ boundaries. The same condition exists for the coefficients, so the maximum will be when the estimated behavior policy is equal to $\pi_{TS}^{\infty}$. Therefore, the minimum value of free energy happens when both the TS policy and the estimated behavior policy of the agents become similar to a one-hot vector when one corresponds to the optimal action. Therefore, the social agent, SA, selects itself unless there are other agents with the estimated behavior policy that reaches $\pi_{TS}^{\infty}$. Thus, we prove that in all situations, the SA converges with the optimal policy. 




\section{Reward distributions} \label{reward}
We conducted experiments using three distinct bandit instances belonging to the class of 10-armed Bernoulli bandits within our environment. In these instances, the rewards are binary, with the probability associated with each reward determining the value of the corresponding action. The expected values of each arm range between zero and one. We have defined the concept of "optimality gap" as the difference between the expected reward of the optimal action and the other actions. To label each Bernoulli distribution with the smallest optimality gap, we designated them as follows: 

- Bernoulli instance with an optimality gap of $\Delta = 0.2$: The expected rewards for the actions in this instance are [ 0.05, 0.1, 0.15, 0.2, 0.3, 0.4, 0.5, 0.6, 0.7, 0.9 ].

- Bernoulli instance with an optimality gap of $\Delta = 0.1$: The expected rewards for the actions in this instance are [ 0.05, 0.1, 0.2, 0.3, 0.4, 0.5, 0.6, 0.7, 0.8, 0.9 ].

- Bernoulli instance with an optimality gap of $\Delta = 0.05$: The expected rewards for the actions in this instance are [ 0.1, 0.2, 0.3, 0.4, 0.5, 0.6, 0.7, 0.8, 0.85, 0.9 ].

\section{Hyperparameters} \label{hyperparameters}
For the TS agent, we utilized the beta distribution as the belief distribution and updated it using the Bayes rule. The update involved adding the obtained reward to the alpha parameter and subtracting the obtained reward from one for the beta parameter. Initially, both alpha and beta were set to one to generate a uniform distribution. As a result, the TS algorithm does not require any fixed hyperparameters. 

As for the UCB (Upper Confidence Bound) algorithm, the only hyperparameter involved is denoted as $C$, which is related to the variance of the Subgaussian distributions of rewards. Considering the Bernoulli distribution being 1/2-subgaussian, we set the value of $C$ to 1/2. In our experimental setup, we employed the epsilon-greedy algorithms with an epsilon value of 0.9. Furthermore, we decayed the epsilon value at a rate of 0.999 per step throughout the experiments. Similar to the UCB algorithm, TUCB algorithm also consists of the hyperparameter $C$. In line with the findings reported in their paper \cite{lupu2019leveraging}, we set the value of $C$ to 2 for all our experiments. 

Regarding OUCB algorithm involves two hyperparameters: $C$, $\beta_1$, and $\beta_2$. Drawing from the recommendations provided in the referenced paper ~\cite{Zong2020SocialBL}, we set $C$ to 2 and both $\beta_1$ and $\beta_2$ to 0.5. In our proposed method, we also have two primary hyperparameters to address: $\lambda$ and $C$. We performed a grid search as part of our hyperparameter tuning process and determined the optimal values to be 0.1 and 0.5, respectively. Additionally, we used 0.15 as the coefficient for combining the uniform distribution to ensure smoothness in our estimated policy.

\newpage

\vfill

\end{document}